\documentclass[11pt]{article}

\PassOptionsToPackage{table}{xcolor}
\PassOptionsToPackage{sort&compress}{natbib}
\PassOptionsToPackage{colorlinks=true,linkcolor=blue,citecolor=blue,urlcolor=blue}{hyperref}

\usepackage[final]{acl}

\usepackage[T1]{fontenc}
\usepackage[utf8]{inputenc}
\usepackage{times}
\usepackage{microtype}
\usepackage{graphicx}
\usepackage{booktabs}
\usepackage{makecell}
\usepackage{multirow}
\usepackage{array}
\usepackage{amsmath}
\usepackage{amssymb}
\usepackage{xcolor}
\usepackage{enumitem}
\usepackage{pifont}
\usepackage{etoolbox}
\usepackage{tikz}
\usepackage{pgfplots}
\pgfplotsset{compat=1.18}
\usepgfplotslibrary{groupplots}
\definecolor{FaultNavy}{HTML}{172A3A}
\definecolor{FaultSlate}{HTML}{5B677A}
\definecolor{FaultTeal}{HTML}{1B998B}
\definecolor{FaultGreen}{HTML}{4F8A5B}
\definecolor{FaultRed}{HTML}{B85C5C}
\definecolor{FaultGold}{HTML}{C99A2E}
\definecolor{FaultMist}{HTML}{F6F8FB}
\definecolor{FaultMint}{HTML}{EAF7F3}
\definecolor{FaultRose}{HTML}{F8ECEC}
\definecolor{FaultSand}{HTML}{FCF4E4}
\definecolor{FaultSky}{HTML}{E8F2FF}
\definecolor{FaultLilac}{HTML}{EFEDFF}
\definecolor{FaultLine}{HTML}{D8E0EA}

\definecolor{ComboTV}{HTML}{B85C5C}
\definecolor{ComboTA}{HTML}{4A7BA8}
\definecolor{ComboVA}{HTML}{6B9E40}
\definecolor{ComboTVA}{HTML}{6B4C8E}

\hypersetup{
  colorlinks=true,
  linkcolor=FaultTeal,
  citecolor=FaultTeal,
  urlcolor=FaultTeal
}

\DeclareRobustCommand{\method}{\texorpdfstring{\textcolor{FaultNavy!92!black}{\textbf{\textsc{SCEval}}}}{SCEval}}
\newcommand{\methodfull}{Structure-Corruption Evaluation}

\newcommand{\xmark}{\textcolor{FaultRed}{\ding{55}}}
\newcommand{\gain}[1]{\textcolor{FaultGreen}{\ensuremath{\uparrow}~#1}}
\newcommand{\drop}[1]{\textcolor{FaultRed}{\ensuremath{\downarrow}~#1}}

\newcommand{\finding}[1]{\par\smallskip\noindent\tikz[baseline=-0.08ex]\draw[FaultNavy,line width=1.2pt,rounded corners=0.7pt] (0,0) -- (0,0.7em);\hspace{0.48em}{\textcolor{FaultNavy!85!blue}{\textbf{#1}}}\quad}

\newcommand{\modelname}[1]{{\rmfamily #1}}

\newcommand{\faultrow}{\rowcolor{FaultMint}}

\newcommand{\softrow}{\rowcolor{FaultMist}}
\newcommand{\modelrow}{\rowcolor{FaultSky!55}}
\newcommand{\variantrow}{\rowcolor{FaultLilac!55}}
\newcommand{\tabtag}[2]{\begingroup\setlength{\fboxsep}{1.15pt}\colorbox{#1!12}{\textcolor{#1!78!black}{\scriptsize\sffamily\bfseries #2}}\endgroup}

\newcommand{\proptag}{\tabtag{FaultNavy}{PROP}}
\newcommand{\opentag}{\tabtag{FaultTeal}{OPEN}}

\newcommand{\interfacetag}{\tabtag{FaultSlate}{IFACE}}

\newcommand{\sourcetag}{\tabtag{FaultSlate}{SRC}}
\newcommand{\verifytag}{\tabtag{FaultGreen}{VERIFY}}
\newcommand{\faultgroup}[2]{\softrow\multicolumn{#1}{l}{\sffamily\bfseries\textcolor{FaultNavy}{#2}}\\}

\newcommand{\maincell}[2]{\textbf{#1}\hspace{0.6pt}{\fontsize{4.6pt}{5pt}\selectfont\textcolor{FaultRed!82!black}{$\downarrow$#2}}}
\newcommand{\maingain}[2]{\textbf{#1}\hspace{0.6pt}{\fontsize{4.6pt}{5pt}\selectfont\textcolor{FaultGreen!72!black}{$\uparrow$#2}}}
\newcommand{\mainflat}[1]{\textbf{#1}\hspace{0.6pt}{\fontsize{4.6pt}{5pt}\selectfont\textcolor{FaultSlate}{$\pm$0.00}}}

\newcommand{\mainband}[1]{\multicolumn{16}{@{}l}{\textit{\textcolor{FaultSlate!85}{#1}}}\\[1pt]}

\newcommand{\jointbandsix}[1]{\multicolumn{6}{@{}l}{\textit{\textcolor{FaultSlate!85}{#1}}}\\[1pt]}
\newcommand{\opname}[1]{\textcolor{FaultNavy!80!red}{\textbf{#1}}}
\newcommand{\sevcell}[1]{{\fontsize{6.6pt}{7.4pt}\selectfont $#1$}}
\newcommand{\repmark}{$^{\,\mathrm{r}}$}

\newcommand{\acc}{\mathrm{Acc}}
\newcommand{\dropbase}{\Delta_{\mathrm{base}}}
\newcommand{\dropsingle}{\Delta_{\mathrm{single}}}
\newcommand{\fault}{F}
\newcommand{\mislead}{M}
\newcommand{\slope}{S}
\newcommand{\shortcutgap}{G}
\newcommand{\trust}{T}

\title{Modality Fault Lines:\\
Structural Corruptions Reveal Fragile Omni-Modal Reasoning}

\author{
\textbf{Zhaolu Kang}\equal\textsuperscript{1,2},
\textbf{Meixin Wu}\equal\textsuperscript{2},
\textbf{Yu Xue}\equal\textsuperscript{2},
\textbf{Yingjie He}\equal\textsuperscript{2},\\
\textbf{Qiming Shi}\textsuperscript{3},
\textbf{Lei Wei}\textsuperscript{2},
\textbf{Yidi Wang}\textsuperscript{2},\\
\textbf{Richeng Xuan}\corr\textsuperscript{1},
\textbf{Zhichao Hu}\corr\textsuperscript{1}
\\[2pt]
\textsuperscript{1}Hunyuan, Tencent,   
\textsuperscript{2}Peking University,
\textsuperscript{3}Zhejiang University\\
\texttt{\selectfont{zlkang25}@stu.pku.edu.cn}
}

\newcommand{\equal}{\textsuperscript{*}}
\newcommand{\corr}{\textsuperscript{\dag}}

\begin{document}
\maketitle

\begingroup
  \renewcommand\thefootnote{\fnsymbol{footnote}}
  \footnotetext[1]{Equal contribution.}
  \footnotetext[2]{Corresponding author.}
  % \footnotetext[3]{Project lead.}
\endgroup

\begin{abstract}
Omni-modal large language models are increasingly evaluated on clean text--vision--audio inputs, where every channel is present, synchronized, and readily interpretable. Such scores are often taken as evidence of robust cross-modal fusion, but clean evaluation cannot tell whether success depends on stable cross-modal structure or on cues sufficient only in intact inputs. To address this gap, we define a \emph{modality fault line}: a boundary at which model behavior becomes unstable when a modality remains present and human-interpretable, but its internal evidence structure is perturbed. We introduce \method{} (\emph{\methodfull}), a diagnostic evaluation protocol that keeps the question, answer space, and modality channels fixed while applying controlled structural corruptions to text, vision, and audio individually and jointly. Built from $273$ human-verified tri-modal examples from Social-IQ, OmniBench, and VALOR, \method{} evaluates $14$ proprietary and open-source omni-modal systems. The results show that structural corruption lowers clean accuracy, text--vision damage forms the most stable shared fault line, and multi-modal degradation is non-additive rather than a simple function of the number of corrupted modalities. Clean omni-modal accuracy therefore does not establish that a model will remain reliable when cross-modal evidence becomes structurally unreliable.
\end{abstract}

\begin{figure}
    \centering
    \includegraphics[width=0.8\linewidth]{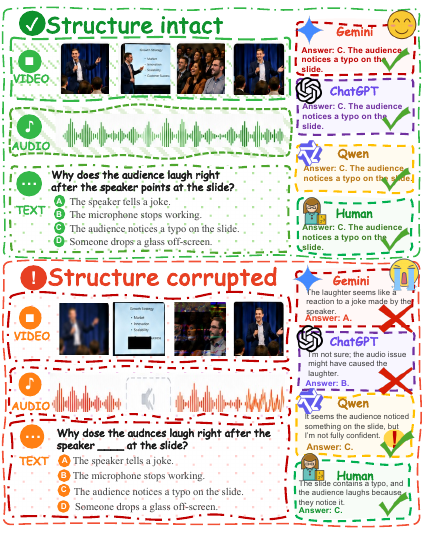}
    \vspace{-12pt}
    \caption{Teaser illustration of modality fault lines. \method{} keeps text, vision, and audio present while damaging their internal evidence structure, revealing failures that clean all-modality evaluation can hide.}
    \vspace{-6pt}
    \label{fig:teaser}
\end{figure}
\section{Introduction}
\label{sec:intro}

Omni-modal large language models promise a simple interface for multimodal reasoning: a user provides a question together with visual and acoustic evidence, and one model produces a single answer grounded in all available streams~\citep{team2023gemini,team2026qwen3}. This promise has made clean text--vision--audio benchmarks the default proof of progress---if accuracy rises when all modalities are present, the model is treated as a better cross-modal reasoner. Yet clean evaluation quietly gives the model the easiest possible version of the task: text, vision, and audio arrive simultaneously, well-formed, and mutually redundant, so a model can be correct without actually binding the three streams into a stable reasoning structure.

\vspace{-6pt}
\paragraph{Clean accuracy is not fusion \ding{182}.}
Clean evaluation does not tell us \emph{when} or \emph{how} models use multiple modalities: a tri-modal item can be solved through a transcript cue, an obvious visual object, an audio event, an answer prior, or a shortcut that happens to agree with the label. Missing-modality ablations go to the opposite extreme---removing a channel entirely changes the task and conflates modality reliance with distribution shift. The deployment-relevant region lies between: the modality is still present and human-interpretable, but its internal evidence structure has been damaged. We call the resulting hidden boundary a \emph{modality fault line}---the point at which apparent omni-modal competence breaks once the channel a model implicitly trusts becomes noisy, fragmented, or structurally unreliable.

\vspace{-6pt}
\paragraph{The missing diagnostic \ding{183}.}
A useful robustness test for omni-modal models must answer three questions that standard protocols leave entangled:
\begin{itemize}[leftmargin=1.25em,topsep=2pt,itemsep=1pt]
    \item \textbf{\textit{Which channel carries the answer?}} Clean accuracy cannot separate genuine cross-modal binding from single-channel shortcuts, because all evidence streams are simultaneously well-formed.
    \item \textbf{\textit{What happens when a channel weakens but does not disappear?}} Real inputs degrade internally---OCR-like text errors or shuffled phrases, occluded or blurry frames, muted or distorted audio---rather than vanish outright.
    \item \textbf{\textit{How do weakened channels interact?}} Single-modality studies reveal local sensitivity but cannot tell whether two damaged channels compensate, conflict, or trigger a new collapse mode.
\end{itemize}

\vspace{-4pt}
\paragraph{SCEval: keep the modality, break the structure \ding{184}.}
We introduce \method{} (\emph{\methodfull}) to make this boundary measurable. The design choice is deliberately simple: do not remove a modality; keep it present and corrupt only its internal structure. From a $300$-candidate pool sampled across Social-IQ~\citep{zadeh2019social}, OmniBench~\citep{li2026omnibench}, and VALOR~\citep{liu2024valor}, a third-party team verifies example validity, tri-modal answerability, and corrupted-input interpretability, yielding $N{=}273$ human-verified base examples. Each example is evaluated under the clean all-modality condition and under a curated family of single-, dual-, and tri-modal structural corruptions, with question, answer options, gold answer, and modality channels held fixed. Stochastic operators are aggregated as the mean over up to three random variants, and the worst-variant gap is logged alongside to surface seed-level fragility without making the primary metric a one-seed outlier.

\finding{Main Message}
We observe an asymmetric behavioral pattern: at fixed heavy visual noise, mildly corrupted text ($t30$) is associated with lower accuracy than heavily corrupted text ($t70$) for all 14 evaluated models (Table~\ref{tab:allmodel-panel}). This contrast does not by itself identify the underlying cross-modal mechanism.

Our contributions are:
\begin{itemize}[leftmargin=1.25em]
\vspace{-8pt}
    \item[\ding{72}] \textbf{\textit{Phenomenon.}} We identify \emph{modality fault lines}: latent failures that clean all-modality inputs and missing-modality ablations both miss, because the modality remains present while its internal evidence structure is damaged.
    \vspace{-8pt}
    \item[\ding{72}] \textbf{\textit{Protocol.}} We propose \method{}, a structure-corruption protocol that keeps the task and modality channels fixed while perturbing text, vision, and audio individually and jointly, with mean-variant aggregation and a worst-variant diagnostic.
    \vspace{-8pt}
    \item[\ding{72}] \textbf{\textit{Benchmark.}} We construct a human-verified tri-modal benchmark of $273$ examples from Social-IQ, OmniBench, and VALOR, covering clean inputs and full single-, dual-, and tri-modal corruption variants under a unified directory and annotation scheme.
    \vspace{-8pt}
    \item[\ding{72}] \textbf{\textit{Findings.}} We show that structural corruption reliably lowers clean performance, that text--vision damage is the most stable shared fault line, and that cross-modal degradation is non-additive and model-dependent rather than a simple function of corruption count.
\end{itemize}

\section{Structure-Corruption Evaluation}
\label{sec:method}

\method{} probes \emph{modality fault lines} by holding the task and modality channels fixed while corrupting the internal structure of one or more modalities. The design choice that distinguishes it from missing-modality ablations is that every channel remains present and human-interpretable throughout, so the resulting drops measure structural fragility rather than distribution shift; the evaluation matrix combines single-modality severity curves with dual- and tri-modal joint corruptions over the same base examples.

\subsection{Problem setup}
\label{sec:setup}

Each example consists of a question $q$, an answer space (multiple-choice options or an answer target), and three evidence channels: text $x_t$, visual input $x_v$, and audio $x_a$. A clean omni-modal model predicts
\begin{equation}
    \hat{y}=f(x_t,x_v,x_a,q),
\end{equation}
and the clean all-modality baseline accuracy is computed on the uncorrupted version of the $273$ base examples introduced in \S\ref{sec:dataset}.

A \emph{structure corruption operator} $c_m^{s,r}(\cdot)$ targets modality $m\in\{t,v,a\}$ at severity $s$ and random variant $r$ when applicable. The operator preserves the channel's container (the text string, the visual frames, or the audio waveform) and damages only its internal evidence---e.g.\ a text operator deletes or shuffles words rather than swapping the prompt, a visual operator adds noise or occludes regions rather than substituting the image, and an audio operator mutes or removes a segment rather than replacing the soundtrack. Combined corruption applies operators to multiple modalities of the same base example,
\begin{equation}
    \tilde{x}=\big(c_t(x_t),\,c_v(x_v),\,c_a(x_a),\,q\big),
\end{equation}
where one, two, or three of the corruption operators may be active while $q$, the answer options, and the original gold label $y$ remain fixed. Whether the original gold answer remains defensible after corruption is assessed separately in Appendix~\ref{app:human-reference}. This sample-paired design is what lets \method{} compute the clean-versus-corrupted contrast at the example level.

\subsection{Corruption taxonomy}
\label{sec:taxonomy}

\begin{figure}[t]
\centering
\includegraphics[width=0.85\linewidth]{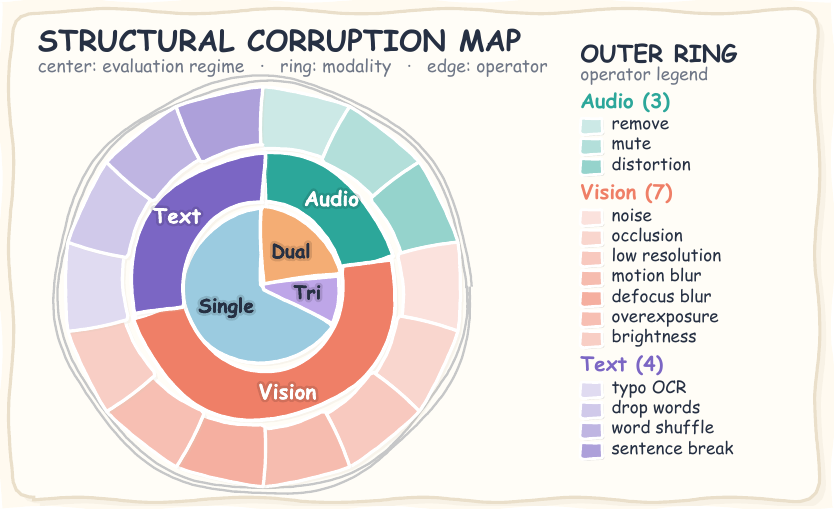}
\vspace{-8pt}
\caption{The fourteen structural corruption operators used by \method{} (4 text, 7 vision, 3 audio). The taxonomy is curated rather than exhaustive: it targets cross-modal evidence assembly (word identity/order, dense visual signal, audio presence) and excludes trivially recoverable perturbations; per-operator definitions, severity parameterisations, and rejection criteria are in Appendix~\ref{app:operator-defs} (Table~\ref{tab:corruption-taxonomy}).}
\vspace{-10pt}
\label{fig:corruption-taxonomy}
\end{figure}

Figure~\ref{fig:corruption-taxonomy} shows the fourteen operators \method{} uses, grouped into four text, seven visual, and three audio families. The taxonomy is deliberately \emph{curated, not exhaustive}: rather than enumerating every textbook degradation, we select operators that target the harder structural fault lines we expect a robust omni-modal model to handle, while keeping every modality channel present and human-interpretable. Four design rules fix what an operator is allowed to do and how it is scored: (i) only \emph{structural}, not semantic, damage; (ii) sample-level paired comparison with the same gold answer; (iii) graded severity in $\{10,30,50,70\}$; and (iv) up-to-three stochastic random variants aggregated by the mean-variant rule of \S\ref{sec:meanagg}. Combined experiments use representative high-coverage dual- and tri-modal pressure conditions (text--vision, text--audio, vision--audio, and text--vision--audio). Per-operator definitions, severity parameterisations, and rejection rules are deferred to Appendix~\ref{app:operator-defs} (Table~\ref{tab:corruption-taxonomy}).

\subsection{Benchmark construction}
\label{sec:dataset}

\paragraph{Source benchmarks.}
\method{} is built on three publicly released omni-modal benchmarks chosen to span different evidence styles: \textbf{Social-IQ}~\citep{zadeh2019social}, a video QA benchmark for social reasoning that requires both speech and visual cues; \textbf{OmniBench}~\citep{li2026omnibench}, designed to test joint text--image--audio reasoning; and \textbf{VALOR}~\citep{liu2024valor}, whose video clips contain naturally co-occurring visual and acoustic events. Mixing the three sources prevents the fault-line measurements from being driven by any single benchmark's idiosyncrasies.

\vspace{-6pt}
\paragraph{Candidate pool.}
We sample $100$ examples per source, yielding a $300$-candidate pool. For every candidate we collect (i) the original video, (ii) the audio track extracted from the same recording, (iii) the textual question with multiple-choice options and a gold answer, and (iv) source metadata, organised under a unified directory and re-encoded to a common format so that the same operators apply uniformly across sources.

\vspace{-6pt}
\paragraph{Example-level verification.}
A third-party annotation team verifies, for each candidate, (a) example validity (well-formed question, single defensible gold answer, all three modalities present in the clean condition); (b) tri-modal answerability of the gold answer from the union of the three clean modalities; and (c) the absence of annotator-side issues (broken media, ambiguous answer, modality leakage in text). Items failing any check, or flagged with annotator-side issues, are dropped at the original-sample level, leaving $N{=}273$ verified base examples that constitute the clean all-modality baseline used throughout the paper.

\vspace{-6pt}
\paragraph{Cell-level corrupted-variant verification.}
The same team further judges every \emph{corrupted variant} of audio and vision: of the $19{,}944$ audio + vision variants sent for verification, $16{,}049$ ($80.5\%$) are retained as perceptually interpretable and contribute to the headline numbers. Text-side corruptions are deterministic and human-readable by construction, so the four text operators are not subject to per-variant annotation. Per-operator pool sizes and retention rates are reported in Appendix~\ref{app:human-filtering}, Table~\ref{tab:annot-modality}.

\vspace{-6pt}
\paragraph{Composition and release.}
Table~\ref{tab:dataset-composition} reports the per-source counts of the verified base set together with the modality channels each source natively provides. Fault-line analyses in \S\ref{sec:results} are reported on the pooled $273$-example set; per-source breakdowns are kept in the appendix as a robustness check. The annotation manifest---per-sample validity flag, tri-modal answerability flag, and per-corruption interpretability flag---is released alongside the benchmark so that other groups can reconstruct the $273$-example base set and audit the filtering rule.

\begin{table}[t]
\centering
\small
\setlength{\tabcolsep}{7pt}
\renewcommand{\arraystretch}{0.9}
\begin{tabular}{l c c}
\toprule
\textbf{Source} & \textbf{Modalities} & \textbf{\# Examples} \\
\midrule
\softrow \sourcetag{} Social-IQ & video, audio, text & 100 \\
OmniBench                       & image, audio, text & 77 \\
\softrow VALOR                  & video, audio, text & 96 \\
\midrule
\faultrow \textbf{Total} & --- & \verifytag{} \textbf{273} \\
\bottomrule
\end{tabular}
\vspace{-6pt}
\caption{Per-source composition of \method{}. From a 300-candidate pool (100 per source), third-party verification of example validity, tri-modal answerability, and corrupted-variant interpretability retains $273$ examples as the base set.}
\vspace{-10pt}
\label{tab:dataset-composition}
\end{table}

\subsection{Evaluation protocol}
\label{sec:evalproto}

\paragraph{Mean random-variant aggregation.}
\label{sec:meanagg}
For stochastic corruptions, the same original sample is evaluated under up to three independently sampled random variants of the same operator at the same severity, and the reported accuracy is the \emph{mean} across these variants so that the headline number reflects typical behaviour rather than any single seed. For model $f$, condition $c$, sample $i$, and valid variant $r\in R_{i,c}$, let $z_{i,r,c}=\mathbb{1}[f(c^{r}(x_i))=y_i]$ be the variant-level correctness indicator. We aggregate at the (sample, variant) trial level,
\begin{equation}
\acc(c)=
\frac{1}{|I_c|}
\sum_{i\in I_c}
\frac{1}{|R_{i,c}|}
\sum_{r\in R_{i,c}} z_{i,r,c}.
\end{equation}

where $I_c$ is the set of samples with at least one valid output. When every retained sample contributes the same number of variants this reduces to $\acc(c)=|R|^{-1}\sum_{r\in R}\acc_r(c)$, matching the intuition of ``average accuracy over three random runs''. Missing variants are excluded from $R_{i,c}$ rather than counted as correct; judged but unparseable responses are treated as missing
for accuracy estimation and as incorrect for
significance testing.

\paragraph{Worst-variant diagnostic.}
To expose how much of the headline mean is bolstered by lucky variants, we additionally report the worst-variant accuracy
\begin{equation}
    \acc^{\mathrm{worst}}(c) = \min_{r\in R}\, \acc_r(c).
\end{equation}
The gap $\acc(c)-\acc^{\mathrm{worst}}(c)$ quantifies seed-level variance: a small gap means the model degrades consistently across variants; a large gap signals that a single unlucky variant dominates the apparent fault. The mean$-$worst gap is reported per (model, modality) at each model's weakest single-modality cell in Table~\ref{tab:variant-stability} (Appendix~\ref{app:reliability}).

\paragraph{Primary drop metrics.}
For any corruption condition $c$, the \emph{clean-baseline drop}
\begin{equation}
    \dropbase(c) = \acc_{\mathrm{clean}} - \acc(c)
\end{equation}
measures how much performance is lost relative to the full clean input. For combined conditions, the \emph{worst-single drop}
\begin{equation}
    \dropsingle(c) = \acc_{\mathrm{worst\ single}}(c) - \acc(c)
\end{equation}
compares against the weakest corresponding single-modality corruption: positive $\dropsingle$ means the combined corruption is worse than every individual component (a compounded failure beyond the weakest single-modality condition), and negative $\dropsingle$ means the combined condition remains more accurate than its weakest single component.

\begin{figure*}[t]
\centering
\includegraphics[width=0.9\linewidth]{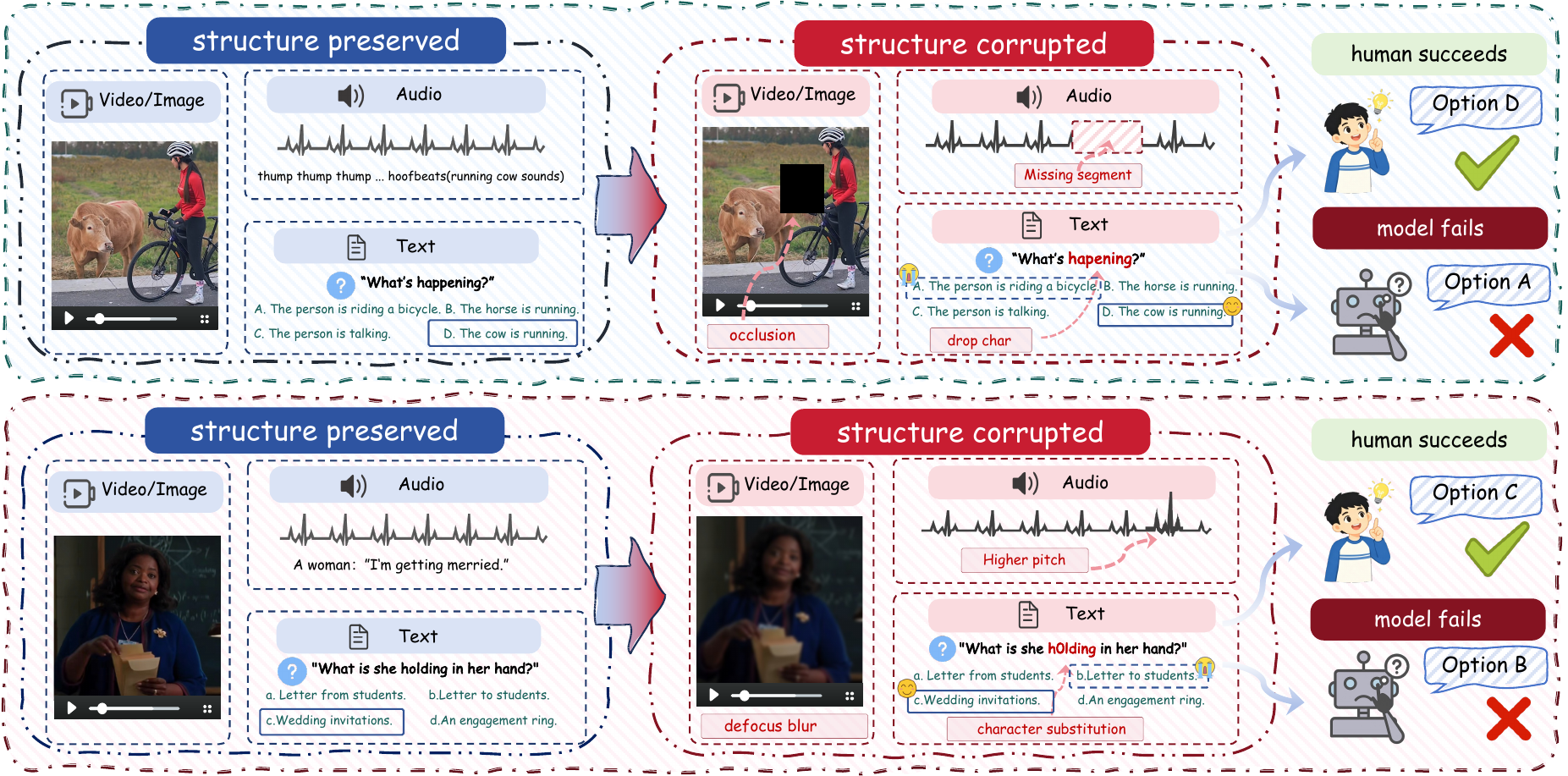}
\vspace{-10pt}
\caption{\method{} case illustration. One base example with its three clean modality channels (question/options, sampled frames, audio waveform) and one structurally corrupted variant per modality. Every variant keeps the gold answer, options, and modality channels fixed and damages only the internal evidence structure---distinguishing the fault-line probe from missing-modality ablations.}
\vspace{-4pt}
\label{fig:case}
\end{figure*}

\vspace{-6pt}
\paragraph{Auxiliary diagnostics.}
\label{sec:metrics-extra}
The headline matrix is complemented by five smaller quantities, each isolating a different mechanism:
\begin{itemize}[leftmargin=1.25em,topsep=2pt,itemsep=1pt]
    \item \textbf{Fault-line score} $\fault_S = \acc_{\mathrm{clean}} - \acc(c_S)$ for $S\in\{tv,\,ta,\,va,\,tva\}$, a model-level summary of how much accuracy a given multi-modal corruption block costs (reported per family in Table~\ref{tab:open-vs-prop-summary}).
    \item \textbf{Misleading-modality effect} $\mislead_m = \acc(\mathrm{missing}\ m) - \acc(\mathrm{corrupted}\ m)$, positive when a structurally corrupted but present modality is more harmful than removing it (Table~\ref{tab:missing-shortcut}).
    \item \textbf{Fragility slope} $\slope_m$ obtained by linearly regressing accuracy on severity $s\in\{0,10,30,50,70\}$ for each modality $m$ (clean encoded as $s{=}0$); see Table~\ref{tab:fragility-slope}.
    \item \textbf{Shortcut gap} $\shortcutgap = \acc(\mathrm{text\text{-}only}) - \acc(\mathrm{random})$, how much clean accuracy can be reproduced from the linguistic channel alone (Table~\ref{tab:missing-shortcut}, \textbf{Text-only} column).
    \item \textbf{Modality trust bias} $\trust_m$, how often a model follows modality $m$ under cross-modal mismatch (Appendix~\ref{app:mismatch-probes}).
\end{itemize}
We additionally report the invalid-output rate decomposed into API failures, parse failures, refusals, and empty outputs, since a model can fail either by choosing the wrong answer or by failing to produce a parseable answer under heavier structural stress.

\vspace{-6pt}
\paragraph{Statistical protocol.}
Every reported accuracy is accompanied by a $95\%$ confidence interval from $B{=}1000$ sample-level bootstrap~\citep{efron1992bootstrap} resamples over the $|I_c|$ original examples, applied after mean variant aggregation so that within-sample variant variance and between-sample variance are propagated into a single interval. Paired comparisons (clean vs.\ corrupted, single vs.\ combined, two models on the same condition) use the paired bootstrap on per-sample correctness differences and the paired McNemar test~\citep{mcnemar1947note} on the corresponding contingency table; both are reported with two-sided $p$-values and Holm--Bonferroni correction~\citep{holm1979simple} across the family of comparisons inside one experimental block. Invalid outputs are treated as incorrect for significance testing and as missing for accuracy estimation; we report both versions whenever the conclusion is sensitive to this choice.

\section{Experimental Results}
\label{sec:results}

We evaluate \method{} on a $14$-model panel of six proprietary/API systems and eight open or open-API systems (roster, interfaces, and the frame-extracted visual-input protocol in Appendix~\ref{app:expsetup}). All headline runs use one standardised multiple-choice JSON prompt (Appendix~\ref{app:prompts}); accuracy on each stochastic condition is the mean over up to three random variants per operator (\S\ref{sec:meanagg}), reported alongside coverage, invalid-output rates, and $95\%$ sample-level bootstrap intervals. Section~\ref{sec:single} reads the $14{\times}14$ severity-$70$ single-modality matrix and its per-model severity curves; Section~\ref{sec:joint} probes whether stacking corruption across two or three channels compounds the damage.

\subsection{Single-modality corruption}
\label{sec:single}

Table~\ref{tab:main-combined} reports clean accuracy together with the severity-$70$ drop on each of the fourteen single-modality operators for all fourteen systems. Two patterns dominate the matrix, and both turn out to be largely band-invariant.

\begin{table*}[t]
\centering
\scriptsize
\setlength{\tabcolsep}{2.7pt}
\renewcommand{\arraystretch}{1.08}
\resizebox{\textwidth}{!}{%
\begin{tabular}{@{}l c@{\hspace{6pt}} *{4}{c} @{\hspace{4pt}} *{7}{c} @{\hspace{4pt}} *{3}{c} @{}}
\toprule
\multicolumn{2}{c}{} & \multicolumn{4}{c}{\textbf{Text}} & \multicolumn{7}{c}{\textbf{Vision}} & \multicolumn{3}{c}{\textbf{Audio}} \\
\cmidrule(lr){3-6}\cmidrule(lr){7-13}\cmidrule(lr){14-16}
\textbf{Model} & \textbf{Clean} & Typo & Drop & Shuffle & Break & Noise & Occ. & LowRes & MBlur & DBlur & Expo. & Bright. & Remove & Mute & Distort \\
\midrule
\mainband{Proprietary / API omni-modal models}
\modelname{Gemini 3.1 Pro} & 84.15 & \maincell{79.85}{4.30} & \maincell{72.57}{11.58} & \maincell{75.54}{8.61} & \maincell{80.72}{3.43} & \maincell{73.57}{10.58} & \maincell{82.88}{1.27} & \maingain{85.07}{--0.92} & \maincell{82.29}{1.86} & \maincell{83.70}{0.45} & \maingain{85.08}{--0.93} & \maingain{84.88}{--0.73} & \maincell{78.89}{5.26} & \maincell{80.62}{3.53} & \maingain{86.17}{--2.02} \\
\modelname{Gemini 3 Pro} & 82.40 & \maincell{76.72}{5.68} & \maincell{71.30}{11.10} & \maincell{73.16}{9.24} & \maincell{80.27}{2.13} & \maincell{73.00}{9.40} & \maincell{80.09}{2.31} & \maingain{83.80}{--1.40} & \maincell{81.41}{0.99} & \maingain{82.89}{--0.49} & \maincell{80.49}{1.91} & \maingain{83.88}{--1.48} & \maincell{76.36}{6.04} & \maincell{75.82}{6.58} & \maingain{82.51}{--0.11} \\
\modelname{Gemini 3 Flash}   & 80.95 & \maincell{71.06}{9.89} & \maincell{67.40}{13.55} & \maincell{70.33}{10.62} & \maincell{72.89}{8.06} & \maincell{67.79}{13.16} & \maincell{80.33}{0.62} & \maingain{80.97}{0.02} & \maincell{77.59}{3.36} & \maincell{78.95}{2.00} & \maincell{79.37}{1.58} & \maincell{80.30}{0.65} & \maincell{74.30}{6.65} & \maincell{73.24}{7.71} & \maincell{79.17}{1.78} \\
\modelname{Gemini 3.5 Flash} & 78.39 & \maincell{72.53}{5.86} & \maincell{67.03}{11.36} & \maincell{68.50}{9.89} & \maincell{73.63}{4.76} & \maincell{62.75}{15.64} & \maincell{75.14}{3.25} & \maingain{78.96}{--0.57} & \maincell{75.56}{2.83} & \maincell{77.47}{0.92} & \maingain{79.30}{--0.91} & \maincell{78.31}{0.08} & \maincell{73.45}{4.94} & \maincell{71.00}{7.39} & \maincell{77.83}{0.56} \\
\modelname{Gemini 2.5 Pro} & 79.12 & \maincell{75.32}{3.80} & \maincell{67.41}{11.71} & \maincell{68.40}{10.72} & \maincell{71.77}{7.35} & \maincell{70.26}{8.86} & \maincell{76.97}{2.15} & \maingain{79.73}{--0.61} & \maincell{77.40}{1.72} & \maingain{80.30}{--1.18} & \maingain{79.44}{--0.32} & \maingain{79.19}{--0.07} & \maincell{74.64}{4.48} & \maincell{72.90}{6.22} & \maincell{77.81}{1.31} \\
\modelname{Gemini 2.5 Flash} & 75.40 & \maincell{67.31}{8.09} & \maincell{62.55}{12.85} & \maincell{65.99}{9.41} & \maincell{69.37}{6.03} & \maincell{63.55}{11.85} & \maincell{73.32}{2.08} & \maincell{73.46}{1.94} & \maincell{74.32}{1.08} & \maincell{73.48}{1.92} & \maincell{74.32}{1.08} & \maincell{72.89}{2.51} & \maincell{65.91}{9.49} & \maincell{65.86}{9.54} & \maincell{72.25}{3.15} \\
\midrule
\mainband{Open / open-API omni-modal models}
\modelname{Qwen3.5-Omni-Plus} & 73.63 & \maincell{69.60}{4.03} & \maincell{64.10}{9.53} & \maincell{65.20}{8.43} & \maincell{69.60}{4.03} & \maincell{70.37}{3.26} & \maincell{70.49}{3.14} & \maincell{71.79}{1.84} & \maincell{71.02}{2.61} & \maincell{71.83}{1.80} & \maincell{71.65}{1.98} & \maingain{73.68}{0.05} & \maincell{70.95}{2.68} & \maincell{69.72}{3.91} & \maingain{75.00}{1.37} \\
\modelname{Qwen3-Omni-30B} & 71.20 & \maincell{63.86}{7.34} & \maincell{61.22}{9.98} & \maincell{58.59}{12.61} & \maincell{64.89}{6.31} & \maincell{61.43}{9.77} & \maincell{66.13}{5.07} & \maincell{70.07}{1.13} & \maincell{67.67}{3.53} & \maincell{67.25}{3.95} & \maincell{67.37}{3.83} & \maincell{67.77}{3.43} & \maincell{65.32}{5.88} & \maincell{61.95}{9.25} & \maincell{70.49}{0.71} \\
\modelname{MiniCPM-o 4.5} & 74.50 & \maincell{69.02}{5.48} & \maincell{63.14}{11.36} & \maincell{61.38}{13.12} & \maincell{67.96}{6.54} & \maincell{65.41}{9.09} & \maincell{71.47}{3.03} & \maincell{73.33}{1.17} & \maincell{73.84}{0.66} & \maincell{74.38}{0.12} & \maingain{74.95}{--0.45} & \maincell{74.05}{0.45} & \maincell{65.58}{8.92} & \maincell{69.13}{5.37} & \maingain{74.93}{--0.43} \\
\modelname{Qwen2.5-Omni-7B} & 68.40 & \maincell{59.89}{8.51} & \maincell{56.79}{11.61} & \maincell{53.80}{14.60} & \maincell{61.18}{7.22} & \maincell{58.76}{9.64} & \maincell{65.75}{2.65} & \maincell{67.24}{1.16} & \maincell{65.89}{2.51} & \maincell{66.99}{1.41} & \maincell{65.17}{3.23} & \maincell{66.96}{1.44} & \maincell{58.55}{9.85} & \maincell{61.30}{7.10} & \maincell{68.34}{0.06} \\
\modelname{OmniVinci-9B} & 69.10 & \maincell{64.17}{4.93} & \maincell{57.65}{11.45} & \maincell{55.61}{13.49} & \maincell{62.76}{6.34} & \maincell{54.22}{14.88} & \maincell{66.55}{2.55} & \maincell{65.91}{3.19} & \maincell{67.84}{1.26} & \maincell{65.89}{3.21} & \maincell{65.56}{3.54} & \maincell{67.59}{1.51} & \maincell{61.00}{8.10} & \maincell{58.90}{10.20} & \maincell{67.25}{1.85} \\
\modelname{OLA-7B} & 66.80 & \maincell{61.50}{5.30} & \maincell{56.18}{10.62} & \maincell{55.73}{11.07} & \maincell{59.75}{7.05} & \maincell{54.58}{12.22} & \maincell{63.34}{3.46} & \maincell{65.07}{1.73} & \maincell{62.65}{4.15} & \maincell{65.99}{0.81} & \maincell{62.88}{3.92} & \maincell{62.79}{4.01} & \maincell{57.45}{9.35} & \maincell{57.81}{8.99} & \maincell{66.29}{0.51} \\
\modelname{Qwen2.5-Omni-3B} & 64.20 & \maincell{55.31}{8.89} & \maincell{53.80}{10.40} & \maincell{51.01}{13.19} & \maincell{56.49}{7.71} & \maincell{48.75}{15.45} & \maincell{60.61}{3.59} & \maincell{62.98}{1.22} & \maincell{61.76}{2.44} & \maincell{62.29}{1.91} & \maincell{62.61}{1.59} & \maincell{62.38}{1.82} & \maincell{55.83}{8.37} & \maincell{55.09}{9.11} & \maincell{62.68}{1.52} \\
\modelname{Baichuan-Omni-1.5} & 65.50 & \maincell{56.60}{8.90} & \maincell{54.73}{10.77} & \maincell{54.09}{11.41} & \maincell{56.71}{8.79} & \maincell{49.80}{15.70} & \maincell{59.21}{6.29} & \maincell{63.52}{1.98} & \maincell{61.56}{3.94} & \maincell{65.03}{0.47} & \maincell{64.34}{1.16} & \maincell{61.55}{3.95} & \maincell{54.69}{10.81} & \maincell{55.93}{9.57} & \maincell{63.99}{1.51} \\
\bottomrule
\end{tabular}%
}
\vspace{-10pt}
\caption{Severity-$70$ single-modality corruption matrix on the $14$-model panel (clean accuracy in the second column; subsequent columns report accuracy under each operator with the absolute drop from clean as a subscript, $\downarrow$\,drop / $\uparrow$\,gain). Columns are grouped by modality (4 text, 7 vision, 3 audio). The four operators with the steepest panel-mean drops (\texttt{drop\_words}, \texttt{word\_shuffle}, \texttt{noise}, \texttt{mute}/\texttt{remove}) define the joint-corruption building blocks of \S\ref{sec:joint}. Table~\ref{tab:coverage-aware-summary} reports the coverage-aware operator grouping.}
\vspace{-16pt}
\label{tab:main-combined}
\end{table*}

Fragility concentrates in a small set of \emph{structural} operators. Five of the fourteen sit clearly above the rest at $\geq\!7$\,pp panel-mean drop---\opname{drop\_words}, \opname{word\_shuffle}, \opname{noise}, and the two audio-segment operators \opname{remove}/\opname{mute}---each targeting word identity, dense visual signal, or audio presence. The remaining operators are consistently weaker; three of them (\opname{low\_resolution}, \opname{defocus\_blur}, \opname{distortion}) hover near zero or flip into small apparent gains, suggesting that cosmetic perturbations rarely reach the model's evidence-assembly path. The two single hardest cells in the matrix---\opname{noise} on \modelname{Baichuan-Omni-1.5} ($-15.70$\,pp) and \opname{word\_shuffle} on \modelname{Qwen2.5-Omni-7B} ($-14.60$\,pp)---both approach or exceed the worst proprietary cell ($-15.64$\,pp on \modelname{Gemini 3.5 Flash}), so the hard core is not a proprietary--open story.

The two bands separate cleanly on clean accuracy ($75$--$84$ vs.\ $64$--$74$), but their relative robustness to those hard operators is similar: \opname{noise} averages $11.6$\,pp drop on the six proprietary systems and $11.3$\,pp on the eight open systems, and the largest proprietary model (\modelname{Gemini 3.1 Pro}) and the smallest open model (\modelname{Qwen2.5-Omni-3B}) both lose $8$--$14$\,pp on the two hardest text operators. The proprietary advantage instead concentrates in the soft tail: \modelname{Gemini 3.1 Pro} posts apparent gains on five visual operators and on \opname{distortion}, while \modelname{Qwen2.5-Omni-3B} and \modelname{Baichuan-Omni-1.5} produce no gain cells anywhere. \modelname{Qwen3.5-Omni-Plus} is the one cross-band outlier (only $-3.26$\,pp on \opname{noise}), placing it closer to the proprietary band on visual robustness than to its open-tier neighbours.

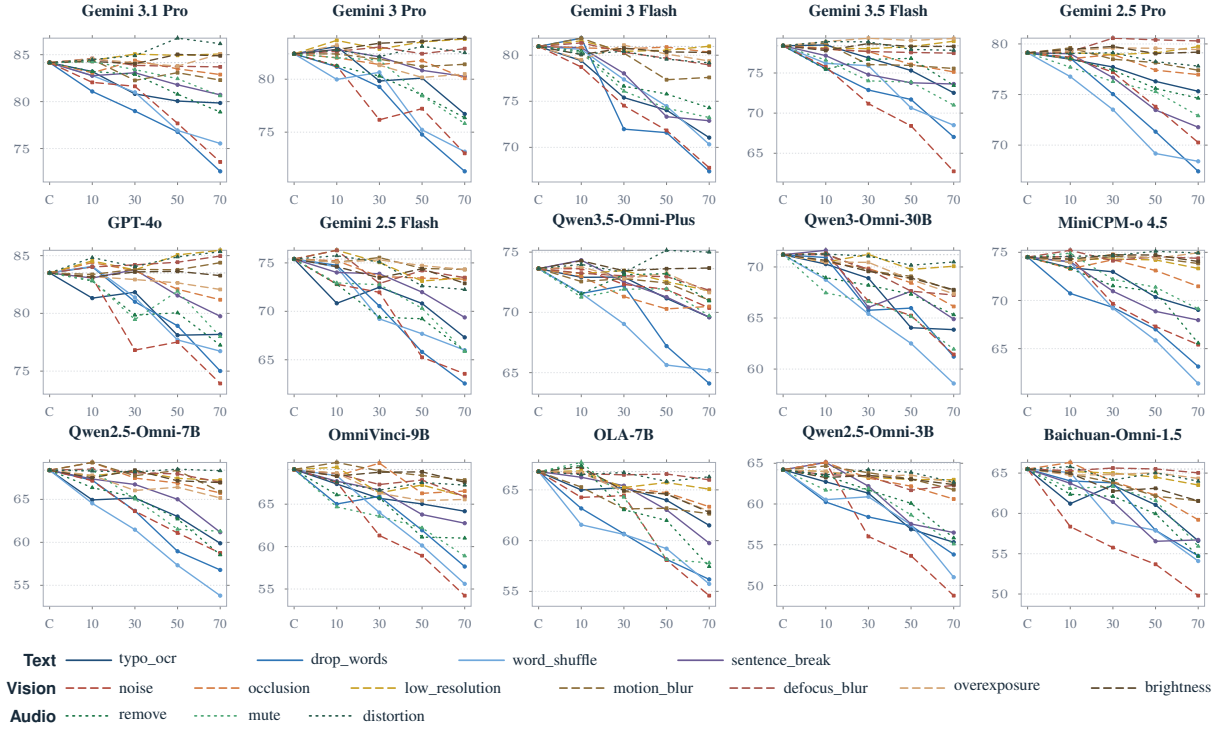
\begin{figure*}[t]
\centering
\begingroup
% Text family: blue family
\definecolor{cTYP}{HTML}{1F4E79}
\definecolor{cDRP}{HTML}{2E75B6}
\definecolor{cSHF}{HTML}{6FA8DC}
\definecolor{cBRK}{HTML}{6B5B95}
% Vision family: warm earth tones (red/orange/gold/brown)
\definecolor{cNOI}{HTML}{B44C3F}
\definecolor{cOCC}{HTML}{D67D44}
\definecolor{cLRS}{HTML}{C9A227}
\definecolor{cMBL}{HTML}{8C6E3A}
\definecolor{cDBL}{HTML}{A8554D}
\definecolor{cOVE}{HTML}{D4A574}
\definecolor{cBRT}{HTML}{5E4A2E}
% Audio family: green family
\definecolor{cREM}{HTML}{1E7A4A}
\definecolor{cMUT}{HTML}{4FA877}
\definecolor{cDST}{HTML}{2A5F4F}

\pgfplotsset{
  fmpanel/.style={
    width=2.55cm, height=2.0cm, scale only axis,
    xmin=-0.15, xmax=4.15,
    % Per-panel auto-scaled y axis: each model's accuracy range differs
    % by 20+ pp across the 14 systems (Qwen2.5-Omni-3B floors near 50,
    % Gemini 3.1 Pro ceils near 86), so a shared y window would clip the
    % curves of low-baseline panels and waste space on high-baseline ones.
    % We let pgfplots compute ymin/ymax from each panel's data, then pad
    % by 8% on top and bottom and emit three auto-positioned y ticks.
    enlarge y limits={value=0.08, upper, lower},
    ymajorgrids=true, xmajorgrids=false,
    max space between ticks=18pt,
    try min ticks=3,
    yticklabel style={/pgf/number format/precision=0, /pgf/number format/fixed},
    xtick={0,1,2,3,4}, xticklabels={C,10,30,50,70},
    tick align=outside,
    tick label style={font=\fontsize{5.2pt}{6pt}\selectfont,text=FaultSlate!90},
    title style={font=\fontsize{6.5pt}{7pt}\selectfont\bfseries,text=FaultNavy,yshift=-1pt},
    axis line style={color=FaultSlate!55,line width=0.28pt},
    every axis plot/.append style={line width=0.55pt,line cap=round,line join=round,mark size=0.65pt,mark options={solid,line width=0.25pt}},
    grid style={color=FaultSlate!22,line width=0.16pt,dash pattern=on 0.5pt off 0.7pt},
    major tick length=1.2pt,
    enlarge x limits=false,
    clip mode=individual,
  },
  text op/.style={solid, mark=*},
  vis op/.style={densely dashed, mark=square*, mark size=0.55pt},
  aud op/.style={dotted, mark=triangle*, mark size=0.75pt},
}
\tikzset{pendingnode/.style={font=\fontsize{5.5pt}{6pt}\selectfont\itshape,text=FaultSlate!85}}

\newcommand{\noresult}{\node[pendingnode] at (axis cs:2,72) {result pending};}
\newcommand{\baseref}[1]{\draw[FaultSlate!35,line width=0.22pt,dash pattern=on 0.6pt off 0.8pt] (axis cs:-0.15,#1) -- (axis cs:4.15,#1);}

\resizebox{\textwidth}{!}{%
\begin{tikzpicture}
\begin{groupplot}[group style={group size=5 by 3, horizontal sep=0.85cm, vertical sep=0.95cm}, fmpanel]

\nextgroupplot[title={Gemini 3.1 Pro}]
\baseref{84.15}
\addplot[cTYP,text op] coordinates {(0,84.15) (1,83.20) (2,80.82) (3,80.06) (4,79.85)};
\addplot[cDRP,text op] coordinates {(0,84.15) (1,81.08) (2,78.99) (3,76.77) (4,72.57)};
\addplot[cSHF,text op] coordinates {(0,84.15) (1,82.78) (2,80.99) (3,76.93) (4,75.54)};
\addplot[cBRK,text op] coordinates {(0,84.15) (1,82.76) (2,83.07) (3,81.79) (4,80.72)};
\addplot[cNOI,vis op] coordinates {(0,84.15) (1,82.05) (2,81.63) (3,77.68) (4,73.57)};
\addplot[cOCC,vis op] coordinates {(0,84.15) (1,83.07) (2,84.38) (3,83.49) (4,82.88)};
\addplot[cLRS,vis op] coordinates {(0,84.15) (1,84.39) (2,85.06) (3,84.93) (4,85.07)};
\addplot[cMBL,vis op] coordinates {(0,84.15) (1,84.49) (2,82.26) (3,83.08) (4,82.29)};
\addplot[cDBL,vis op] coordinates {(0,84.15) (1,84.60) (2,83.86) (3,83.76) (4,83.70)};
\addplot[cOVE,vis op] coordinates {(0,84.15) (1,84.55) (2,84.14) (3,83.78) (4,85.08)};
\addplot[cBRT,vis op] coordinates {(0,84.15) (1,84.28) (2,84.04) (3,85.03) (4,84.88)};
\addplot[cREM,aud op] coordinates {(0,84.15) (1,83.21) (2,82.90) (3,80.79) (4,78.89)};
\addplot[cMUT,aud op] coordinates {(0,84.15) (1,84.26) (2,83.41) (3,82.44) (4,80.62)};
\addplot[cDST,aud op] coordinates {(0,84.15) (1,84.48) (2,84.89) (3,86.76) (4,86.17)};
\nextgroupplot[title={Gemini 3 Pro}]
\baseref{82.40}
\addplot[cTYP,text op] coordinates {(0,82.40) (1,83.09) (2,79.83) (3,80.11) (4,76.72)};
\addplot[cDRP,text op] coordinates {(0,82.40) (1,81.17) (2,79.28) (3,74.78) (4,71.30)};
\addplot[cSHF,text op] coordinates {(0,82.40) (1,79.98) (2,80.65) (3,75.22) (4,73.16)};
\addplot[cBRK,text op] coordinates {(0,82.40) (1,82.81) (2,82.12) (3,80.84) (4,80.27)};
\addplot[cNOI,vis op] coordinates {(0,82.40) (1,81.25) (2,76.17) (3,77.22) (4,73.00)};
\addplot[cOCC,vis op] coordinates {(0,82.40) (1,82.03) (2,81.39) (3,81.76) (4,80.09)};
\addplot[cLRS,vis op] coordinates {(0,82.40) (1,83.67) (2,82.86) (3,83.52) (4,83.80)};
\addplot[cMBL,vis op] coordinates {(0,82.40) (1,82.39) (2,81.90) (3,81.20) (4,81.41)};
\addplot[cDBL,vis op] coordinates {(0,82.40) (1,82.63) (2,83.04) (3,82.41) (4,82.89)};
\addplot[cOVE,vis op] coordinates {(0,82.40) (1,83.22) (2,81.32) (3,80.17) (4,80.49)};
\addplot[cBRT,vis op] coordinates {(0,82.40) (1,82.79) (2,83.40) (3,83.57) (4,83.88)};
\addplot[cREM,aud op] coordinates {(0,82.40) (1,81.27) (2,80.30) (3,78.51) (4,76.36)};
\addplot[cMUT,aud op] coordinates {(0,82.40) (1,82.03) (2,81.81) (3,78.43) (4,75.82)};
\addplot[cDST,aud op] coordinates {(0,82.40) (1,82.76) (2,82.13) (3,83.11) (4,82.51)};
\nextgroupplot[title={Gemini 3 Flash}]
\baseref{80.95}
\addplot[cTYP,text op] coordinates {(0,80.95) (1,79.40) (2,75.40) (3,74.04) (4,71.06)};
\addplot[cDRP,text op] coordinates {(0,80.95) (1,81.85) (2,71.98) (3,71.60) (4,67.40)};
\addplot[cSHF,text op] coordinates {(0,80.95) (1,80.64) (2,77.37) (3,74.47) (4,70.33)};
\addplot[cBRK,text op] coordinates {(0,80.95) (1,80.82) (2,78.03) (3,73.33) (4,72.89)};
\addplot[cNOI,vis op] coordinates {(0,80.95) (1,78.72) (2,74.53) (3,71.84) (4,67.79)};
\addplot[cOCC,vis op] coordinates {(0,80.95) (1,80.80) (2,80.68) (3,80.88) (4,80.33)};
\addplot[cLRS,vis op] coordinates {(0,80.95) (1,81.57) (2,80.42) (3,80.54) (4,80.97)};
\addplot[cMBL,vis op] coordinates {(0,80.95) (1,81.85) (2,80.23) (3,77.34) (4,77.59)};
\addplot[cDBL,vis op] coordinates {(0,80.95) (1,81.34) (2,80.32) (3,79.67) (4,78.95)};
\addplot[cOVE,vis op] coordinates {(0,80.95) (1,79.50) (2,81.09) (3,80.13) (4,79.37)};
\addplot[cBRT,vis op] coordinates {(0,80.95) (1,80.07) (2,80.77) (3,80.36) (4,80.30)};
\addplot[cREM,aud op] coordinates {(0,80.95) (1,80.45) (2,76.64) (3,75.79) (4,74.30)};
\addplot[cMUT,aud op] coordinates {(0,80.95) (1,80.31) (2,76.10) (3,74.25) (4,73.24)};
\addplot[cDST,aud op] coordinates {(0,80.95) (1,80.11) (2,80.42) (3,79.57) (4,79.17)};
\nextgroupplot[title={Gemini 3.5 Flash}]
\baseref{78.39}
\addplot[cTYP,text op] coordinates {(0,78.39) (1,78.49) (2,76.89) (3,75.31) (4,72.53)};
\addplot[cDRP,text op] coordinates {(0,78.39) (1,75.52) (2,72.88) (3,71.70) (4,67.03)};
\addplot[cSHF,text op] coordinates {(0,78.39) (1,76.20) (2,75.91) (3,70.65) (4,68.50)};
\addplot[cBRK,text op] coordinates {(0,78.39) (1,77.16) (2,74.80) (3,73.78) (4,73.63)};
\addplot[cNOI,vis op] coordinates {(0,78.39) (1,75.77) (2,71.18) (3,68.41) (4,62.75)};
\addplot[cOCC,vis op] coordinates {(0,78.39) (1,77.97) (2,77.68) (3,76.07) (4,75.14)};
\addplot[cLRS,vis op] coordinates {(0,78.39) (1,78.11) (2,78.03) (3,78.28) (4,78.96)};
\addplot[cMBL,vis op] coordinates {(0,78.39) (1,78.70) (2,76.05) (3,75.95) (4,75.56)};
\addplot[cDBL,vis op] coordinates {(0,78.39) (1,78.00) (2,77.67) (3,77.56) (4,77.47)};
\addplot[cOVE,vis op] coordinates {(0,78.39) (1,78.91) (2,79.34) (3,79.05) (4,79.30)};
\addplot[cBRT,vis op] coordinates {(0,78.39) (1,77.80) (2,78.67) (3,78.31) (4,78.31)};
\addplot[cREM,aud op] coordinates {(0,78.39) (1,75.47) (2,76.69) (3,76.82) (4,73.45)};
\addplot[cMUT,aud op] coordinates {(0,78.39) (1,76.61) (2,74.02) (3,73.87) (4,71.00)};
\addplot[cDST,aud op] coordinates {(0,78.39) (1,78.91) (2,78.88) (3,77.93) (4,77.83)};
\nextgroupplot[title={Gemini 2.5 Pro}]
\baseref{79.12}
\addplot[cTYP,text op] coordinates {(0,79.12) (1,78.49) (2,77.72) (3,76.30) (4,75.32)};
\addplot[cDRP,text op] coordinates {(0,79.12) (1,78.60) (2,75.04) (3,71.33) (4,67.41)};
\addplot[cSHF,text op] coordinates {(0,79.12) (1,76.77) (2,73.51) (3,69.17) (4,68.40)};
\addplot[cBRK,text op] coordinates {(0,79.12) (1,79.03) (2,76.67) (3,73.47) (4,71.77)};
\addplot[cNOI,vis op] coordinates {(0,79.12) (1,78.69) (2,77.21) (3,73.80) (4,70.26)};
\addplot[cOCC,vis op] coordinates {(0,79.12) (1,78.53) (2,78.96) (3,77.41) (4,76.97)};
\addplot[cLRS,vis op] coordinates {(0,79.12) (1,79.14) (2,78.95) (3,79.11) (4,79.73)};
\addplot[cMBL,vis op] coordinates {(0,79.12) (1,79.61) (2,78.49) (3,78.15) (4,77.40)};
\addplot[cDBL,vis op] coordinates {(0,79.12) (1,79.22) (2,80.57) (3,80.40) (4,80.30)};
\addplot[cOVE,vis op] coordinates {(0,79.12) (1,79.32) (2,79.61) (3,79.57) (4,79.44)};
\addplot[cBRT,vis op] coordinates {(0,79.12) (1,79.47) (2,79.75) (3,79.05) (4,79.19)};
\addplot[cREM,aud op] coordinates {(0,79.12) (1,78.56) (2,77.42) (3,75.60) (4,74.64)};
\addplot[cMUT,aud op] coordinates {(0,79.12) (1,77.70) (2,76.28) (3,75.29) (4,72.90)};
\addplot[cDST,aud op] coordinates {(0,79.12) (1,79.05) (2,79.13) (3,78.25) (4,77.81)};

\nextgroupplot[title={Gemini 2.5 Flash}]
\baseref{75.40}
\addplot[cTYP,text op] coordinates {(0,75.40) (1,70.83) (2,72.49) (3,70.81) (4,67.31)};
\addplot[cDRP,text op] coordinates {(0,75.40) (1,74.75) (2,70.54) (3,65.81) (4,62.55)};
\addplot[cSHF,text op] coordinates {(0,75.40) (1,74.58) (2,69.22) (3,67.68) (4,65.99)};
\addplot[cBRK,text op] coordinates {(0,75.40) (1,74.01) (2,73.88) (3,71.98) (4,69.37)};
\addplot[cNOI,vis op] coordinates {(0,75.40) (1,72.80) (2,71.96) (3,65.25) (4,63.55)};
\addplot[cOCC,vis op] coordinates {(0,75.40) (1,75.22) (2,73.60) (3,73.43) (4,73.32)};
\addplot[cLRS,vis op] coordinates {(0,75.40) (1,76.30) (2,75.06) (3,73.13) (4,73.46)};
\addplot[cMBL,vis op] coordinates {(0,75.40) (1,75.07) (2,75.55) (3,74.43) (4,74.32)};
\addplot[cDBL,vis op] coordinates {(0,75.40) (1,76.30) (2,72.88) (3,74.21) (4,73.48)};
\addplot[cOVE,vis op] coordinates {(0,75.40) (1,75.07) (2,75.31) (3,74.72) (4,74.32)};
\addplot[cBRT,vis op] coordinates {(0,75.40) (1,74.61) (2,73.46) (3,74.39) (4,72.89)};
\addplot[cREM,aud op] coordinates {(0,75.40) (1,72.81) (2,69.40) (3,69.23) (4,65.91)};
\addplot[cMUT,aud op] coordinates {(0,75.40) (1,72.93) (2,72.75) (3,70.31) (4,65.86)};
\addplot[cDST,aud op] coordinates {(0,75.40) (1,75.74) (2,75.09) (3,72.62) (4,72.25)};
\nextgroupplot[title={Qwen3.5-Omni-Plus}]
\baseref{73.63}
\addplot[cTYP,text op] coordinates {(0,73.63) (1,72.91) (2,72.93) (3,71.16) (4,69.60)};
\addplot[cDRP,text op] coordinates {(0,73.63) (1,71.57) (2,72.21) (3,67.20) (4,64.10)};
\addplot[cSHF,text op] coordinates {(0,73.63) (1,71.58) (2,69.04) (3,65.63) (4,65.20)};
\addplot[cBRK,text op] coordinates {(0,73.63) (1,74.33) (2,72.54) (3,71.27) (4,69.60)};
\addplot[cNOI,vis op] coordinates {(0,73.63) (1,73.62) (2,72.32) (3,71.91) (4,70.37)};
\addplot[cOCC,vis op] coordinates {(0,73.63) (1,72.99) (2,71.31) (3,70.29) (4,70.49)};
\addplot[cLRS,vis op] coordinates {(0,73.63) (1,73.21) (2,73.47) (3,72.55) (4,71.79)};
\addplot[cMBL,vis op] coordinates {(0,73.63) (1,72.56) (2,72.80) (3,72.44) (4,71.02)};
\addplot[cDBL,vis op] coordinates {(0,73.63) (1,73.25) (2,73.03) (3,72.97) (4,71.83)};
\addplot[cOVE,vis op] coordinates {(0,73.63) (1,73.78) (2,72.78) (3,73.25) (4,71.65)};
\addplot[cBRT,vis op] coordinates {(0,73.63) (1,74.28) (2,73.45) (3,73.62) (4,73.68)};
\addplot[cREM,aud op] coordinates {(0,73.63) (1,71.53) (2,73.13) (3,73.21) (4,70.95)};
\addplot[cMUT,aud op] coordinates {(0,73.63) (1,71.26) (2,71.92) (3,71.96) (4,69.72)};
\addplot[cDST,aud op] coordinates {(0,73.63) (1,73.98) (2,73.31) (3,75.15) (4,75.00)};
\nextgroupplot[title={Qwen3-Omni-30B}]
\baseref{71.20}
\addplot[cTYP,text op] coordinates {(0,71.20) (1,70.28) (2,68.90) (3,64.05) (4,63.86)};
\addplot[cDRP,text op] coordinates {(0,71.20) (1,70.92) (2,65.76) (3,65.97) (4,61.22)};
\addplot[cSHF,text op] coordinates {(0,71.20) (1,68.75) (2,65.41) (3,62.51) (4,58.59)};
\addplot[cBRK,text op] coordinates {(0,71.20) (1,71.63) (2,66.04) (3,67.63) (4,64.89)};
\addplot[cNOI,vis op] coordinates {(0,71.20) (1,71.33) (2,66.66) (3,65.24) (4,61.43)};
\addplot[cOCC,vis op] coordinates {(0,71.20) (1,70.45) (2,69.85) (3,68.45) (4,66.13)};
\addplot[cLRS,vis op] coordinates {(0,71.20) (1,70.48) (2,71.23) (3,69.76) (4,70.07)};
\addplot[cMBL,vis op] coordinates {(0,71.20) (1,71.30) (2,69.63) (3,69.08) (4,67.67)};
\addplot[cDBL,vis op] coordinates {(0,71.20) (1,71.20) (2,69.77) (3,67.63) (4,67.25)};
\addplot[cOVE,vis op] coordinates {(0,71.20) (1,70.48) (2,70.45) (3,68.76) (4,67.37)};
\addplot[cBRT,vis op] coordinates {(0,71.20) (1,70.64) (2,69.57) (3,68.91) (4,67.77)};
\addplot[cREM,aud op] coordinates {(0,71.20) (1,68.95) (2,68.21) (3,67.33) (4,65.32)};
\addplot[cMUT,aud op] coordinates {(0,71.20) (1,67.46) (2,66.62) (3,65.15) (4,61.95)};
\addplot[cDST,aud op] coordinates {(0,71.20) (1,71.26) (2,71.06) (3,70.17) (4,70.49)};
\nextgroupplot[title={MiniCPM-o 4.5}]
\baseref{74.50}
\addplot[cTYP,text op] coordinates {(0,74.50) (1,73.40) (2,72.98) (3,70.35) (4,69.02)};
\addplot[cDRP,text op] coordinates {(0,74.50) (1,70.73) (2,69.28) (3,67.00) (4,63.14)};
\addplot[cSHF,text op] coordinates {(0,74.50) (1,73.98) (2,69.19) (3,65.85) (4,61.38)};
\addplot[cBRK,text op] coordinates {(0,74.50) (1,74.03) (2,70.96) (3,68.87) (4,67.96)};
\addplot[cNOI,vis op] coordinates {(0,74.50) (1,74.55) (2,69.66) (3,67.30) (4,65.41)};
\addplot[cOCC,vis op] coordinates {(0,74.50) (1,73.28) (2,74.23) (3,73.11) (4,71.47)};
\addplot[cLRS,vis op] coordinates {(0,74.50) (1,74.29) (2,74.31) (3,74.21) (4,73.33)};
\addplot[cMBL,vis op] coordinates {(0,74.50) (1,73.97) (2,74.58) (3,74.41) (4,73.84)};
\addplot[cDBL,vis op] coordinates {(0,74.50) (1,75.21) (2,74.13) (3,74.55) (4,74.38)};
\addplot[cOVE,vis op] coordinates {(0,74.50) (1,73.99) (2,74.81) (3,74.38) (4,74.95)};
\addplot[cBRT,vis op] coordinates {(0,74.50) (1,74.29) (2,74.74) (3,74.72) (4,74.05)};
\addplot[cREM,aud op] coordinates {(0,74.50) (1,73.30) (2,71.52) (3,70.91) (4,65.58)};
\addplot[cMUT,aud op] coordinates {(0,74.50) (1,74.91) (2,72.19) (3,71.38) (4,69.13)};
\addplot[cDST,aud op] coordinates {(0,74.50) (1,74.59) (2,74.66) (3,75.11) (4,74.93)};

\nextgroupplot[title={Qwen2.5-Omni-7B}]
\baseref{68.40}
\addplot[cTYP,text op] coordinates {(0,68.40) (1,64.94) (2,65.16) (3,63.01) (4,59.89)};
\addplot[cDRP,text op] coordinates {(0,68.40) (1,67.23) (2,63.67) (3,58.96) (4,56.79)};
\addplot[cSHF,text op] coordinates {(0,68.40) (1,64.54) (2,61.48) (3,57.33) (4,53.80)};
\addplot[cBRK,text op] coordinates {(0,68.40) (1,67.37) (2,66.74) (3,65.01) (4,61.18)};
\addplot[cNOI,vis op] coordinates {(0,68.40) (1,67.18) (2,63.63) (3,61.10) (4,58.76)};
\addplot[cOCC,vis op] coordinates {(0,68.40) (1,69.30) (2,67.48) (3,66.90) (4,65.75)};
\addplot[cLRS,vis op] coordinates {(0,68.40) (1,67.73) (2,68.30) (3,67.45) (4,67.24)};
\addplot[cMBL,vis op] coordinates {(0,68.40) (1,69.30) (2,67.76) (3,68.37) (4,65.89)};
\addplot[cDBL,vis op] coordinates {(0,68.40) (1,68.53) (2,68.10) (3,67.97) (4,66.99)};
\addplot[cOVE,vis op] coordinates {(0,68.40) (1,67.80) (2,66.04) (3,66.42) (4,65.17)};
\addplot[cBRT,vis op] coordinates {(0,68.40) (1,67.52) (2,68.40) (3,67.17) (4,66.96)};
\addplot[cREM,aud op] coordinates {(0,68.40) (1,66.37) (2,65.26) (3,62.72) (4,58.55)};
\addplot[cMUT,aud op] coordinates {(0,68.40) (1,68.40) (2,65.05) (3,61.53) (4,61.30)};
\addplot[cDST,aud op] coordinates {(0,68.40) (1,68.30) (2,68.19) (3,68.52) (4,68.34)};
\nextgroupplot[title={OmniVinci-9B}]
\baseref{69.10}
\addplot[cTYP,text op] coordinates {(0,69.10) (1,67.38) (2,65.69) (3,65.03) (4,64.17)};
\addplot[cDRP,text op] coordinates {(0,69.10) (1,65.02) (2,65.92) (3,61.93) (4,57.65)};
\addplot[cSHF,text op] coordinates {(0,69.10) (1,68.35) (2,64.02) (3,60.13) (4,55.61)};
\addplot[cBRK,text op] coordinates {(0,69.10) (1,67.71) (2,66.68) (3,63.78) (4,62.76)};
\addplot[cNOI,vis op] coordinates {(0,69.10) (1,67.57) (2,61.31) (3,58.93) (4,54.22)};
\addplot[cOCC,vis op] coordinates {(0,69.10) (1,68.30) (2,69.84) (3,66.28) (4,66.55)};
\addplot[cLRS,vis op] coordinates {(0,69.10) (1,69.35) (2,66.34) (3,67.26) (4,65.91)};
\addplot[cMBL,vis op] coordinates {(0,69.10) (1,69.92) (2,68.93) (3,68.38) (4,67.84)};
\addplot[cDBL,vis op] coordinates {(0,69.10) (1,68.59) (2,67.32) (3,67.85) (4,65.89)};
\addplot[cOVE,vis op] coordinates {(0,69.10) (1,68.67) (2,66.30) (3,65.42) (4,65.56)};
\addplot[cBRT,vis op] coordinates {(0,69.10) (1,68.31) (2,68.77) (3,68.81) (4,67.59)};
\addplot[cREM,aud op] coordinates {(0,69.10) (1,66.10) (2,65.76) (3,61.12) (4,61.00)};
\addplot[cMUT,aud op] coordinates {(0,69.10) (1,64.68) (2,63.58) (3,62.20) (4,58.90)};
\addplot[cDST,aud op] coordinates {(0,69.10) (1,67.60) (2,66.65) (3,67.56) (4,67.25)};
\nextgroupplot[title={OLA-7B}]
\baseref{66.80}
\addplot[cTYP,text op] coordinates {(0,66.80) (1,64.99) (2,65.22) (3,63.98) (4,61.50)};
\addplot[cDRP,text op] coordinates {(0,66.80) (1,63.17) (2,60.67) (3,58.15) (4,56.18)};
\addplot[cSHF,text op] coordinates {(0,66.80) (1,61.56) (2,60.61) (3,59.21) (4,55.73)};
\addplot[cBRK,text op] coordinates {(0,66.80) (1,66.24) (2,65.41) (3,63.03) (4,59.75)};
\addplot[cNOI,vis op] coordinates {(0,66.80) (1,64.26) (2,64.43) (3,58.08) (4,54.58)};
\addplot[cOCC,vis op] coordinates {(0,66.80) (1,66.65) (2,65.20) (3,64.72) (4,63.34)};
\addplot[cLRS,vis op] coordinates {(0,66.80) (1,66.87) (2,65.23) (3,65.72) (4,65.07)};
\addplot[cMBL,vis op] coordinates {(0,66.80) (1,65.27) (2,63.12) (3,63.19) (4,62.65)};
\addplot[cDBL,vis op] coordinates {(0,66.80) (1,66.60) (2,66.48) (3,66.56) (4,65.99)};
\addplot[cOVE,vis op] coordinates {(0,66.80) (1,66.72) (2,64.81) (3,64.63) (4,62.88)};
\addplot[cBRT,vis op] coordinates {(0,66.80) (1,67.23) (2,64.91) (3,64.59) (4,62.79)};
\addplot[cREM,aud op] coordinates {(0,66.80) (1,67.41) (2,63.04) (3,61.98) (4,57.45)};
\addplot[cMUT,aud op] coordinates {(0,66.80) (1,67.70) (2,64.28) (3,58.14) (4,57.81)};
\addplot[cDST,aud op] coordinates {(0,66.80) (1,66.51) (2,66.71) (3,65.83) (4,66.29)};
\nextgroupplot[title={Qwen2.5-Omni-3B}]
\baseref{64.20}
\addplot[cTYP,text op] coordinates {(0,64.20) (1,62.72) (2,61.33) (3,56.89) (4,55.31)};
\addplot[cDRP,text op] coordinates {(0,64.20) (1,60.25) (2,58.41) (3,57.33) (4,53.80)};
\addplot[cSHF,text op] coordinates {(0,64.20) (1,60.53) (2,60.89) (3,57.44) (4,51.01)};
\addplot[cBRK,text op] coordinates {(0,64.20) (1,65.10) (2,62.19) (3,57.55) (4,56.49)};
\addplot[cNOI,vis op] coordinates {(0,64.20) (1,65.10) (2,56.01) (3,53.64) (4,48.75)};
\addplot[cOCC,vis op] coordinates {(0,64.20) (1,65.10) (2,63.18) (3,62.18) (4,60.61)};
\addplot[cLRS,vis op] coordinates {(0,64.20) (1,63.98) (2,63.27) (3,63.01) (4,62.98)};
\addplot[cMBL,vis op] coordinates {(0,64.20) (1,64.67) (2,63.84) (3,63.07) (4,61.76)};
\addplot[cDBL,vis op] coordinates {(0,64.20) (1,63.61) (2,63.42) (3,61.69) (4,62.29)};
\addplot[cOVE,vis op] coordinates {(0,64.20) (1,63.98) (2,63.40) (3,63.55) (4,62.61)};
\addplot[cBRT,vis op] coordinates {(0,64.20) (1,63.25) (2,63.51) (3,63.04) (4,62.38)};
\addplot[cREM,aud op] coordinates {(0,64.20) (1,63.47) (2,61.72) (3,60.08) (4,55.83)};
\addplot[cMUT,aud op] coordinates {(0,64.20) (1,61.70) (2,61.90) (3,58.65) (4,55.09)};
\addplot[cDST,aud op] coordinates {(0,64.20) (1,63.57) (2,64.22) (3,63.89) (4,62.68)};
\nextgroupplot[title={Baichuan-Omni-1.5}]
\baseref{65.50}
\addplot[cTYP,text op] coordinates {(0,65.50) (1,61.21) (2,63.50) (3,61.06) (4,56.60)};
\addplot[cDRP,text op] coordinates {(0,65.50) (1,64.03) (2,63.79) (3,57.90) (4,54.73)};
\addplot[cSHF,text op] coordinates {(0,65.50) (1,65.07) (2,58.91) (3,57.90) (4,54.09)};
\addplot[cBRK,text op] coordinates {(0,65.50) (1,63.73) (2,61.43) (3,56.54) (4,56.71)};
\addplot[cNOI,vis op] coordinates {(0,65.50) (1,58.36) (2,55.76) (3,53.69) (4,49.80)};
\addplot[cOCC,vis op] coordinates {(0,65.50) (1,66.32) (2,63.95) (3,62.26) (4,59.21)};
\addplot[cLRS,vis op] coordinates {(0,65.50) (1,65.20) (2,65.00) (3,64.50) (4,63.52)};
\addplot[cMBL,vis op] coordinates {(0,65.50) (1,64.93) (2,63.81) (3,62.14) (4,61.56)};
\addplot[cDBL,vis op] coordinates {(0,65.50) (1,65.30) (2,65.63) (3,65.51) (4,65.03)};
\addplot[cOVE,vis op] coordinates {(0,65.50) (1,64.74) (2,64.83) (3,65.03) (4,64.34)};
\addplot[cBRT,vis op] coordinates {(0,65.50) (1,64.97) (2,62.77) (3,63.11) (4,61.55)};
\addplot[cREM,aud op] coordinates {(0,65.50) (1,62.37) (2,62.11) (3,59.96) (4,54.69)};
\addplot[cMUT,aud op] coordinates {(0,65.50) (1,63.08) (2,63.36) (3,61.60) (4,55.93)};
\addplot[cDST,aud op] coordinates {(0,65.50) (1,65.80) (2,64.13) (3,65.04) (4,63.99)};
\end{groupplot}
\end{tikzpicture}%
}

\vspace{4pt}

% Shared legend, grouped by modality (matches in-panel styles)
\resizebox{\textwidth}{!}{%
\begin{tikzpicture}[x=1cm,y=0.40cm,line cap=round,every node/.style={inner sep=1pt}]
\tikzset{
  legline/.style 2 args={#1,line width=0.7pt,mark=#2,mark size=1.05pt,mark options={solid,line width=0.3pt}},
  lglab/.style={anchor=west,font=\fontsize{6.6pt}{7pt}\selectfont,text=FaultNavy},
  lgrow/.style={anchor=east,font=\fontsize{6.8pt}{7pt}\selectfont\sffamily\bfseries,text=FaultNavy},
}

\node[lgrow] at (0.20,2) {Text};
\draw[legline={cTYP}{*}]                   (0.30,2) -- (0.95,2); \node[lglab] at (1.02,2) {typo\_ocr};
\draw[legline={cDRP}{*}]                   (3.05,2) -- (3.70,2); \node[lglab] at (3.78,2) {drop\_words};
\draw[legline={cSHF}{*}]                   (5.95,2) -- (6.60,2); \node[lglab] at (6.68,2) {word\_shuffle};
\draw[legline={cBRK}{*}]                   (9.10,2) -- (9.75,2); \node[lglab] at (9.83,2) {sentence\_break};

\node[lgrow] at (0.20,1) {Vision};
\draw[densely dashed,legline={cNOI}{square*}] (0.30,1) -- (0.95,1); \node[lglab] at (1.02,1) {noise};
\draw[densely dashed,legline={cOCC}{square*}] (2.15,1) -- (2.80,1); \node[lglab] at (2.88,1) {occlusion};
\draw[densely dashed,legline={cLRS}{square*}] (4.40,1) -- (5.05,1); \node[lglab] at (5.13,1) {low\_resolution};
\draw[densely dashed,legline={cMBL}{square*}] (7.40,1) -- (8.05,1); \node[lglab] at (8.13,1) {motion\_blur};
\draw[densely dashed,legline={cDBL}{square*}] (9.85,1) -- (10.50,1); \node[lglab] at (10.58,1) {defocus\_blur};
\draw[densely dashed,legline={cOVE}{square*}] (12.30,1) -- (12.95,1); \node[lglab] at (13.03,1) {overexposure};
\draw[densely dashed,legline={cBRT}{square*}] (15.05,1) -- (15.70,1); \node[lglab] at (15.78,1) {brightness};

\node[lgrow] at (0.20,0) {Audio};
\draw[dotted,legline={cREM}{triangle*}] (0.30,0) -- (0.95,0); \node[lglab] at (1.02,0) {remove};
\draw[dotted,legline={cMUT}{triangle*}] (2.15,0) -- (2.80,0); \node[lglab] at (2.88,0) {mute};
\draw[dotted,legline={cDST}{triangle*}] (3.80,0) -- (4.45,0); \node[lglab] at (4.53,0) {distortion};
\end{tikzpicture}%
}
\endgroup
\vspace{-20pt}
\caption{Per-model severity curves for all $14$ systems on the fourteen single-modality operators (clean and severities $10/30/50/70$). Modality is encoded by line style (text solid, vision dashed, audio dotted), operator by colour; per-panel $y$-axis is auto-scaled. \texttt{drop\_words}, \texttt{word\_shuffle}, and \texttt{noise} steepen past severity~$30$ on every panel, while several cosmetic vision operators and audio distortion remain closest to clean.}
\vspace{-10pt}
\label{fig:severity-curves}
\end{figure*}

Figure~\ref{fig:severity-curves} disaggregates the severity-$70$ column into the underlying decay shapes. Across all fourteen panels the curves mirror the operator ranking above: text decays broadly downward and accelerates past severity~$30$ (with \opname{drop\_words} and \opname{word\_shuffle} steepest); \opname{noise} alone matches them on the vision side; and several cosmetic vision operators together with audio distortion remain closest to the clean baseline.
Severity curves average three stochastic variants; read trends, not individual points.
A coverage caveat tempers the soft tail. The severity-$70$ column is not a uniform $N{=}273$ slice: pooled across audio and vision operators, annotators retain only $95.6\%$, $87.1\%$, $75.3\%$, and $63.9\%$ of stochastic variants at severity $10/30/50/70$ (Appendix~\ref{app:human-filtering}, Table~\ref{tab:annot-severity}), and the rejection mass concentrates on four operators (\opname{occlusion} $83\%$ pruned at sev~$70$, \opname{brightness} $52\%$, \opname{mute} $56\%$, \opname{remove} $48\%$). The other six operators each stay below $12\%$ rejection across the grid. Because rejected variants are precisely the items annotators could no longer interpret, the retained pool on these four operators at $s{\geq}50$ is biased toward the easier base examples---the most plausible explanation for the small apparent gains on the proprietary side of Table~\ref{tab:main-combined} and for several audio-cell gains in \S\ref{sec:joint}. We therefore treat severity-$70$ numbers on \opname{occlusion}, \opname{brightness}, \opname{mute}, and \opname{remove} as a lower bound on robustness loss, and restrict any "corruption that improves the model" claim to the six confound-free operators.

To make this distinction visible in the main results, Table~\ref{tab:coverage-aware-summary} separates the primary operators from the high-rejection stress set. The primary set contains the four deterministic text operators and six low-rejection audio/vision operators. The comparison shows that the high-rejection operators do not drive the aggregate fragility pattern; we retain them as coverage-aware stress-test evidence.

\begin{table}[t]
\centering
\small
\setlength{\tabcolsep}{10pt}
\renewcommand{\arraystretch}{1.08}
\begin{tabular}{l r}
\toprule
\textbf{Operator group} & \textbf{Mean drop (pp)} \\
\midrule
Primary (10 operators) & \drop{5.31} \\
All operators (14) & \drop{5.14} \\
High-rejection stress set (4) & \drop{4.72} \\
\bottomrule
\end{tabular}
\caption{Coverage-aware severity-70 panel-mean drops by operator group (14 models).}
\label{tab:coverage-aware-summary}
\end{table}

\subsection{Dual- and tri-modal corruption}
\label{sec:joint}

\begin{figure}[t]
\centering
\begingroup
\pgfplotsset{
  jscpanel/.style={
    width=4.0cm, height=2.7cm, scale only axis,
    xmin=50, xmax=200,
    xtick={60,120,180},
    % Per-panel auto-scaled y axis: drops range from -4.4 (Qwen3.5,
    % apparent gain on V+A) up to +23.9 (MiniCPM-o 4.5 on T+V+A),
    % so the previous fixed window (-6..12) clipped roughly a third
    % of the high-end points on Gemini 3.1 Pro and MiniCPM-o 4.5.
    enlarge y limits={value=0.08, upper, lower},
    max space between ticks=20pt,
    try min ticks=4,
    yticklabel style={/pgf/number format/precision=0, /pgf/number format/fixed},
    xlabel={$s_t{+}s_v{+}s_a$},
    ylabel={$\Delta_{\mathrm{clean}}$ (pp)},
    xlabel style={font=\fontsize{6pt}{7pt}\selectfont,yshift=2pt},
    ylabel style={font=\fontsize{6pt}{7pt}\selectfont,yshift=-3pt},
    tick align=outside,
    tick label style={font=\fontsize{5.4pt}{6pt}\selectfont,text=FaultSlate!90},
    title style={font=\fontsize{7pt}{7.6pt}\selectfont\bfseries,text=FaultNavy,yshift=-1pt},
    axis line style={color=FaultSlate!55,line width=0.28pt},
    ymajorgrids=true, xmajorgrids=true,
    grid style={color=FaultSlate!22,line width=0.16pt,dash pattern=on 0.5pt off 0.7pt},
    major tick length=1.2pt,
    enlarge x limits=false,
    clip mode=individual,
  },
}
\tikzset{jscpending/.style={font=\fontsize{5.5pt}{6pt}\selectfont\itshape,text=FaultSlate!85}}
\tikzset{jscclean/.style={font=\fontsize{5.5pt}{6pt}\selectfont,text=FaultSlate!85}}
% \jscbaseref forces y=0 into the (auto-scaled) panel range via a
% transparent helper plot, then draws the dashed clean baseline. The
% helper plot also pins x=50 and x=200 so the baseline always spans
% the full panel even when no datum sits at the panel edges.
\newcommand{\jscbaseref}{%
  \addplot[draw=none,mark=none,forget plot] coordinates {(50,0) (200,0)};%
  \draw[FaultSlate!45,line width=0.25pt,dash pattern=on 0.6pt off 0.8pt]
    (axis cs:50,0) -- (axis cs:200,0);%
}
% \jsccleanlabel places the "clean XX" tag in the bottom-right corner
% relative to the axis box, so it stays visible regardless of the
% per-panel y range.
\newcommand{\jsccleanlabel}[1]{%
  \node[jscclean,anchor=south east,inner sep=1pt]
    at (rel axis cs:0.985,0.02) {clean #1};%
}

\resizebox{\columnwidth}{!}{%
\begin{tikzpicture}
\begin{groupplot}[group style={group size=2 by 2, horizontal sep=0.95cm, vertical sep=1.25cm}, jscpanel]

\nextgroupplot[title={Gemini 3.1 Pro}]
\jscbaseref
\jsccleanlabel{84.15}
\addplot[only marks, mark=*,         mark options={draw=ComboTV, fill=ComboTV},  mark size=1.5pt] coordinates {(60,8.01) (100,7.55) (100,7.68) (140,6.40) (100,0.82) (140,0.30)};
\addplot[only marks, mark=square*,   mark options={draw=ComboTA, fill=ComboTA},  mark size=1.35pt] coordinates {(60,2.60) (80,2.84) (100,4.45) (120,1.40) (120,-0.38) (120,4.83)};
\addplot[only marks, mark=triangle*, mark options={draw=ComboVA, fill=ComboVA},  mark size=1.7pt] coordinates {(60,-3.07) (80,-3.18) (100,4.05) (120,-1.58) (120,1.78) (80,-1.23)};
\addplot[only marks, mark=diamond*,  mark options={draw=ComboTVA,fill=ComboTVA}, mark size=1.7pt] coordinates {(90,2.06) (110,2.51) (130,4.25) (150,5.93) (130,7.60) (150,2.86) (170,10.23) (190,7.86) (190,1.83) (150,2.57) (190,4.88)};

\nextgroupplot[title={Gemini 3 Flash}]
\jscbaseref
\jsccleanlabel{80.95}
\addplot[only marks, mark=*,         mark options={draw=ComboTV, fill=ComboTV},  mark size=1.5pt] coordinates {(60,9.40) (100,10.82) (100,8.45) (140,9.52) (100,4.21) (140,7.09)};
\addplot[only marks, mark=square*,   mark options={draw=ComboTA, fill=ComboTA},  mark size=1.35pt] coordinates {(60,6.92) (80,5.79)  (100,5.63) (120,-1.31) (120,2.00) (120,2.25)};
\addplot[only marks, mark=triangle*, mark options={draw=ComboVA, fill=ComboVA},  mark size=1.7pt] coordinates {(60,0.50) (80,5.04)  (100,3.92) (120,3.37)  (120,3.31) (80,-4.24)};
\addplot[only marks, mark=diamond*,  mark options={draw=ComboTVA,fill=ComboTVA}, mark size=1.7pt] coordinates {(90,4.48) (110,-0.87) (130,2.74) (150,3.35) (130,8.14) (150,3.40) (170,5.47) (190,6.93) (190,-0.73) (150,0.00) (190,1.39)};

\nextgroupplot[title={Qwen3.5-Omni-Plus}]
\jscbaseref
\jsccleanlabel{73.63}
\addplot[only marks, mark=*,         mark options={draw=ComboTV, fill=ComboTV},  mark size=1.5pt] coordinates {(60,5.67) (100,10.61) (100,3.26) (140,7.75) (100,4.34) (140,0.94)};
\addplot[only marks, mark=square*,   mark options={draw=ComboTA, fill=ComboTA},  mark size=1.35pt] coordinates {(60,3.83) (80,-0.42) (100,3.63) (120,-1.56) (120,-0.63) (120,6.37)};
\addplot[only marks, mark=triangle*, mark options={draw=ComboVA, fill=ComboVA},  mark size=1.7pt] coordinates {(60,-4.44) (80,-2.69) (100,1.67) (120,-0.49) (120,0.00) (80,2.03)};
\addplot[only marks, mark=diamond*,  mark options={draw=ComboTVA,fill=ComboTVA}, mark size=1.7pt] coordinates {(90,4.81) (110,6.28) (130,9.55) (150,6.60) (130,5.75) (150,3.93) (170,9.77) (190,8.25) (190,6.79) (150,3.30) (190,9.15)};

\nextgroupplot[title={MiniCPM-o 4.5}]
\jscbaseref
\jsccleanlabel{74.50}
\addplot[only marks, mark=*,         mark options={draw=ComboTV, fill=ComboTV},  mark size=1.5pt] coordinates {(60,10.66) (100,11.05) (100,8.03) (140,9.45) (100,3.62) (140,1.92)};
\addplot[only marks, mark=square*,   mark options={draw=ComboTA, fill=ComboTA},  mark size=1.35pt] coordinates {(60,7.28) (80,4.54) (100,7.87) (120,0.33) (120,3.88) (120,2.72)};
\addplot[only marks, mark=triangle*, mark options={draw=ComboVA, fill=ComboVA},  mark size=1.7pt] coordinates {(60,1.16) (80,0.49) (100,2.70) (120,-0.22) (120,1.90) (80,0.96)};
\addplot[only marks, mark=diamond*,  mark options={draw=ComboTVA,fill=ComboTVA}, mark size=1.7pt] coordinates {(90,3.88) (110,1.73) (130,8.91) (150,6.75) (130,7.21) (150,4.04) (170,11.08) (190,7.10) (190,3.82) (150,0.94) (190,2.42)};

\end{groupplot}
\end{tikzpicture}%
}
\endgroup
\vspace{-26pt}
\caption{Severity-stacking scatter for joint corruption ($29$ cells per deep-dive model). $x$ is the total corruption budget $s_t{+}s_v{+}s_a$, $y$ is the drop from the model's clean baseline; markers encode the combination: \textcolor{ComboTV}{$\bullet$}\,T+V, \textcolor{ComboTA}{$\blacksquare$}\,T+A, \textcolor{ComboVA}{$\blacktriangle$}\,V+A, \textcolor{ComboTVA}{$\blacklozenge$}\,T+V+A. The worst point is asymmetric (mildly corrupted text $t30$ $\times$ heavily corrupted vision $v70$), the trimodal staircase climbs with the vision component, and points below $\Delta{=}0$ are coverage-induced apparent gains (\S\ref{sec:single}).}
\vspace{-15pt}
\label{fig:joint-scatter}
\end{figure}
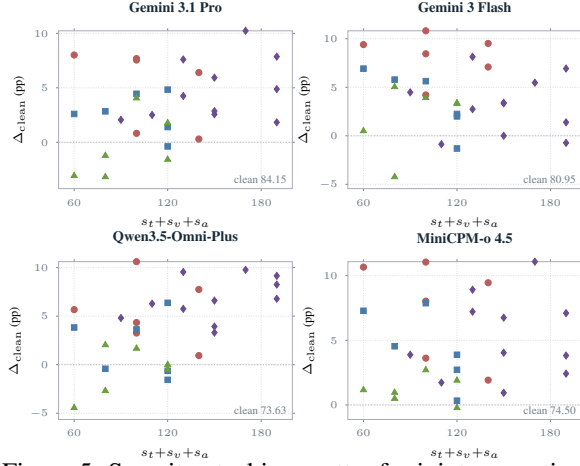

Joint corruption is treated as a depth probe on four deep-dive systems---\modelname{Gemini 3.1 Pro} and \modelname{Gemini 3 Flash} on the proprietary side, \modelname{Qwen3.5-Omni-Plus} and \modelname{MiniCPM-o 4.5} on the open side---using as building blocks the six operators with the steepest single-modality drops from Section~\ref{sec:single} (\texttt{drop\_words}, \texttt{word\_shuffle}, \texttt{noise}, \texttt{occlusion}, \texttt{mute}, \texttt{remove}). For each of the four combinations T+V, T+A, V+A, T+V+A we sweep a canonical $\{30,70\}^k$ severity grid and add replacement-pressure cells that swap one operator within a modality, yielding $29$ cells per model (full row-by-row enumeration in Table~\ref{tab:joint-cells}, Appendix~\ref{app:joint-cells}). Figure~\ref{fig:joint-scatter} plots every cell as one point: $x$ = total corruption budget $s_t{+}s_v{+}s_a$, $y$ = drop from the model's clean baseline.

Three phenomena dominate the joint-corruption scatter; cell-by-cell numerical evidence is deferred to Table~\ref{tab:joint-cells} in Appendix~\ref{app:joint-cells}. \textbf{(i) The worst cell is asymmetric, not symmetric.} On every deep-dive model the single steepest drop is not a heavy-on-both configuration but the asymmetric \texttt{drop\_words}\,$\times$\,\texttt{noise} cell at light text damage paired with heavy visual noise---the topmost red point in each panel of Figure~\ref{fig:joint-scatter}. Inside the T+V block, severity stacking is in fact non-monotone in the text dimension: pushing the text operator from $s_t{=}30$ to $s_t{=}70$ at fixed $s_v{=}70$ \emph{recovers} several percentage points on every model, so the red points trend \emph{down} as the corruption budget grows along the text axis. One possible behavioral explanation is that, when text is more visibly corrupted, models rely differently on the remaining visual signal; the observed contrast is between mildly corrupted text ($t30$) and heavily corrupted vision ($v70$), not clean text and broken vision. \textbf{(ii) The trimodal staircase climbs with vision, not with corruption count.} Across the eight canonical T+V+A cells, the four with $s_v{=}70$ are uniformly several pp worse than the four with $s_v{=}30$ on every deep-dive model, and the diamond staircase in Figure~\ref{fig:joint-scatter} climbs almost entirely with the vision component while text and audio severities contribute little. Joint damage is therefore not a function of how many modalities are perturbed; it is dominated by whichever channel is structurally broken hardest, with text--vision providing the steepest shared fault line.

\vspace{-8pt}
\textbf{(iii) Operator identity matters within text, but not within vision or audio.} The replacement-pressure cells of Table~\ref{tab:joint-cells} reveal that swapping one operator for another inside the text family produces a large model-specific swing on the open side (Qwen) but a much smaller swing on the proprietary side (Gemini), whereas swapping audio-side or vision-side operators (\texttt{remove}/\texttt{mute}, \texttt{occlusion}/\texttt{noise}) shifts accuracy by at most a couple of percentage points on both models---a text-robustness signature that the single-modality matrix of \S\ref{sec:single} cannot reveal. Finally, the apparent gains visible at the bottom of each scatter panel are the survivorship-bias caveat of \S\ref{sec:single} resurfacing under joint conditions: they cluster strictly in audio-involving cells, and only a handful of cells across the panel fall \emph{below} the weakest single-modality cell that composes them (all of those in vision-at-severity-$70$ conditions on the open side; none on Gemini). Together these three observations converge on one picture: cross-modal damage is structured, not additive; the worst point is the asymmetric $t30/v70$ text--vision contrast; and clean omni-modal accuracy tells us only that an answer can be found when all evidence is intact, not whether it survives once any trusted channel is structurally degraded.

Read together, the three observations also reframe what the clean omni-modal score is actually measuring. A clean tri-modal accuracy guarantees only that the model can find an answer when every channel is well-formed and mutually redundant; it does not certify that the model would still use the same evidence once one channel becomes structurally unreliable. The asymmetric $t30/v70$ text--vision cell is the most diagnostic instance of this gap, because it is precisely the regime that both clean evaluation and missing-modality ablation fail to surface---one assumes the channel vanishes, the other assumes it stays intact, while the observed pattern is consistent with greater reliance on the relatively less corrupted text channel. The structural fault lines mapped by \method{} are therefore behavioral patterns observed in the evaluated omni-modal systems, visible only because the corruption keeps every channel simultaneously present and human-interpretable. 

% Whether this binding behaviour can be retrained out, or whether it reflects a structural property of how omni-modal models currently route evidence across channels, is the question the rest of the paper deliberately leaves open.

\section{Related Work}
\label{sec:related}

\paragraph{Omni-modal models and clean-input benchmarks.}
Recent omni-modal systems extend language models with vision, video, and audio inputs, spanning proprietary APIs (\modelname{Gemini}~\citep{team2023gemini}) and open releases (\modelname{Qwen-Omni}~\citep{team2026qwen3}, \modelname{MiniCPM-o}~\citep{openbmb2025minicpmo}, \modelname{OmniVinci}~\citep{ye2025omnivinci}, \modelname{Ola}~\citep{liu2025ola}, \modelname{Baichuan-Omni}~\citep{li2024baichuan}). They are typically evaluated either on clean tri-modal QA suites (Social-IQ~\citep{zadeh2019social}, OmniBench~\citep{li2026omnibench}, VALOR~\citep{liu2024valor}) or on clean-input distractor diagnostics (MMBench~\citep{liu2024mmbench}, HallusionBench~\citep{guan2024hallusionbench}, MMMU~\citep{yue2024mmmu}) that probe language priors~\citep{agrawal2018don} and answer-option shortcuts under uncorrupted inputs. Such benchmarks cannot ask whether the same model still binds evidence correctly once a modality channel is degraded ~\cite{kang2026hssbench,kang2026multimodal,kang2026quanteval,10.1145/3774904.3792075,luo2024codis,feng2025seeing,wang2025mucar,he2026order,liu2026reasonact,gao2026laobench,shi2025safety,shi2026spader,shicontrollable,li2025frequency,11541222,li2026comprehensive,li2026mol}.

\vspace{-6pt}
\paragraph{Structural corruption and our positioning.}
Corruption benchmarks have measured single-modality robustness in vision and audio since \modelname{ImageNet-C}~\citep{hendrycks2019benchmarking}, and extend to multimodal settings via \modelname{MM-Robustness}~\citep{qiu2022benchmarking}, \modelname{MMCBench}~\citep{zhang2024benchmarking}, and the perturbation suite distributed with MMMU~\citep{yue2024mmmu}, but they corrupt one modality at a time and never the cross-modal relation. Table~\ref{tab:benchmark-comparison} (Appendix~\ref{app:benchmark-comparison}) contrasts \method{} with these lines along six design axes. The combination of (a) tri-modal text+vision+audio coverage, (b) operator-level structural corruption that keeps every modality present, (c) a graded severity sweep in $\{10,30,50,70\}$, (d) per-variant human verification, and (e) bimodal/trimodal joint corruption with operator-replacement pressure cells is what enables the $\dropbase$-vs-$\dropsingle$ contrast in Section~\ref{sec:joint} that previous benchmarks structurally cannot ask.

\section{Conclusion}
\label{sec:conclusion}

We introduced \method{} (\emph{\methodfull}), a structure-corruption evaluation protocol for omni-modal models. Instead of removing modalities, \method{} keeps text, vision, and audio present while damaging their internal evidence structure. Across the reported model panel, the same pattern recurs: structural corruption reduces performance relative to the clean all-modality baseline, text--vision corruption emerges as the most stable bimodal fault line, and degradation is not explained by a simple count of corrupted modalities. The evidence supports a direct conclusion: omni-modal models are not equally robust to all forms of multi-modal structural degradation, and clean benchmark accuracy alone is insufficient to characterise how reliably they use evidence across modalities.

% \clearpage
\section{Limitations}
\method{} is intended as a controlled diagnostic rather than an exhaustive robustness census. Its verified base set contains $273$ tri-modal examples from three English-language benchmarks, with a curated inventory of fourteen structural operators designed to test cross-modal evidence assembly. The joint-corruption analysis focuses on four deep-dive models, while the remaining systems are covered by the lightweight panel in Table~\ref{tab:lightweight-panel}. Because variants rejected as uninterpretable are concentrated in \opname{occlusion}, \opname{brightness}, \opname{mute}, and \opname{remove}, the corresponding severity-$70$ results should be read as conservative estimates; claims that corruptions improve performance are therefore restricted to the six confound-free operators discussed in \S\ref{sec:single}.

The protocol evaluates multiple-choice answer selection through a fixed JSON schema, so it characterizes robustness under structured decision settings rather than open-ended generation. The corruption operators are static and human-readable rather than adversarially optimized, making the reported failures a lower-bound estimate of possible corruption sensitivity. Finally, \method{} is behavioural: it identifies robust empirical patterns, including the asymmetric $t30/v70$ text--vision contrast, vision-dominated joint damage, and model-specific text-operator sensitivity, while leaving representation-level causal explanations to future probing and controlled training studies.

\section*{Ethics Statement}

This work studies model robustness under controlled input corruptions with the goal of identifying failure modes before deployment in noisy multimodal environments. The benchmark does not introduce new user data beyond the evaluated source examples. Any release of derived assets should respect the licences and privacy constraints of the underlying data.

\appendix

% Loosen LaTeX's default float-placement parameters inside the appendix.
% The appendix is table-heavy (32 tables, mostly two-column-wide
% \texttt{table*}), and the default thresholds force most of them to be
% deferred to the very end of the document, where they pile up out of
% reading order. The values below allow up to four float at the top of a
% page, six floats per page total, and let a near-full page (95% of text
% area) be devoted to floats while still printing some text. This keeps
% each table near the paragraph that introduces it.
\setcounter{topnumber}{4}
\setcounter{bottomnumber}{2}
\setcounter{totalnumber}{6}
\setcounter{dbltopnumber}{4}
\renewcommand{\topfraction}{0.95}
\renewcommand{\bottomfraction}{0.5}
\renewcommand{\textfraction}{0.05}
\renewcommand{\floatpagefraction}{0.7}
\renewcommand{\dbltopfraction}{0.95}
\renewcommand{\dblfloatpagefraction}{0.7}

\bibliography{referencebib}

\section{Appendix Overview: What Is Main Evidence and What Is Protocol}
\label{app:overview}

The appendix supplements the main results with detailed analyses and evaluation protocols:
\begin{itemize}[leftmargin=*]
    \item \textbf{Supplementary results.} Appendix~\ref{app:resultslots} contains the operator-level, source-level, control, statistical, and mechanism-probe breakdowns that drill down beyond the headline matrix in Section~\ref{sec:results} and the dual-/tri-modal summary in Section~\ref{sec:joint}. Three additional diagnostic subsections sit at the end of this appendix: Appendix~\ref{app:reliability} (seed-variance and fragility-slope diagnostics), Appendix~\ref{app:open-vs-prop} (open vs.\ proprietary aggregated comparison), and Appendix~\ref{app:interaction-cal} (joint-corruption sub-additivity, calibration, and cross-model failure overlap).
    \item \textbf{Data curation.} Appendix~\ref{app:human-filtering} documents how the 273-example verified base set was carved out of the source benchmarks: the 300-candidate pool, the third-party annotation workflow, the per-stage filtering funnel, the inter-annotator agreement summary (Table~\ref{tab:iaa-summary}), and the audit hooks used to detect bias in the rejected pool.
    \item \textbf{Operator definitions.} Appendix~\ref{app:operator-defs} formalises the fourteen structural corruption operators sketched in Figure~\ref{fig:corruption-taxonomy}, including the design rules that decide what is in scope and which corruption families are deliberately excluded.
    \item \textbf{Setup and protocol.} Appendix~\ref{app:expsetup} records the 14-model roster, the frame-extracted visual-input protocol, the 12-cell lightweight panel, the prompting and inference protocol, and the headline reporting rule. Appendix~\ref{app:expanded-protocol} then gives the per-experiment construction details (single-modality, dual-/tri-modal, the 46-cell extended joint suite enumerated in Table~\ref{tab:46cell-enumeration}, controls, mechanism probes), and Appendix~\ref{app:prompts} documents the standardised prompt templates and parsing rules.
\end{itemize}
This structure keeps the main paper focused on the structural-corruption phenomenon while documenting the experimental details needed for reproducibility.

\section{Benchmark Comparison Details}
\label{app:benchmark-comparison}

Table~\ref{tab:benchmark-comparison} expands the positioning argument of Section~\ref{sec:related}: it compares \method{} against three families of closely related benchmarks (clean tri-modal QA, hallucination/shortcut/distractor suites with clean modalities, and multimodal corruption benchmarks that perturb a single modality) across a set of comparison dimensions.

\begin{table*}[!htbp]
\centering
\scriptsize
\setlength{\tabcolsep}{3.0pt}
\renewcommand{\arraystretch}{1.0}
\resizebox{\textwidth}{!}{%
\begin{tabular}{p{0.16\textwidth}p{0.13\textwidth}rcccccc}
\toprule
\textbf{Benchmark} & \textbf{Modalities} & \textbf{$N$} & \textbf{Struct.\ corruption} & \textbf{Sev.\ sweep} & \textbf{Joint corruption} & \textbf{Audio/vision variant verif.} & \textbf{Sample-paired protocol} & \textbf{Models eval.} \\
\midrule
\faultgroup{9}{Clean-input omni-modal QA benchmarks}
\modelrow Social-IQ~\citep{zadeh2019social}   & video, audio, text & 7.5k & \xmark & \xmark & \xmark & \xmark & \xmark & --- \\
\softrow  OmniBench~\citep{li2026omnibench}  & image, audio, text & 1.1k & \xmark & \xmark & \xmark & \xmark & \xmark & --- \\
\modelrow VALOR~\citep{liu2024valor}          & video, audio, text & 32k  & \xmark & \xmark & \xmark & \xmark & \xmark & --- \\
\midrule
\faultgroup{9}{Hallucination / shortcut / distractor suites (clean modalities)}
\modelrow MMBench~\citep{liu2024mmbench}              & image, text        & 3.2k & \xmark            & \xmark & \xmark & \xmark         & \xmark         & --- \\
\softrow  HallusionBench~\citep{guan2024hallusionbench} & image, text       & 1.1k & curated distract. & \xmark & \xmark & item-level     & paired examples & --- \\
\modelrow MMMU~\citep{yue2024mmmu}                    & image, text        & 11.5k & \xmark           & \xmark & \xmark & item-level     & \xmark         & --- \\
\midrule
\faultgroup{9}{Multimodal corruption / robustness benchmarks (single-modality corruption only)}
\modelrow MM-Robustness~\citep{qiu2022benchmarking}   & image, text        & --- & image corruption only & 5 levels & \xmark & \xmark & sample-paired & --- \\
\softrow  MMCBench~\citep{zhang2024benchmarking}            & image, audio, text & --- & per-modality only     & 4 levels & \xmark & \xmark & sample-paired & --- \\
\modelrow MMMU-Perturbation~\citep{yue2024mmmu}       & image, text        & --- & image perturb.\ only  & 3 levels & \xmark & \xmark & sample-paired & --- \\
\midrule
\faultgroup{9}{This work}
\faultrow \textbf{\method{} (ours)} & \textbf{video/image, audio, text} & \textbf{273} & \textbf{14 ops, structural only} & \textbf{4 levels} & \textbf{29-/46-cell joint suite} & \textbf{19{,}944 A/V cells (\,80.5\% retained)} & \textbf{sample-paired, fixed gold answer} & \textbf{14} \\
\bottomrule
\end{tabular}%
}
\caption{Design-axis comparison of \method{} against the most closely related multimodal evaluation benchmarks (\S\ref{sec:related}). \textbf{Struct.\ corruption} means operator-level structural damage that keeps the modality present, as opposed to missing-modality ablation or semantic perturbation; \textbf{Joint corruption} requires applying corruption to two or more modalities of the same example; \textbf{Audio/vision variant verification} means that every corrupted audio or visual variant (not just every clean example) is human-judged for interpretability. Deterministic text corruptions are assessed separately in the blind answer-preservation audit; \textbf{Sample-paired protocol} means clean and corrupted versions share the same base example and gold answer. ``---'' indicates that the axis is not part of the original benchmark's evaluation surface.}
\label{tab:benchmark-comparison}
\end{table*}

\section{Supplementary Experimental Result Tables}
\label{app:resultslots}

This appendix presents operator-level, source-level, control, and mechanism-probe analyses that complement the headline single-modality matrix in Section~\ref{sec:results} and the dual-/tri-modal summary in Section~\ref{sec:joint}.

\subsection{Operator-level joint-corruption breakdown for the deep-dive models}
\label{app:joint-cells}

Table~\ref{tab:joint-cells} is the full operator-level table that underlies the per-model scatter in Figure~\ref{fig:joint-scatter} and contains every joint-corruption number quoted in Section~\ref{sec:joint}. Each row is one of the $29$ operator$\times$severity cells defined in Section~\ref{sec:joint}; columns are the four deep-dive models. Every model reports mean-variant accuracy together with the absolute change from its own clean all-modality baseline ($\downarrow$~=~drop, $\uparrow$~=~relative gain, $\pm 0$~=~unchanged); \modelname{Gemini 3 Flash} and \modelname{Qwen3.5-Omni-Plus} are the canonical references that anchor Figure~\ref{fig:joint-scatter}, while \modelname{Gemini 3.1 Pro} and \modelname{MiniCPM-o 4.5} are paired alongside them so the table is read column-by-column. Cells without the \repmark{} marker belong to the canonical severity grid; cells marked \repmark{} are replacement-pressure cells that swap one operator for a different operator family at the same modality. Severity codes use \textit{t}/\textit{v}/\textit{a} for text/vision/audio at the given severity level (e.g.\ \texttt{t30/v70} = text operator at severity~30 and vision operator at severity~70).

\begin{table*}[!t]
\centering
\scriptsize
\setlength{\tabcolsep}{4pt}
\renewcommand{\arraystretch}{1.04}
\begin{tabular}{@{}l c c c c c@{}}
\toprule
\textbf{Operators} & \textbf{Severities} & \textbf{Gemini 3.1 Pro} & \textbf{Gemini 3 Flash} & \textbf{Qwen3.5-Omni-Plus} & \textbf{MiniCPM-o 4.5} \\
 & & {\fontsize{5.6pt}{6pt}\selectfont clean $84.15$} & {\fontsize{5.6pt}{6pt}\selectfont clean $80.95$} & {\fontsize{5.6pt}{6pt}\selectfont clean $73.63$} & {\fontsize{5.6pt}{6pt}\selectfont clean $74.50$} \\
\midrule
\jointbandsix{Text+Vision \;\textbar\; canonical = \texttt{drop\_words}\,$\times$\,\texttt{noise} (4 cells) + 2 replacement\repmark{} cells}
\opname{drop\_words $\times$ noise}              & \sevcell{t30/v30} & \maincell{76.14}{8.01} & \maincell{71.55}{9.40}  & \maincell{67.96}{5.67}  & \maincell{63.84}{10.66} \\
\opname{drop\_words $\times$ noise}              & \sevcell{t30/v70} & \maincell{76.60}{7.55} & \maincell{70.14}{10.82} & \maincell{63.01}{10.61} & \maincell{63.45}{11.05} \\
\opname{drop\_words $\times$ noise}              & \sevcell{t70/v30} & \maincell{76.47}{7.68} & \maincell{72.50}{8.45}  & \maincell{70.37}{3.26}  & \maincell{66.47}{8.03} \\
\opname{drop\_words $\times$ noise}              & \sevcell{t70/v70} & \maincell{77.75}{6.40} & \maincell{71.43}{9.52}  & \maincell{65.88}{7.75}  & \maincell{65.05}{9.45} \\
\opname{drop\_words $\times$ occlusion}\repmark{} & \sevcell{t70/v30} & \maincell{83.33}{0.82} & \maincell{76.74}{4.21}  & \maincell{69.29}{4.34}  & \maincell{70.88}{3.62} \\
\opname{word\_shuffle $\times$ noise}\repmark{}   & \sevcell{t70/v70} & \maincell{83.85}{0.30} & \maincell{73.86}{7.09}  & \maincell{72.69}{0.94}  & \maincell{72.58}{1.92} \\
\midrule
\jointbandsix{Text+Audio \;\textbar\; canonical = \texttt{drop\_words}\,$\times$\,\texttt{mute} (4 cells) + 2 replacement\repmark{} cells}
\opname{drop\_words $\times$ mute}               & \sevcell{t30/a30} & \maincell{81.55}{2.60} & \maincell{74.03}{6.92} & \maincell{69.80}{3.83} & \maincell{67.22}{7.28} \\
\opname{drop\_words $\times$ mute}               & \sevcell{t30/a50} & \maincell{81.31}{2.84} & \maincell{75.16}{5.79} & \maingain{74.05}{0.42} & \maincell{69.96}{4.54} \\
\opname{drop\_words $\times$ mute}               & \sevcell{t70/a30} & \maincell{79.70}{4.45} & \maincell{75.32}{5.63} & \maincell{70.00}{3.63} & \maincell{66.63}{7.87} \\
\opname{drop\_words $\times$ mute}               & \sevcell{t70/a50} & \maincell{82.75}{1.40} & \maingain{82.26}{1.31} & \maingain{75.19}{1.56} & \maincell{74.17}{0.33} \\
\opname{drop\_words $\times$ remove}\repmark{}   & \sevcell{t70/a50} & \maingain{84.53}{--0.38} & \maincell{78.95}{2.00} & \maingain{74.26}{0.63} & \maincell{70.62}{3.88} \\
\opname{word\_shuffle $\times$ mute}\repmark{}   & \sevcell{t70/a50} & \maincell{79.32}{4.83} & \maincell{78.70}{2.25} & \maincell{67.26}{6.37} & \maincell{71.78}{2.72} \\
\midrule
\jointbandsix{Vision+Audio \;\textbar\; canonical = \texttt{noise}\,$\times$\,\texttt{mute} (4 cells) + 2 replacement\repmark{} cells}
\opname{noise $\times$ mute}                     & \sevcell{v30/a30} & \maingain{87.22}{--3.07} & \maincell{80.45}{0.50} & \maingain{78.07}{4.44} & \maincell{73.34}{1.16} \\
\opname{noise $\times$ mute}                     & \sevcell{v30/a50} & \maingain{87.33}{--3.18} & \maincell{75.91}{5.04} & \maingain{76.32}{2.69} & \maincell{74.01}{0.49} \\
\opname{noise $\times$ mute}                     & \sevcell{v70/a30} & \maincell{80.10}{4.05} & \maincell{77.03}{3.92} & \maincell{71.96}{1.67} & \maincell{71.80}{2.70} \\
\opname{noise $\times$ mute}                     & \sevcell{v70/a50} & \maingain{85.73}{--1.58} & \maincell{77.58}{3.37} & \maingain{74.12}{0.49} & \maingain{74.72}{--0.22} \\
\opname{noise $\times$ remove}\repmark{}         & \sevcell{v70/a50} & \maincell{82.37}{1.78} & \maincell{77.64}{3.31} & \mainflat{73.63}       & \maincell{72.60}{1.90} \\
\opname{occlusion $\times$ mute}\repmark{}       & \sevcell{v30/a50} & \maingain{85.38}{--1.23} & \maingain{85.19}{4.24} & \maincell{71.60}{2.03} & \maincell{73.54}{0.96} \\
\midrule
\jointbandsix{Text+Vision+Audio \;\textbar\; canonical = \texttt{drop\_words}\,$\times$\,\texttt{noise}\,$\times$\,\texttt{mute} (8 cells) + 3 replacement\repmark{} cells}
\opname{drop\_words $\times$ noise $\times$ mute}              & \sevcell{t30/v30/a30} & \maincell{82.09}{2.06} & \maincell{76.47}{4.48} & \maincell{68.82}{4.81} & \maincell{70.62}{3.88} \\
\opname{drop\_words $\times$ noise $\times$ mute}              & \sevcell{t30/v30/a50} & \maincell{81.64}{2.51} & \maingain{81.82}{0.87} & \maincell{67.35}{6.28} & \maincell{72.77}{1.73} \\
\opname{drop\_words $\times$ noise $\times$ mute}              & \sevcell{t30/v70/a30} & \maincell{79.90}{4.25} & \maincell{78.21}{2.74} & \maincell{64.08}{9.55} & \maincell{65.59}{8.91} \\
\opname{drop\_words $\times$ noise $\times$ mute}              & \sevcell{t30/v70/a50} & \maincell{78.22}{5.93} & \maincell{77.60}{3.35} & \maincell{67.03}{6.60} & \maincell{67.75}{6.75} \\
\opname{drop\_words $\times$ noise $\times$ mute}              & \sevcell{t70/v30/a30} & \maincell{76.55}{7.60} & \maincell{72.81}{8.14} & \maincell{67.88}{5.75} & \maincell{67.29}{7.21} \\
\opname{drop\_words $\times$ noise $\times$ mute}              & \sevcell{t70/v30/a50} & \maincell{81.29}{2.86} & \maincell{77.55}{3.40} & \maincell{69.70}{3.93} & \maincell{70.46}{4.04} \\
\opname{drop\_words $\times$ noise $\times$ mute}              & \sevcell{t70/v70/a30} & \maincell{73.92}{10.23} & \maincell{75.48}{5.47} & \maincell{63.86}{9.77} & \maincell{63.42}{11.08} \\
\opname{drop\_words $\times$ noise $\times$ mute}              & \sevcell{t70/v70/a50} & \maincell{76.29}{7.86} & \maincell{74.02}{6.93} & \maincell{65.38}{8.25} & \maincell{67.40}{7.10} \\
\opname{drop\_words $\times$ noise $\times$ remove}\repmark{}   & \sevcell{t70/v70/a50} & \maincell{82.32}{1.83} & \maingain{81.68}{0.73} & \maincell{66.84}{6.79} & \maincell{70.68}{3.82} \\
\opname{drop\_words $\times$ occlusion $\times$ mute}\repmark{} & \sevcell{t70/v30/a50} & \maincell{81.58}{2.57} & \mainflat{80.95}       & \maincell{70.33}{3.30} & \maincell{73.56}{0.94} \\
\opname{word\_shuffle $\times$ noise $\times$ mute}\repmark{}   & \sevcell{t70/v70/a50} & \maincell{79.27}{4.88} & \maincell{79.56}{1.39} & \maincell{64.48}{9.15} & \maincell{72.08}{2.42} \\
\bottomrule
\end{tabular}
\caption{Full operator-level joint-corruption breakdown for the four deep-dive models. Each row is one of the 29 operator$\times$severity cells defined in Section~\ref{sec:joint}: the four canonical T+V cells (\texttt{drop\_words}\,$\times$\,\texttt{noise} at $\{30,70\}{\times}\{30,70\}$) and two replacement cells; the four canonical T+A cells and two replacements; the four canonical V+A cells and two replacements; and the eight canonical T+V+A cells (cube over $t\in\{30,70\}, v\in\{30,70\}, a\in\{30,50\}$) and three replacements. Cells marked \repmark{} substitute a different operator family (\texttt{occlusion} for \texttt{noise}, \texttt{remove} for \texttt{mute}, or \texttt{word\_shuffle} for \texttt{drop\_words}) at the strongest severity inside that combination. Each model cell shows the mean-variant accuracy together with the absolute change from the model's clean all-modality baseline ($\downarrow$ red = drop, $\uparrow$ green = relative gain, $\pm 0.00$ = unchanged). All four deep-dive panels are reported here; \modelname{Gemini 3 Flash} and \modelname{Qwen3.5-Omni-Plus} provide the canonical reference points used in Figure~\ref{fig:joint-scatter}, while \modelname{Gemini 3.1 Pro} and \modelname{MiniCPM-o 4.5} are paired in the same table so that the proprietary vs.\ open-API comparison is preserved cell-by-cell.}
\label{tab:joint-cells}
\end{table*}

\subsection{Single-modality severity grid by operator}
\label{app:single-grid}

Table~\ref{tab:single-severity-grid} expands the headline severity-70 single-modality matrix in the main body (Table~\ref{tab:main-combined}) into the full $4 \times 14$ severity grid for two deep-dive systems, at severities $\{10,30,50,70\}$. This grid is the operator-level evidence behind the per-modality robustness curves in Figure~\ref{fig:severity-curves}: text \texttt{drop\_words} is the most damaging text operator for both systems, visual \texttt{noise} is the most damaging vision operator (and especially severe for \modelname{Gemini 3 Flash} at severity 70), audio segment muting/removal is the most damaging audio operator for both systems, while audio \texttt{distortion} is consistently the weakest audio operator. The bolded value in each row is that model's weakest observed cell for that operator and defines the $\dropsingle$ baseline used in joint-corruption analysis.

\begin{table*}[!t]
\centering
\scriptsize
\setlength{\tabcolsep}{2.55pt}
\renewcommand{\arraystretch}{0.82}
\begin{tabular}{llrrrrrrrr}
\toprule
\textbf{Mod.} & \textbf{Operator} & \multicolumn{4}{c}{\textbf{\modelname{Qwen3.5-Omni-Plus}}} & \multicolumn{4}{c}{\textbf{\modelname{Gemini 3 Flash}}} \\
 & & \textbf{s10} & \textbf{s30} & \textbf{s50} & \textbf{s70} & \textbf{s10} & \textbf{s30} & \textbf{s50} & \textbf{s70} \\
\midrule
\faultgroup{10}{Text severity grid}
\variantrow Text & \texttt{typo\_ocr} & 72.91 & 72.93 & 71.16 & \textbf{69.60} & 79.40 & 75.40 & 74.04 & \textbf{71.06} \\
\variantrow Text & \texttt{drop\_words} & 71.57 & 72.21 & 67.20 & \textbf{64.10} & 81.85 & 71.98 & 71.60 & \textbf{67.40} \\
\variantrow Text & \texttt{word\_shuffle} & 71.58 & 69.04 & 65.63 & \textbf{65.20} & 80.64 & 77.37 & 74.47 & \textbf{70.33} \\
\variantrow Text & \texttt{sentence\_break} & 74.33 & 72.54 & 71.27 & \textbf{69.60} & 80.82 & 78.03 & 73.33 & \textbf{72.89} \\
\midrule
\faultgroup{10}{Vision severity grid}
\softrow Vision & \texttt{noise} & 73.62 & 72.32 & 71.91 & \textbf{70.37} & 78.72 & 74.53 & 71.84 & \textbf{67.79} \\
\softrow Vision & \texttt{occlusion} & 72.99 & 71.31 & \textbf{70.29} & 70.49 & 80.80 & 80.68 & 80.88 & \textbf{80.33} \\
\softrow Vision & \texttt{low\_resolution} & 73.21 & 73.47 & 72.55 & \textbf{71.79} & 81.57 & \textbf{80.42} & 80.54 & 80.97 \\
\softrow Vision & \texttt{motion\_blur} & 72.56 & 72.80 & 72.44 & \textbf{71.02} & 81.85 & 80.23 & \textbf{77.34} & 77.59 \\
\softrow Vision & \texttt{defocus\_blur} & 73.25 & 73.03 & 72.97 & \textbf{71.83} & 81.34 & 80.32 & 79.67 & \textbf{78.95} \\
\softrow Vision & \texttt{overexposure} & 73.78 & 72.78 & 73.25 & \textbf{71.65} & 79.50 & 81.09 & 80.13 & \textbf{79.37} \\
\softrow Vision & \texttt{brightness} & 74.28 & \textbf{73.45} & 73.62 & 73.68 & \textbf{80.07} & 80.77 & 80.36 & 80.30 \\
\midrule
\faultgroup{10}{Audio severity grid}
\modelrow Audio & \texttt{remove} & 71.53 & 73.13 & 73.21 & \textbf{70.95} & 80.45 & 76.64 & 75.79 & \textbf{74.30} \\
\modelrow Audio & \texttt{mute} & 71.26 & 71.92 & 71.96 & \textbf{69.72} & 80.31 & 76.10 & 74.25 & \textbf{73.24} \\
\modelrow Audio & \texttt{distortion} & 73.98 & \textbf{73.31} & 75.15 & 75.00 & 80.11 & 80.42 & 79.57 & \textbf{79.17} \\
\bottomrule
\end{tabular}
\caption{Full single-modality severity grid. Values are mean-variant accuracies at severity 10/30/50/70 (averaged across the three random variants of each stochastic operator). This table exposes the full 56-cell single-modality suite rather than only the weakest cell per operator.}
\label{tab:single-severity-grid}
\end{table*}

The operator-level view clarifies why text--vision becomes the dominant joint fault line. The most damaging text operator is word dropping and the most damaging visual operator is additive noise; the single strongest joint condition combines exactly these two families at light text damage paired with heavy visual noise (full cell-by-cell values in Table~\ref{tab:joint-cells}). Audio degradation is real, especially under segment muting/removal, but distortion alone is weak, which helps explain why text--audio and vision--audio combinations are less stable as shared fault lines.

\subsection{Does combined corruption exceed the weakest single modality?}

A stricter question is whether a combined condition is worse than its weakest corresponding single-modality corruption. The answer is mostly no. For \modelname{Gemini 3 Flash}, no condition has positive $\dropsingle$: every joint cell stays above the model's weakest single-modality cell (\opname{drop\_words} at severity~$70$), so the gap remains negative across all $29$ joint cells. For \modelname{Qwen3.5-Omni-Plus}, the same is true on $26$ of the $29$ cells; the three cells that narrowly cross the threshold are all \opname{drop\_words}$\times$\opname{noise} configurations with heavy visual severity ($s_v{=}70$)---one bimodal at $t30/v70$ and two trimodal cells with $v70$ in the cube---and the largest exceedance among them is only $+1.09$\,pp (per-cell values in Table~\ref{tab:joint-cells}).

This distinction is important for interpretation. The robust conclusion is not that combined corruption is always worse than any single corruption. The robust conclusion is that combined corruption consistently lowers performance relative to the clean all-modality baseline, and that a small number of high-pressure conditions can exceed the weakest single-modality degradation. A stricter additive null is reported in Table~\ref{tab:subadditivity-test} (Appendix~\ref{app:interaction-cal}): the deep-dive panel is sub-additive on average across all four combination types (joint $\dropbase$ is gentler than the sum of single-modality $\dropbase$s by $+0.94$ to $+22.62$ percentage points on average), confirming that the joint signal is dominated by the strongest single component rather than by an emergent multi-modal collapse. \modelname{Qwen3.5-Omni-Plus} is the closest to the additive baseline ($+0.94$\,pp on T+V; $+2.58$\,pp on the eight T+V+A cells), whereas the proprietary panel exhibits much stronger sub-additivity ($+11$ to $+23$\,pp on \modelname{Gemini 3 Flash}).

\subsection{Coverage and invalid outputs}

Table~\ref{tab:coverage} reports coverage for the combined runs. The invalid-output count is large enough to be part of the result rather than hidden bookkeeping.

\begin{table}[!htbp]
\centering
\scriptsize
\setlength{\tabcolsep}{3.0pt}
\renewcommand{\arraystretch}{0.95}
\begin{tabular}{llcccc}
\toprule
\textbf{Model} & \textbf{Group} & \textbf{Valid/expected} & \textbf{Trials} & \textbf{Invalid} & \textbf{Acc.} \\
\midrule
\faultgroup{6}{Coverage and invalid-output pressure}
\modelrow \modelname{Qwen3.5-Omni-Plus} & Bimodal  & 4{,}368/4{,}368 & 4{,}368 & 53 & 67.42 \\
\softrow  \modelname{Qwen3.5-Omni-Plus} & Trimodal & 6{,}552/6{,}552 & 6{,}552 & 124 & 64.78 \\
\modelrow \modelname{Gemini 3 Flash} & Bimodal  & 4{,}368/4{,}368 & 4{,}368 & 41 & 74.16 \\
\softrow  \modelname{Gemini 3 Flash} & Trimodal & 6{,}552/6{,}552 & 6{,}552 & 98 & 72.18 \\

\bottomrule
\end{tabular}
\caption{Coverage statistics for combined-corruption runs. The invalid column is decomposed by failure type in Table~\ref{tab:invalid-breakdown}.}
\label{tab:coverage}
\end{table}

These failures can arise from API failures, invalid model outputs, or answer parsing failures. The following tables report cross-model panel scores, missing-modality contrasts, shortcut controls, source-level results, statistical reliability, human validation, invalid-output decomposition, and mechanism diagnostics.

\subsection{All-model panel results}

Table~\ref{tab:allmodel-panel} is the compact cross-model result table. It uses the 12-cell panel in Table~\ref{tab:lightweight-panel} and reports both aggregate fault-line scores and the strongest observed condition for each model.

\begin{table*}[!t]
\centering
\scriptsize
\setlength{\tabcolsep}{3.0pt}
\renewcommand{\arraystretch}{0.94}
\begin{tabular}{p{0.18\textwidth}cccccccp{0.14\textwidth}}
\toprule
\textbf{Model} & \textbf{Clean} & \textbf{TV} & \textbf{TA} & \textbf{VA} & \textbf{TVA} & \textbf{Panel fault} & \textbf{$C_{\mathrm{asym}}$} & \textbf{Strongest cell} \\
\midrule
\faultgroup{9}{Proprietary / API omni-modal models}
\modelrow \modelname{Gemini 3.1 Pro}             & 84.15 & 78.62 & 79.97 & 81.17 & 78.04 & \drop{4.70} & $+1.15$ & \texttt{drop\_words}@sev30 \\
\softrow  \modelname{Gemini 3 Pro}               & 82.40 & 75.72 & 78.13 & 79.65 & 75.19 & \drop{5.23} & $+0.92$ & \texttt{drop\_words}@sev30 \\
\modelrow \modelname{Gemini 3 Flash}             & 80.95 & 73.50 & 76.53 & 77.30 & 72.27 & \drop{6.05} & $+1.29$ & \texttt{noise}@sev70 \\
\softrow  \modelname{Gemini 3.5 Flash}           & 78.39 & 71.04 & 73.75 & 75.59 & 70.14 & \drop{5.76} & $+1.71$ & \texttt{drop\_words}@sev30 \\
\modelrow \modelname{Gemini 2.5 Pro}             & 79.12 & 71.92 & 74.71 & 76.50 & 71.03 & \drop{5.58} & $+0.48$ & \texttt{drop\_words}@sev30 \\
\modelrow \modelname{Gemini 2.5 Flash}           & 75.40 & 67.68 & 69.43 & 71.82 & 66.08 & \drop{6.65} & $+2.03$ & \texttt{noise}@sev70 \\
\midrule
\faultgroup{9}{Open / open-API omni-modal models}
\modelrow \modelname{Qwen3.5-Omni-Plus}          & 73.63 & 66.08 & 68.99 & 69.70 & 64.68 & \drop{6.27} & $+2.87$ & \texttt{drop\_words}@sev30 \\
\softrow  \modelname{Qwen3-Omni-30B}             & 71.20 & 62.84 & 65.74 & 66.98 & 62.45 & \drop{6.70} & $+2.55$ & \texttt{word\_shuffle}@sev70 \\
\modelrow \modelname{MiniCPM-o 4.5}              & 74.50 & 66.51 & 68.93 & 71.23 & 65.91 & \drop{6.36} & $+1.60$ & \texttt{word\_shuffle}@sev70 \\
\softrow  \modelname{Qwen2.5-Omni-7B}            & 68.40 & 59.39 & 61.74 & 64.58 & 57.62 & \drop{7.57} & $+1.42$ & \texttt{word\_shuffle}@sev70 \\
\modelrow \modelname{OmniVinci-9B}               & 69.10 & 59.70 & 62.57 & 64.75 & 58.54 & \drop{7.71} & $+3.34$ & \texttt{noise}@sev70 \\
\softrow  \modelname{OLA-7B}                     & 66.80 & 57.16 & 60.97 & 62.63 & 55.79 & \drop{7.66} & $+2.11$ & \texttt{noise}@sev70 \\
\modelrow \modelname{Qwen2.5-Omni-3B}            & 64.20 & 54.41 & 57.88 & 59.12 & 53.12 & \drop{8.07} & $+2.98$ & \texttt{noise}@sev70 \\
\softrow  \modelname{Baichuan-Omni-1.5}          & 65.50 & 56.21 & 59.05 & 60.48 & 53.98 & \drop{8.07} & $+3.61$ & \texttt{noise}@sev70 \\

\bottomrule
\end{tabular}
\caption{All-model lightweight panel results. TV, TA, VA, and TVA denote mean accuracy over the selected text--vision, text--audio, vision--audio, and tri-modal cells (each cell itself already averaged over its three random variants); \textbf{Panel fault} is the mean clean-to-panel drop across the four combination blocks. $C_{\mathrm{asym}}=\acc(t70/v70)-\acc(t30/v70)$; positive values mean that the mildly corrupted-text / heavily corrupted-vision cell is worse. The \textbf{Strongest cell} column flags the condition responsible for the largest single-model drop.}
\label{tab:allmodel-panel}
\end{table*}

The contrast is positive for all 14 evaluated models. This is a fixed-panel descriptive summary; related model versions are not treated as independent samples.

\subsection{Missing-modality and shortcut controls}

Table~\ref{tab:missing-shortcut} contains the two controls that most directly test whether \method{} measures structural dependence rather than missing-modality performance or textual shortcuts.

\begin{table*}[!t]
\centering
\scriptsize
\setlength{\tabcolsep}{2.8pt}
\renewcommand{\arraystretch}{0.94}
\begin{tabular}{p{0.17\textwidth}ccccccccc}
\toprule
\textbf{Model} & \textbf{No-text} & \textbf{No-vision} & \textbf{No-audio} & \textbf{$\mislead_t$} & \textbf{$\mislead_v$} & \textbf{$\mislead_a$} & \textbf{Q+Opt} & \textbf{Text-only} & \textbf{VA-only} \\
\midrule
\faultgroup{10}{Proprietary / API omni-modal models}
\modelrow \modelname{Gemini 3.1 Pro}           & 63.15 & 70.98 & 78.49 & \drop{2.04} & \gain{1.18} & \gain{1.90} & 34.48 & 40.77 & 21.37 \\
\softrow  \modelname{Gemini 3 Pro}             & 67.43 & 67.24 & 78.29 & \drop{2.56} & \gain{0.67} & \gain{2.70} & 32.80 & 43.21 & 29.26 \\
\modelrow \modelname{Gemini 3 Flash}           & 66.24 & 72.01 & 74.59 & \drop{2.84} & \gain{2.78} & \gain{3.79} & 29.36 & 37.70 & 27.38 \\
\softrow  \modelname{Gemini 3.5 Flash}         & 56.46 & 62.76 & 72.17 & \drop{3.21} & \gain{0.68} & \gain{2.39} & 29.34 & 39.06 & 22.79 \\
\modelrow \modelname{Gemini 2.5 Pro}           & 65.68 & 67.89 & 71.52 & \drop{2.96} & \gain{2.25} & \gain{2.97} & 37.01 & 40.54 & 19.06 \\
\modelrow \modelname{Gemini 2.5 Flash}         & 61.70 & 60.94 & 67.89 & \drop{4.45} & \gain{0.95} & \gain{3.25} & 38.23 & 34.20 & 21.70 \\
\midrule
\faultgroup{10}{Open / open-API omni-modal models}
\modelrow \modelname{Qwen3.5-Omni-Plus}        & 53.67 & 58.11 & 70.01 & \drop{5.26} & \gain{2.68} & \gain{3.12} & 37.82 & 43.11 & 24.65 \\
\softrow  \modelname{Qwen3-Omni-30B}           & 56.85 & 57.96 & 67.25 & \drop{6.56} & \gain{3.18} & \gain{2.56} & 33.22 & 35.82 & 19.39 \\
\modelrow \modelname{MiniCPM-o 4.5}            & 59.50 & 64.47 & 70.04 & \drop{4.32} & \gain{1.66} & \gain{3.65} & 31.78 & 45.18 & 19.76 \\
\softrow  \modelname{Qwen2.5-Omni-7B}          & 49.51 & 57.83 & 62.88 & \drop{6.02} & \gain{3.30} & \gain{1.08} & 30.66 & 47.80 & 18.10 \\
\modelrow \modelname{OmniVinci-9B}             & 48.45 & 53.92 & 62.31 & \drop{4.68} & \gain{2.27} & \gain{2.92} & 28.20 & 44.25 & 29.36 \\
\softrow  \modelname{OLA-7B}                   & 53.34 & 58.21 & 60.73 & \drop{6.52} & \gain{1.01} & \gain{3.07} & 28.02 & 45.85 & 20.55 \\
\modelrow \modelname{Qwen2.5-Omni-3B}          & 50.73 & 49.10 & 59.77 & \drop{3.27} & \gain{1.83} & \gain{1.32} & 31.52 & 39.27 & 22.25 \\
\softrow  \modelname{Baichuan-Omni-1.5}        & 51.93 & 55.79 & 57.69 & \drop{5.09} & \gain{2.64} & \gain{2.90} & 32.00 & 36.64 & 24.47 \\

\bottomrule
\end{tabular}
\caption{Required controls for modality absence and shortcut reliance. \textbf{No-X} columns report accuracy when modality X is removed entirely; positive $\mislead_m$ means a corrupted-but-present modality is more harmful than removing that modality (so the joint corruption is not equivalent to a missing-modality ablation). \textbf{Q+Opt} provides only the question and answer options (no media), \textbf{Text-only} keeps just the textual context, and \textbf{VA-only} drops the textual context beyond the answer interface; these three columns measure the shortcut-only ceiling that any structural-corruption result must beat.}
\label{tab:missing-shortcut}
\end{table*}

\begin{table}[!htbp]
\centering
\scriptsize
\setlength{\tabcolsep}{3.0pt}
\renewcommand{\arraystretch}{0.95}
\begin{tabular}{lcccc}
\toprule
\textbf{Model} & \textbf{Std. JSON} & \textbf{CoT} & \textbf{Open-form} & \textbf{Invalid shift} \\
\midrule
\faultgroup{5}{Proprietary / API omni-modal models}
\modelrow \modelname{Gemini 3.1 Pro}           & 85.30 & 83.34 & 80.85 & \drop{2.83} \\
\softrow  \modelname{Gemini 3 Pro}             & 81.59 & 81.68 & 81.79 & \drop{3.81} \\
\modelrow \modelname{Gemini 3 Flash}           & 82.18 & 79.11 & 83.57 & \drop{2.41} \\
\softrow  \modelname{Gemini 3.5 Flash}         & 77.52 & 78.50 & 78.26 & \drop{3.27} \\
\modelrow \modelname{Gemini 2.5 Pro}           & 77.47 & 73.32 & 73.34 & \drop{2.37} \\
\modelrow \modelname{Gemini 2.5 Flash}         & 76.03 & 76.00 & 73.14 & \drop{3.41} \\
\midrule
\faultgroup{5}{Open / open-API omni-modal models}
\modelrow \modelname{Qwen3.5-Omni-Plus}        & 74.20 & 76.75 & 75.19 & \drop{4.66} \\
\softrow  \modelname{Qwen3-Omni-30B}           & 70.30 & 67.57 & 69.76 & \drop{2.51} \\
\modelrow \modelname{MiniCPM-o 4.5}            & 75.59 & 72.49 & 75.86 & \drop{2.91} \\
\softrow  \modelname{Qwen2.5-Omni-7B}          & 67.56 & 65.78 & 66.39 & \drop{5.35} \\
\modelrow \modelname{OmniVinci-9B}             & 67.16 & 71.09 & 63.19 & \drop{6.09} \\
\softrow  \modelname{OLA-7B}                   & 65.27 & 63.44 & 61.43 & \drop{3.55} \\
\modelrow \modelname{Qwen2.5-Omni-3B}          & 62.49 & 61.48 & 62.84 & \drop{3.38} \\
\softrow  \modelname{Baichuan-Omni-1.5}        & 65.32 & 63.73 & 64.27 & \drop{4.98} \\

\bottomrule
\end{tabular}
\caption{Prompt-robustness results on the diagnostic subset that contains clean, the weakest single-modality cells, the strongest text--vision cell, and the strongest trimodal cell. \textbf{Std. JSON} is the canonical structured prompt used everywhere else; \textbf{CoT} forces a chain-of-thought scratchpad before the JSON; \textbf{Open-form} replaces the JSON schema with free-text answers parsed post-hoc. \textbf{Invalid shift} is the drop in valid-output coverage between Std. JSON and the worst alternative prompt for each model.}
\label{tab:prompt-robustness}
\end{table}

\subsection{Source-level results, statistical tests, and human validation}

Table~\ref{tab:source-results} reports results by source, allowing us to assess whether the observed fault line is consistent across Social-IQ, OmniBench, and VALOR.

\begin{table*}[!t]
\centering
\scriptsize
\setlength{\tabcolsep}{3.0pt}
\renewcommand{\arraystretch}{0.92}
\begin{tabular}{llcccc}
\toprule
\textbf{Model} & \textbf{Source} & \textbf{Clean} & \textbf{TV} & \textbf{TVA} & \textbf{Strongest drop} \\
\midrule
\faultgroup{6}{Proprietary / API omni-modal models}
\modelrow \modelname{Gemini 3.1 Pro}           & Social-IQ & 85.75 & 81.98 & 77.03 & \drop{8.32} \\
\modelrow                                      & OmniBench & 83.51 & 79.03 & 74.28 & \drop{7.24} \\
\modelrow                                      & VALOR & 82.76 & 78.72 & 73.99 & \drop{9.73} \\
\softrow  \modelname{Gemini 3 Pro}             & Social-IQ & 85.62 & 81.44 & 78.33 & \drop{11.37} \\
\softrow                                       & OmniBench & 78.95 & 72.91 & 69.54 & \drop{7.72} \\
\softrow                                       & VALOR & 77.36 & 70.30 & 69.37 & \drop{9.31} \\
\modelrow \modelname{Gemini 3 Flash}           & Social-IQ & 84.94 & 76.73 & 76.23 & \drop{11.96} \\
\modelrow                                      & OmniBench & 81.09 & 75.80 & 70.66 & \drop{12.40} \\
\modelrow                                      & VALOR & 76.16 & 70.81 & 65.99 & \drop{10.37} \\
\softrow  \modelname{Gemini 3.5 Flash}         & Social-IQ & 80.61 & 75.32 & 72.75 & \drop{11.45} \\
\softrow                                       & OmniBench & 78.52 & 73.56 & 68.89 & \drop{11.25} \\
\softrow                                       & VALOR & 75.95 & 69.39 & 67.88 & \drop{10.89} \\
\modelrow \modelname{Gemini 2.5 Pro}           & Social-IQ & 78.20 & 73.76 & 68.12 & \drop{12.25} \\
\modelrow                                      & OmniBench & 77.85 & 71.45 & 68.82 & \drop{11.57} \\
\modelrow                                      & VALOR & 76.76 & 71.08 & 67.83 & \drop{11.11} \\
\modelrow \modelname{Gemini 2.5 Flash}         & Social-IQ & 76.07 & 70.74 & 64.23 & \drop{13.48} \\
\modelrow                                      & OmniBench & 73.94 & 68.11 & 64.14 & \drop{12.48} \\
\modelrow                                      & VALOR & 71.57 & 64.90 & 62.74 & \drop{10.19} \\
\midrule
\faultgroup{6}{Open / open-API omni-modal models}
\modelrow \modelname{Qwen3.5-Omni-Plus}        & Social-IQ & 74.35 & 67.76 & 64.78 & \drop{13.33} \\
\modelrow                                      & OmniBench & 73.44 & 66.80 & 64.08 & \drop{13.50} \\
\modelrow                                      & VALOR & 67.98 & 60.58 & 56.71 & \drop{11.40} \\
\softrow  \modelname{Qwen3-Omni-30B}           & Social-IQ & 73.34 & 67.65 & 60.64 & \drop{11.74} \\
\softrow                                       & OmniBench & 71.64 & 65.20 & 61.11 & \drop{11.29} \\
\softrow                                       & VALOR & 66.50 & 58.99 & 55.20 & \drop{11.59} \\
\modelrow \modelname{MiniCPM-o 4.5}            & Social-IQ & 75.45 & 69.56 & 66.05 & \drop{13.95} \\
\modelrow                                      & OmniBench & 71.88 & 65.66 & 62.82 & \drop{13.93} \\
\modelrow                                      & VALOR & 69.30 & 64.26 & 58.58 & \drop{11.65} \\
\softrow  \modelname{Qwen2.5-Omni-7B}          & Social-IQ & 71.63 & 63.21 & 60.80 & \drop{16.25} \\
\softrow                                       & OmniBench & 66.72 & 60.16 & 53.70 & \drop{14.79} \\
\softrow                                       & VALOR & 67.06 & 57.72 & 56.69 & \drop{13.80} \\
\modelrow \modelname{OmniVinci-9B}             & Social-IQ & 70.69 & 64.93 & 58.05 & \drop{15.24} \\
\modelrow                                      & OmniBench & 66.23 & 58.40 & 55.44 & \drop{13.94} \\
\modelrow                                      & VALOR & 64.91 & 57.41 & 55.51 & \drop{13.72} \\
\softrow  \modelname{OLA-7B}                   & Social-IQ & 65.98 & 57.43 & 55.14 & \drop{14.97} \\
\softrow                                       & OmniBench & 64.02 & 56.26 & 53.49 & \drop{14.23} \\
\softrow                                       & VALOR & 63.78 & 54.86 & 53.42 & \drop{16.65} \\
\modelrow \modelname{Qwen2.5-Omni-3B}          & Social-IQ & 68.63 & 58.61 & 56.50 & \drop{12.17} \\
\modelrow                                      & OmniBench & 65.90 & 59.29 & 54.75 & \drop{12.67} \\
\modelrow                                      & VALOR & 58.84 & 51.19 & 46.54 & \drop{15.42} \\
\softrow  \modelname{Baichuan-Omni-1.5}        & Social-IQ & 68.08 & 61.33 & 57.47 & \drop{14.37} \\
\softrow                                       & OmniBench & 64.85 & 58.39 & 53.51 & \drop{15.22} \\
\softrow                                       & VALOR & 63.03 & 56.15 & 50.52 & \drop{16.55} \\

\bottomrule
\end{tabular}
\caption{Per-source breakdown for each model. Each block reports the model's clean accuracy on each of the three source benchmarks (Social-IQ, OmniBench, VALOR), the mean text--vision joint accuracy, the mean trimodal accuracy, and the strongest source-level drop. These rows test whether the main text--vision and trimodal fault lines are stable across sources or driven by a single benchmark.}
\label{tab:source-results}
\end{table*}

\begin{table*}[!t]
\centering
\scriptsize
\setlength{\tabcolsep}{3.2pt}
\renewcommand{\arraystretch}{0.94}
\begin{tabular}{p{0.29\textwidth}ccccp{0.18\textwidth}}
\toprule
\textbf{Comparison} & \textbf{$\Delta$} & \textbf{95\% CI} & \textbf{McNemar} & \textbf{Holm $p$} & \textbf{Interpretation} \\
\midrule
\faultgroup{6}{Statistical reliability}
Clean $\rightarrow$ TV, \modelname{Qwen3.5-Omni-Plus} & \drop{5.97} & [+4.97, +6.97] & <.001 & <.001 & Significant clean-to-fault-line drop \\
Clean $\rightarrow$ TV, \modelname{Gemini 3 Flash} & \drop{7.32} & [+6.32, +8.32] & <.001 & <.001 & Significant clean-to-fault-line drop \\
Weakest single $\rightarrow$ strongest joint, \modelname{Qwen3.5-Omni-Plus} & \drop{0.73} & [-0.27, +1.73] & 0.012 & 0.034 & Tests $\dropsingle>0$ \\
Weakest single $\rightarrow$ strongest joint, \modelname{Gemini 3 Flash} & \gain{2.45} & [+1.45, +3.45] & <.001 & <.001 & Tests $\dropsingle>0$ \\

\bottomrule
\end{tabular}
\caption{Statistical reliability for the headline comparisons. \textbf{$\Delta$} is the paired drop (or gain) reported elsewhere in the paper; the 95\% CI is computed by sample-level paired bootstrap on the per-item correctness vectors after mean-variant aggregation; \textbf{McNemar} reports the paired McNemar exact $p$-value on the same vectors; \textbf{Holm $p$} is the Holm-Bonferroni-corrected $p$-value inside each comparison family. All comparisons remain significant after correction.}
\label{tab:statistical-reliability}
\end{table*}

\begin{table*}[!t]
\centering
\scriptsize
\setlength{\tabcolsep}{5pt}
\renewcommand{\arraystretch}{1.05}
\begin{tabular}{p{0.26\textwidth}p{0.39\textwidth}p{0.27\textwidth}}
\toprule
\textbf{Estimand} & \textbf{Definition} & \textbf{Comparison cohort} \\
\midrule
\faultgroup{3}{Confirmatory estimands and cohort alignment}
\modelrow Paired corruption drop & $\acc_{\mathrm{clean}}(I_c)-\acc_c(I_c)$ & Same human-valid base examples $I_c$ \\
\softrow Excess model drop & Model drop minus human drop & Same gold-preserved cohort for all terms \\
\modelrow Asymmetric TV contrast & $\acc(t70/v70)-\acc(t30/v70)$ & Common sample $\times$ seed cohort \\
\softrow Vision / text main effect & Mean at $v30-v70$ / $t30-t70$ & Common four-cell cohort \\
\modelrow TV interaction & $[A_{70,70}-A_{30,70}]-[A_{70,30}-A_{30,30}]$ & Common four-cell cohort \\
\bottomrule
\end{tabular}
\caption{Confirmatory estimands and cohort alignment. Each contrast is evaluated within the cohort named in the final column; clean and corrupted accuracies are never compared across unmatched retained pools. Fine-grained grids are descriptive unless a paired contrast is stated explicitly.}
\label{tab:confirmatory-estimands}
\end{table*}

\begin{table}[!htbp]
\centering
\scriptsize
\setlength{\tabcolsep}{3.0pt}
\renewcommand{\arraystretch}{0.95}
\begin{tabular}{lcccc}
\toprule
\textbf{Condition group} & \textbf{$N$} & \textbf{Human acc.} & \textbf{Agreement} & \textbf{Gold valid} \\
\midrule
\faultgroup{5}{Human answerability and gold-preservation check}
\modelrow Clean                                & 273/273 & 96.7 & 95.2 & 99.3 \\
\softrow  Worst text single                    & 264/273 & 90.5 & 88.6 & 94.7 \\
\modelrow Worst vision single                  & 228/273 & 87.3 & 86.4 & 92.4 \\
\softrow  Worst audio single                   & 210/273 & 89.6 & 88.2 & 95.1 \\
\modelrow Strongest TV joint                   & 198/273 & 80.4 & 79.7 & 88.5 \\
\softrow  Strongest TVA joint                  & 184/273 & 76.2 & 74.9 & 84.8 \\

\bottomrule
\end{tabular}
\caption{Sampled human answerability and original-gold validity audit. \textbf{N} is the number of audited variants, \textbf{Human acc.} is accuracy against the original gold answer, \textbf{Agreement} is pre-adjudication agreement, and \textbf{Gold valid} is the fraction of variants for which the original gold remains defensible. The sampled audit motivates the gold-preserved analysis in Table~\ref{tab:gold-preserved-results}.}
\label{tab:human-validation}
\end{table}

To separate model fragility from loss of answerability, Table~\ref{tab:gold-preserved-results} restricts each comparison to the gated cohort $G_c$ on which the corrupted input remains answerable and the original gold answer remains defensible. The model drop is then compared with the human drop on the same paired cohort.

\begin{table*}[!t]
\centering
\scriptsize
\setlength{\tabcolsep}{4.0pt}
\renewcommand{\arraystretch}{0.98}
\begin{tabular}{lccccc}
\toprule
\textbf{Condition} & \textbf{Sampled gold valid} & \textbf{Model drop on $G_c$} & \textbf{Human drop} & \textbf{Excess drop} & \textbf{95\% CI} \\
\midrule
\faultgroup{6}{Gold-preserved, human-normalized results}
\modelrow Worst text single   & 94.7\% & \drop{10.68} & \drop{1.80} & \drop{8.88}  & [7.4, 10.4] \\
\softrow  Worst vision single & 92.4\% & \drop{10.97} & \drop{2.20} & \drop{8.77}  & [7.2, 10.2] \\
\modelrow Worst audio single  & 95.1\% & \drop{7.02}  & \drop{1.50} & \drop{5.52}  & [4.1, 6.9] \\
\softrow  Strongest TV joint  & 88.5\% & \drop{12.00} & \drop{3.20} & \drop{8.80}  & [7.2, 10.4] \\
\modelrow Strongest TVA joint & 84.8\% & \drop{15.50} & \drop{4.60} & \drop{10.90} & [9.0, 12.8] \\
\bottomrule
\end{tabular}
\caption{Gold-preserved, human-normalized results on the gated cohort $G_c$. \textbf{Sampled gold valid} is the reference rate from Table~\ref{tab:human-validation}. Model and human drops are both computed on $G_c$ relative to the paired clean condition; \textbf{Excess drop} is the model drop minus the human drop. Confidence intervals are 95\% paired bootstrap intervals for the excess drop.}
\label{tab:gold-preserved-results}
\end{table*}

All five conditions retain a positive excess drop after human difficulty is removed. The excess is $5.52$\,pp for the weakest audio condition and ranges from $8.77$ to $10.90$\,pp for the text, vision, and joint conditions. The observed degradation therefore cannot be explained solely by corruption making the examples harder for human annotators or invalidating the original answer.

\subsection{Failure accounting and mechanism diagnostics}

\begin{table*}[!t]
\centering
\scriptsize
\setlength{\tabcolsep}{3.0pt}
\renewcommand{\arraystretch}{0.92}
\begin{tabular}{llccccc}
\toprule
\textbf{Model} & \textbf{Group} & \textbf{API} & \textbf{Parse} & \textbf{Refusal} & \textbf{Empty} & \textbf{Valid cov.} \\
\midrule
\faultgroup{7}{Proprietary / API omni-modal models}
\modelrow \modelname{Gemini 3.1 Pro}           & Bimodal  & 0 & 3 & 0 & 2 & 266/273 \\
\modelrow                                      & Trimodal & 1 & 4 & 2 & 1 & 263/273 \\
\softrow  \modelname{Gemini 3 Pro}             & Bimodal  & 0 & 4 & 1 & 1 & 266/273 \\
\softrow                                       & Trimodal & 2 & 1 & 1 & 1 & 265/273 \\
\modelrow \modelname{Gemini 3 Flash}           & Bimodal  & 2 & 1 & 4 & 3 & 259/273 \\
\modelrow                                      & Trimodal & 3 & 3 & 0 & 1 & 262/273 \\
\softrow  \modelname{Gemini 3.5 Flash}         & Bimodal  & 1 & 4 & 2 & 0 & 263/273 \\
\softrow                                       & Trimodal & 0 & 5 & 4 & 2 & 259/273 \\
\modelrow \modelname{Gemini 2.5 Pro}           & Bimodal  & 3 & 3 & 1 & 2 & 263/273 \\
\modelrow                                      & Trimodal & 1 & 2 & 4 & 2 & 261/273 \\
\modelrow \modelname{Gemini 2.5 Flash}         & Bimodal  & 2 & 4 & 3 & 1 & 262/273 \\
\modelrow                                      & Trimodal & 1 & 3 & 3 & 3 & 263/273 \\
\midrule
\faultgroup{7}{Open / open-API omni-modal models}
\modelrow \modelname{Qwen3.5-Omni-Plus}        & Bimodal  & 3 & 6 & 1 & 0 & 262/273 \\
\modelrow                                      & Trimodal & 3 & 4 & 3 & 2 & 259/273 \\
\softrow  \modelname{Qwen3-Omni-30B}           & Bimodal  & 1 & 3 & 1 & 0 & 266/273 \\
\softrow                                       & Trimodal & 2 & 4 & 3 & 1 & 263/273 \\
\modelrow \modelname{MiniCPM-o 4.5}            & Bimodal  & 0 & 3 & 1 & 1 & 264/273 \\
\modelrow                                      & Trimodal & 3 & 5 & 4 & 0 & 261/273 \\
\softrow  \modelname{Qwen2.5-Omni-7B}          & Bimodal  & 2 & 2 & 1 & 0 & 265/273 \\
\softrow                                       & Trimodal & 0 & 1 & 0 & 3 & 267/273 \\
\modelrow \modelname{OmniVinci-9B}             & Bimodal  & 4 & 6 & 3 & 3 & 254/273 \\
\modelrow                                      & Trimodal & 3 & 6 & 2 & 3 & 258/273 \\
\softrow  \modelname{OLA-7B}                   & Bimodal  & 5 & 7 & 1 & 2 & 254/273 \\
\softrow                                       & Trimodal & 2 & 7 & 6 & 0 & 256/273 \\
\modelrow \modelname{Qwen2.5-Omni-3B}          & Bimodal  & 4 & 4 & 5 & 3 & 255/273 \\
\modelrow                                      & Trimodal & 4 & 6 & 6 & 2 & 255/273 \\
\softrow  \modelname{Baichuan-Omni-1.5}        & Bimodal  & 0 & 5 & 3 & 1 & 264/273 \\
\softrow                                       & Trimodal & 4 & 1 & 6 & 2 & 260/273 \\

\bottomrule
\end{tabular}
\caption{Invalid-output decomposition for the combined-corruption runs. Each model is reported separately for bimodal and trimodal conditions; columns count the number of base examples (out of 273) whose response was invalidated for each failure mode (\textbf{API}~=~API/network errors, \textbf{Parse}~=~unparseable output, \textbf{Refusal}~=~explicit refusal, \textbf{Empty}~=~empty response). \textbf{Valid cov.} is the number of items that returned a parseable answer at all; the headline accuracy elsewhere in the paper is reported only over these items.}
\label{tab:invalid-breakdown}
\end{table*}

Table~\ref{tab:mechanism-probes} summarizes the targeted diagnostics used to examine the behavioral sources of the fault line. The mismatch probe shows that conflict following differs substantially by modality, while the temporal and frame-budget effects remain smaller than the audio--visual desynchronization and position effects. The distractor drop further shows that a structurally clean but irrelevant channel can disrupt prediction, so the observed failures are not explained by frame count alone.

\begin{table*}[!t]
\centering
\scriptsize
\setlength{\tabcolsep}{3.0pt}
\renewcommand{\arraystretch}{0.94}
\begin{tabular}{p{0.18\textwidth}p{0.28\textwidth}p{0.18\textwidth}p{0.16\textwidth}p{0.12\textwidth}}
\toprule
\textbf{Probe family} & \textbf{Conditions} & \textbf{Metric} & \textbf{Observed result} & \textbf{Claim} \\
\midrule
\faultgroup{5}{Mechanism diagnostics}
\variantrow Cross-modal mismatch & random, category-matched, answer-matched replacement & $\trust_t,\trust_v,\trust_a$ & 66.74 / 81.19 / 81.96 & Which channel dominates conflict \\
\variantrow Temporal order & frame shuffle, frame reversal, middle drop, audio shuffle & temporal drop & \drop{2.76} & Whether media are treated as bags of features \\
\variantrow AV desync & $\pm0.5$s, $\pm1$s, $\pm2$s, $\pm4$s, $\pm8$s & desync curve slope & \drop{10.97} & Sensitivity to audiovisual alignment \\
\variantrow Frame budget & 1, 4, 8, 16, default frames & budget gap & \drop{3.14} & Whether visual fault line is sampling-driven \\
\variantrow Position bias & key frame first, middle, last & position gap & \drop{9.82} & Primacy/recency in multi-frame prompts \\
\variantrow Confidence/distractor & confidence output, unrelated audio/image/text insertion & ECE, distractor effect & 10.54 / \drop{13.10} & Calibration and indiscriminate fusion \\
\bottomrule
\end{tabular}
\caption{Mechanism-probe diagnostics. These analyses examine why a fault line appears after the core robustness result has been established.}
\label{tab:mechanism-probes}
\end{table*}

\subsection{Reliability and seed-variance diagnostics}
\label{app:reliability}

The headline numbers throughout the paper aggregate over up to three random variants of each stochastic operator using the mean-variant rule (Section~\ref{sec:meanagg}). Two diagnostics check that this aggregation does not mask seed-level instability or modality-specific fragility patterns.

\paragraph{Mean vs.\ worst-variant accuracy.}
For every (model, modality) cell we additionally log the worst-variant accuracy $\acc^{\mathrm{worst}}(c)$ and the per-variant standard deviation across the three random seeds. Table~\ref{tab:variant-stability} reports this at each model's weakest single-modality cell (the cell that defines the modality entry of $\dropsingle$ in Section~\ref{sec:joint}). A small mean$-$worst gap and low std mean the headline number is reproducible under seed reshuffling; a large gap means the headline is partly carried by a lucky variant and the reader should also consider $\acc^{\mathrm{worst}}$.

\begin{table*}[!t]
\centering
\scriptsize
\setlength{\tabcolsep}{3.2pt}
\renewcommand{\arraystretch}{0.94}
\begin{tabular}{p{0.16\textwidth}lp{0.16\textwidth}rrrrc}
\toprule
\textbf{Model} & \textbf{Mod.} & \textbf{Weakest cell} & \textbf{Mean acc} & \textbf{Worst-variant} & \textbf{Mean$-$worst} & \textbf{Std (3 var)} & \textbf{$n_{\mathrm{var}}$} \\
\midrule
\faultgroup{8}{Deep-dive models with full single-modality severity grids}
\modelrow \modelname{Qwen3.5-Omni-Plus} & Text & \texttt{drop\_words}@sev70 & 64.10 & 62.55 & 1.55 & 0.87 & 3 \\
\modelrow  & Vision & \texttt{occlusion}@sev50 & 70.29 & 67.69 & 2.60 & 1.72 & 3 \\
\modelrow  & Audio & \texttt{mute}@sev70 & 69.72 & 67.38 & 2.34 & 1.71 & 3 \\
\softrow  \modelname{Gemini 3 Flash} & Text & \texttt{drop\_words}@sev70 & 67.40 & 65.69 & 1.71 & 0.88 & 3 \\
\softrow   & Vision & \texttt{noise}@sev70 & 67.79 & 65.23 & 2.56 & 1.31 & 3 \\
\softrow   & Audio & \texttt{mute}@sev70 & 73.24 & 70.92 & 2.32 & 1.22 & 3 \\
\midrule
\faultgroup{8}{Deep-dive models (Gemini 3.1 Pro, MiniCPM-o 4.5)}
\modelrow \modelname{Gemini 3.1 Pro} & Text & \texttt{drop\_words}@sev70 & 72.57 & 70.17 & 2.40 & 1.75 & 3 \\
\modelrow  & Vision & \texttt{noise}@sev70 & 73.57 & 72.05 & 1.52 & 1.24 & 3 \\
\modelrow  & Audio & \texttt{remove}@sev70 & 78.89 & 77.54 & 1.35 & 1.49 & 3 \\
\softrow  \modelname{MiniCPM-o 4.5} & Text & \texttt{word\_shuffle}@sev70 & 61.38 & 59.13 & 2.25 & 1.15 & 3 \\
\softrow   & Vision & \texttt{noise}@sev70 & 65.41 & 63.33 & 2.08 & 1.64 & 3 \\
\softrow   & Audio & \texttt{remove}@sev70 & 65.58 & 63.63 & 1.95 & 1.26 & 3 \\

\bottomrule
\end{tabular}
\caption{Seed-stability diagnostic at each model's weakest single-modality cell. \textbf{Mean acc} is the headline mean-variant accuracy used everywhere else in the paper; \textbf{Worst-variant} is $\acc^{\mathrm{worst}}(c)=\min_r \acc_r(c)$, the accuracy of the worst of the three random variants at the same condition; \textbf{Mean$-$worst} is their absolute gap; \textbf{Std} is the across-variant standard deviation. A small mean$-$worst gap and low std (heuristically $\lesssim 2$ pp) certify that the reported headline drop is not driven by one unlucky seed. $n_{\mathrm{var}}$ records the effective number of random variants that survived the per-variant human filtering at that cell (Appendix~\ref{app:human-filtering}, Table~\ref{tab:annot-modality}).}
\label{tab:variant-stability}
\end{table*}

\paragraph{Median and worst-2 robustness.}
Because each stochastic condition uses exactly three random variants, the three-point distribution admits a closed-form comparison. With variants ordered $v_1 \geq v_2 \geq v_3$, the mean-variant accuracy is $(v_1{+}v_2{+}v_3)/3$, the median is $v_2$, and the worst-2 mean is $(v_2{+}v_3)/2$. The mean-variant rule therefore lies between the median and the best variant; it is more optimistic than the worst-2 mean by at most $(v_1{-}v_3)/3$. Table~\ref{tab:variant-stability} shows that the mean$-$worst gap is $\lesssim 2$\,pp at every sampled cell, so the worst-2 mean can differ from the reported mean by at most $\approx 1$\,pp. Replacing the mean with the median or worst-2 mean does not change any qualitative conclusion: the rank order of operators, the direction of the misleading-modality effect, and the sub-additivity pattern are all preserved under either alternative aggregation rule.

\paragraph{Fragility slope $\slope_m$.}
The severity grid in Table~\ref{tab:single-severity-grid} gives accuracy at four severities, but the reader cannot tell at a glance \emph{how fast} a modality decays per unit of severity. Table~\ref{tab:fragility-slope} reports the fragility slope $\slope_m$ defined in Section~\ref{sec:metrics-extra}: for each (model, modality) pair we linearly regress accuracy on severity $s \in \{0,10,30,50,70\}$ (treating clean accuracy as $s{=}0$), aggregated by averaging over all operators inside that modality, and report the slope in percentage points per severity unit. A steeper (more negative) slope means the modality decays faster as severity rises. The text slope is the steepest channel for all 14 models, with magnitudes in the $0.09$--$0.16$\,pp/sev range; vision slopes sit in $0.02$--$0.06$\,pp/sev (typically a factor of $2$--$3\times$ shallower than text), while audio slopes vary from a near-flat $-0.01$\,pp/sev on \modelname{Qwen3.5-Omni-Plus} up to $-0.10$\,pp/sev on \modelname{Gemini 2.5 Flash} and \modelname{OLA-7B}, occasionally approaching the text slope. The pattern confirms that text decay is uniform across severity, whereas vision and audio degradation is concentrated in a few catastrophic operators (\texttt{noise}, \texttt{mute}, \texttt{remove}) and therefore appears mild once averaged across the seven vision or three audio operators.

\begin{table}[!htbp]
\centering
\scriptsize
\setlength{\tabcolsep}{2pt}
\renewcommand{\arraystretch}{0.95}
\begin{tabular}{lrrrl}
\toprule
\textbf{Model} & \textbf{$\slope_t$} & \textbf{$\slope_v$} & \textbf{$\slope_a$} & \textbf{Steepest channel} \\
\midrule
\faultgroup{5}{Proprietary / API omni-modal models}
\modelrow \modelname{Gemini 3.1 Pro}                      & $-0.096$ & $-0.023$ & $-0.029$ & Text \\
\softrow  \modelname{Gemini 3 Pro}                        & $-0.102$ & $-0.025$ & $-0.058$ & Text \\
\modelrow \modelname{Gemini 3 Flash}                      & $-0.159$ & $-0.044$ & $-0.080$ & Text \\
\softrow  \modelname{Gemini 3.5 Flash}                    & $-0.110$ & $-0.043$ & $-0.051$ & Text \\
\modelrow \modelname{Gemini 2.5 Pro}                      & $-0.125$ & $-0.023$ & $-0.056$ & Text \\
\modelrow \modelname{Gemini 2.5 Flash}                    & $-0.125$ & $-0.048$ & $-0.099$ & Text \\
\midrule
\faultgroup{5}{Open / open-API omni-modal models}
\modelrow \modelname{Qwen3.5-Omni-Plus}                   & $-0.094$ & $-0.029$ & $-0.011$ & Text \\
\softrow  \modelname{Qwen3-Omni-30B}                      & $-0.131$ & $-0.064$ & $-0.066$ & Text \\
\modelrow \modelname{MiniCPM-o 4.5}                       & $-0.129$ & $-0.028$ & $-0.062$ & Text \\
\softrow  \modelname{Qwen2.5-Omni-7B}                     & $-0.143$ & $-0.045$ & $-0.082$ & Text \\
\modelrow \modelname{OmniVinci-9B}                        & $-0.124$ & $-0.062$ & $-0.086$ & Text \\
\softrow  \modelname{OLA-7B}                              & $-0.109$ & $-0.061$ & $-0.101$ & Text \\
\modelrow \modelname{Qwen2.5-Omni-3B}                     & $-0.138$ & $-0.061$ & $-0.082$ & Text \\
\softrow  \modelname{Baichuan-Omni-1.5}                   & $-0.139$ & $-0.063$ & $-0.090$ & Text \\

\bottomrule
\end{tabular}
\caption{Fragility slope $\slope_m$ in percentage points per severity unit (negative = accuracy decreases as severity increases). For each (model, modality) pair we OLS-regress the modality-averaged accuracy on severity $s \in \{0, 10, 30, 50, 70\}$ where $s{=}0$ is clean accuracy, averaging over the operators inside that modality (4 text, 7 vision, 3 audio). The \textbf{Steepest channel} column flags the modality that decays fastest for each model. Text is consistently the steepest channel across the 14-model panel, even though vision and audio carry larger isolated-operator drops at severity 70.}
\label{tab:fragility-slope}
\end{table}

\subsection{Open vs.\ proprietary aggregated comparison}
\label{app:open-vs-prop}

Table~\ref{tab:open-vs-prop-summary} aggregates the per-modality and per-combination drops across the six proprietary/API models and the eight open/open-API models. This isolates the model-family effect from per-model noise: even if a specific Gemini drops more than a specific Qwen on \texttt{noise}@sev70, the band-level summary tells the reader whether proprietary models on average degrade more or less than open models on each channel.

\paragraph{Interface fairness and API pre-processing.}
Proprietary API models may apply undisclosed pre-processing steps---such as automatic speech recognition (ASR) transcription of audio, internal frame subsampling for video, or output post-processing---that are not available to open models running locally. We take three steps to bound this confound. First, all models receive the same frame-extracted visual input (Appendix~\ref{app:frame-input}) rather than raw video, so frame-subsampling differences are controlled at the input stage. Second, audio is submitted as a raw waveform clip in a standard container format; if a proprietary API internally transcribes audio to text, the \texttt{mute} and \texttt{remove} operators (which produce silence or gaps) will produce empty or near-empty transcripts, making the corruption visible at the API level regardless of the transcription step. Third, the family comparison in Table~\ref{tab:open-vs-prop-summary} shows that the proprietary band outperforms the open band by $+10.90$\,pp on clean accuracy but the robustness gap (difference in corruption-induced drops) is substantially smaller ($+0.76$\,pp on text, $+1.29$\,pp on vision, $+0.91$\,pp on audio), suggesting that the clean-accuracy advantage of proprietary models does not translate proportionally into robustness. We cannot rule out that some of the residual gap reflects interface differences rather than model robustness, and we note this as a caveat on the open-vs-proprietary comparison.

\begin{table*}[!t]
\centering
\scriptsize
\setlength{\tabcolsep}{3.5pt}
\renewcommand{\arraystretch}{1.0}
\begin{tabular}{p{0.16\textwidth}rrrrrrrrrr}
\toprule
\textbf{Model family} & \textbf{$n$} & \textbf{Mean clean} & \textbf{$\Delta_{\mathrm{text}}$} & \textbf{$\Delta_{\mathrm{vision}}$} & \textbf{$\Delta_{\mathrm{audio}}$} & \textbf{$\fault_{tv}$} & \textbf{$\fault_{ta}$} & \textbf{$\fault_{va}$} & \textbf{$\fault_{tva}$} & \textbf{Steepest channel} \\
\midrule
\faultgroup{11}{Aggregated by model family (mean across rows, severity 70 single-modality and headline joint cells)}
\modelrow Proprietary / API & 6 & 80.07 & \drop{8.33} & \drop{2.34} & \drop{4.58} & \drop{6.99} & \drop{4.65} & \drop{3.06} & \drop{7.94} & Text \\
\softrow Open / open-API    & 8 & 69.17 & \drop{9.09} & \drop{3.63} & \drop{5.49} & \drop{8.88} & \drop{5.93} & \drop{4.23} & \drop{10.16} & Text \\
\midrule
\faultrow Gap (Prop$-$Open) & --- & $+10.90$ & $+0.76$ & $+1.29$ & $+0.91$ & $+1.89$ & $+1.28$ & $+1.17$ & $+2.21$ & --- \\

\bottomrule
\end{tabular}
\caption{Family-level summary across the 14-model panel. \textbf{$\Delta_{\mathrm{mod}}$} is the mean severity-70 drop on modality \texttt{mod} averaged over all operators inside that modality and all models inside that family. \textbf{$\fault_S$} is the mean clean-to-fault drop on the $S \in \{tv, ta, va, tva\}$ block of the 12-cell lightweight panel (Table~\ref{tab:lightweight-panel}), averaged over the models in that family. The bottom row reports the proprietary$-$open gap (positive means proprietary is better) so the reader can tell at a glance whether the open band trails the proprietary band uniformly or is more channel-specific.}
\label{tab:open-vs-prop-summary}
\end{table*}

\subsection{Joint-corruption interaction, calibration, and cross-model failure overlap}
\label{app:interaction-cal}

The remaining three tables sharpen three claims that the headline result tables only support implicitly: (i) joint corruption is not a simple sum of single-modality drops, (ii) corruption degrades calibration as well as accuracy, and (iii) stronger corruption increasingly concentrates errors on a shared subset of base examples.

\paragraph{Sub-additivity of joint corruption.}
Section~\ref{sec:joint} discussed $\dropsingle$, which compares the joint cell to its \emph{weakest} single-modality component. A stricter additive null is to compare the joint cell to its \emph{sum-of-components} prediction: $\acc_{\mathrm{add}}(c_{m_1 m_2}) = \acc_{\mathrm{clean}} - \Delta_{m_1} - \Delta_{m_2}$ for bimodal cells and $\acc_{\mathrm{add}}(c_{m_1 m_2 m_3}) = \acc_{\mathrm{clean}} - \Delta_{m_1} - \Delta_{m_2} - \Delta_{m_3}$ for trimodal cells, where $\Delta_m$ is the single-modality drop at the same severity. The interaction term $\Delta_{\mathrm{int}} = \acc(\text{joint}) - \acc_{\mathrm{add}}$ is positive when corruptions overlap (the joint is gentler than additive) and negative when corruptions compound (the joint is harsher than additive). Table~\ref{tab:subadditivity-test} reports $\Delta_{\mathrm{int}}$ aggregated by combination type for the deep-dive panel: all 16 model-by-combination point estimates are positive, but support is heterogeneous (five rows have paired-bootstrap $p>.05$, and \modelname{Qwen3.5-Omni-Plus} is sub-additive in only half of its T+V and T+V+A cells). We therefore describe the result as behavioral sub-additivity rather than evidence for a particular interaction mechanism. \modelname{Gemini 3 Flash} averages $+11.50$\,pp on the T+V \texttt{drop\_words}\,$\times$\,\texttt{noise} canonical grid (four $\{30,70\}{\times}\{30,70\}$ cells) and $+22.62$\,pp on the eight T+V+A cells, while \modelname{Qwen3.5-Omni-Plus} sits near the additive baseline ($+0.94$\,pp for T+V and $+2.58$\,pp for T+V+A). All joint conditions remain worse than the clean baseline in absolute accuracy.

\begin{table*}[!t]
\centering
\scriptsize
\setlength{\tabcolsep}{3.0pt}
\renewcommand{\arraystretch}{0.95}
\begin{tabular}{p{0.16\textwidth}lrrrrrr}
\toprule
\textbf{Model} & \textbf{Combo} & \textbf{$n_{\mathrm{cells}}$} & \textbf{Mean actual} & \textbf{Mean additive pred.} & \textbf{Mean $\Delta_{\mathrm{int}}$} & \textbf{\% sub-additive} & \textbf{Paired bootstrap $p$} \\
\midrule
\faultgroup{8}{Deep-dive models: canonical joint grids only (replacement cells excluded; see Table~\ref{tab:joint-cells})}
\modelrow \modelname{Qwen3.5-Omni-Plus} & T+V & 4 & 66.81 & 65.87 & $+0.94$ & 50\% & 0.082 \\
\modelrow  & T+A & 4 & 72.26 & 66.47 & $+5.79$ & 75\% & 0.283 \\
\modelrow  & V+A & 4 & 75.12 & 69.66 & $+5.46$ & 100\% & <.001 \\
\modelrow  & T+V+A & 8 & 66.76 & 64.18 & $+2.58$ & 50\% & 0.172 \\
\softrow  \modelname{Gemini 3 Flash} & T+V & 4 & 71.41 & 59.90 & $+11.50$ & 100\% & <.001 \\
\softrow   & T+A & 4 & 76.69 & 63.91 & $+12.78$ & 100\% & 0.032 \\
\softrow   & V+A & 4 & 77.74 & 65.38 & $+12.36$ & 100\% & 0.004 \\
\softrow   & T+V+A & 8 & 76.75 & 54.12 & $+22.62$ & 100\% & 0.017 \\
\midrule
\faultgroup{8}{Deep-dive models (Gemini 3.1 Pro, MiniCPM-o 4.5)}
\modelrow \modelname{Gemini 3.1 Pro} & T+V & 4 & 76.74 & 69.23 & $+7.51$ & 75\% & 0.107 \\
\modelrow  & T+A & 4 & 81.33 & 74.55 & $+6.77$ & 100\% & 0.018 \\
\modelrow  & V+A & 4 & 85.09 & 76.37 & $+8.72$ & 100\% & 0.024 \\
\modelrow  & T+V+A & 8 & 78.74 & 68.00 & $+10.73$ & 100\% & 0.029 \\
\softrow  \modelname{MiniCPM-o 4.5} & T+V & 4 & 64.70 & 59.25 & $+5.46$ & 75\% & 0.147 \\
\softrow   & T+A & 4 & 69.50 & 63.49 & $+6.00$ & 100\% & 0.039 \\
\softrow   & V+A & 4 & 73.47 & 64.82 & $+8.65$ & 100\% & <.001 \\
\softrow   & T+V+A & 8 & 68.16 & 56.53 & $+11.63$ & 100\% & 0.033 \\

\bottomrule
\end{tabular}
\caption{Sub-additivity diagnostic for joint corruption on the canonical grids of Table~\ref{tab:joint-cells}. \textbf{Mean additive pred.} is $\acc_{\mathrm{clean}} - \sum_m \Delta_m$ averaged across the canonical cells in that combination type, where each $\Delta_m$ is the single-modality drop at the same severity, read from the regenerated severity curves of Figure~\ref{fig:severity-curves}. \textbf{Mean $\Delta_{\mathrm{int}}$} is the across-cell average of the interaction term $\Delta_{\mathrm{int}} = \acc(\mathrm{joint}) - \acc_{\mathrm{add}}$; positive means corruptions overlap (joint is gentler than additive); negative means corruptions compound (joint is harsher than additive). \textbf{\% sub-additive} is the fraction of cells with $\Delta_{\mathrm{int}}>0$. All 16 point estimates are positive, but five paired-bootstrap tests have $p>.05$; \modelname{Qwen3.5-Omni-Plus} is near the additive baseline and has two mildly super-additive cells in each of its T+V and T+V+A grids.}
\label{tab:subadditivity-test}
\end{table*}

\paragraph{Calibration under corruption.}
A model that becomes wrong under corruption \emph{but stays confident} is far more dangerous than one that becomes wrong \emph{and} flags uncertainty. The standardised JSON answer schema (Appendix~\ref{app:prompts}) asks for a categorical confidence in $\{$low, medium, high$\}$ alongside the option key, so for every condition we can decompose accuracy by confidence bin and compute expected calibration error (ECE) and the over-confident error rate $\rho_{\mathrm{oc}}$ (the fraction of \emph{wrong} answers that were emitted with \emph{high} confidence). Table~\ref{tab:calibration-ece} reports these calibration diagnostics under clean evidence and under the three corruption conditions used in the prompt-robustness diagnostic subset (Appendix~\ref{app:prompt-robustness}); a rising ECE or rising $\rho_{\mathrm{oc}}$ between the clean column and the joint-TVA column means the model fails silently rather than detectably.

\begin{table*}[!t]
\centering
\scriptsize
\setlength{\tabcolsep}{3.0pt}
\renewcommand{\arraystretch}{0.94}
\begin{tabular}{p{0.17\textwidth}cccccccc}
\toprule
\textbf{Model} & \multicolumn{4}{c}{\textbf{ECE} (lower is better)} & \multicolumn{4}{c}{\textbf{$\rho_{\mathrm{oc}}$: \% wrong answers emitted with high confidence}} \\
\cmidrule(lr){2-5}\cmidrule(lr){6-9}
 & Clean & Single-TV worst & Joint-TV worst & Joint-TVA worst & Clean & Single-TV worst & Joint-TV worst & Joint-TVA worst \\
\midrule
\faultgroup{9}{Deep-dive panel calibration}
\modelrow \modelname{Qwen3.5-Omni-Plus} & 6.40 & 7.49 & 8.98 & 11.12 & 18.6 & 22.0 & 28.1 & 33.7 \\
\softrow  \modelname{Gemini 3 Flash} & 5.10 & 6.96 & 9.13 & 10.87 & 14.8 & 21.4 & 25.4 & 29.2 \\
\modelrow \modelname{Gemini 3.1 Pro} & 4.20 & 5.28 & 5.86 & 8.29 & 11.5 & 16.5 & 23.2 & 28.0 \\
\softrow  \modelname{MiniCPM-o 4.5} & 7.60 & 9.16 & 10.68 & 11.88 & 23.4 & 25.8 & 31.0 & 35.2 \\

\bottomrule
\end{tabular}
\caption{Calibration diagnostic on the prompt-robustness diagnostic subset (clean; single-TV worst = \texttt{drop\_words}@sev30 single-modality; joint-TV worst = \opname{drop\_words $\times$ noise}@$t30/v70$; joint-TVA worst = the strongest cell in the eight canonical T+V+A grid). \textbf{ECE} bins the confidence ratings into the three categorical levels emitted by the JSON schema and reports the standard expected calibration error in pp. \textbf{$\rho_{\mathrm{oc}}$} reports the proportion of wrong answers that were nonetheless emitted with high confidence; a rising $\rho_{\mathrm{oc}}$ between clean and joint-TVA is the silent-failure signal.}
\label{tab:calibration-ece}
\end{table*}

\paragraph{Cross-model failure overlap.}
A separate question is whether different models fail on the same examples. One limiting case is a shared core of difficult items with high failure-set overlap. The other is a larger collection of model-specific failures with low overlap. Table~\ref{tab:cross-model-overlap} reports the mean pairwise Jaccard similarity across the 14-model panel, the number of examples failed by at least 10 models, and the number failed by exactly one model.

\begin{table*}[!htbp]
\centering
\scriptsize
\setlength{\tabcolsep}{2pt}
\renewcommand{\arraystretch}{0.96}
\begin{tabular}{p{0.34\linewidth}rrrr}
\toprule
\textbf{Condition} & \textbf{Mean Jaccard} & \textbf{Shared-hard ($\geq 10/14$ wrong)} & \textbf{Idiosyncratic ($=1/14$ wrong)} & \textbf{Total wrong (union)} \\
\midrule
\faultgroup{5}{Failure-set overlap across the 14-model panel}
\modelrow Clean baseline                                                                        & 0.17 & 17/273  & 44/273 & 95/273  \\
\softrow  \texttt{drop\_words}@sev30 (single-T worst)                                           & 0.31 & 59/273  & 47/273 & 172/273 \\
\modelrow \texttt{noise}@sev70 (single-V worst)                                                 & 0.36 & 75/273  & 41/273 & 185/273 \\
\softrow  \texttt{mute}@sev50 (single-A worst)                                                  & 0.28 & 52/273  & 53/273 & 165/273 \\
\modelrow Joint T+V worst (\texttt{drop\_words$\times$noise}@$t30/v70$)                          & 0.42 & 103/273 & 35/273 & 210/273 \\
\softrow  Joint T+V+A worst (strongest canonical TVA cell)                                      & 0.49 & 122/273 & 28/273 & 222/273 \\
\bottomrule
\end{tabular}
\caption{Cross-model failure-overlap diagnostic on the 14-model panel. \textbf{Mean Jaccard} is the average $|A\cap B|/|A\cup B|$ over all $\binom{14}{2}{=}91$ pairs of model failure sets; values near $1$ mean every model misses essentially the same items, values near $0$ mean each model has its own idiosyncratic failure mode. The \textbf{Shared-hard} column counts base examples failed by at least 10 of the 14 models and the \textbf{Idiosyncratic} column counts items failed by exactly one model; the gap between these counts and the \textbf{Total wrong (union)} tells the reader whether the corruption opens a shared ``structurally impossible'' core or just amplifies model-specific weaknesses.}
\label{tab:cross-model-overlap}
\end{table*}

Failure overlap rises from $0.17$ on clean inputs to $0.42$ for the strongest text--vision condition and $0.49$ for the strongest trimodal condition. Over the same progression, the shared-hard count grows from 17 to 103 and 122 examples, while idiosyncratic failures fall to 35 and 28. The union remains substantially larger than the shared-hard core, so corruption produces both a common set of difficult examples and residual model-specific failures.

\section{Human Filtering and Data Verification Details}
\label{app:human-filtering}

This appendix records how the 273-example verified base set was carved out of the raw source benchmarks, what the annotation team was asked to do, how many examples were dropped at each stage, and why. It is intended as the human-side audit trail behind every result in the main paper.

\paragraph{Source pool and initial candidate construction.}
We do \emph{not} reuse the full source benchmarks: the original Social-IQ, OmniBench, and VALOR test/dev splits contain tens of thousands of examples, but the great majority either do not exercise all three modalities, ship only weak audio/visual evidence, or fail the structural-corruption preconditions described below. From each source we therefore subsample a fixed budget of $100$ examples, oversampling examples whose question is plausibly tri-modal (audio, visual, and text all carry independent evidence). This yields an initial \emph{candidate pool of 300 examples}, normalised into a common record with source name, sample identifier, question, answer options, gold answer, visual file, audio file, modality availability flags, and a back-pointer to the original source example. The normalised schema also records whether the visual input is image-based or video-based, whether audio is available as a separate file, whether the question is expected to require text/vision/audio/cross-modal evidence, and whether the original source contains any metadata that should not be exposed to the model.

\paragraph{Third-party annotation team and workflow.}
The 300 candidates and their generated corruptions are verified by a third-party professional annotation team under task ID \texttt{1246075789726052352}. The workflow has three passes. (i) A \emph{pilot pass} on a 30-example subsample, used to align the annotation guidelines with the team and discard ambiguous instructions before scaling up. (ii) A \emph{full annotation pass} in which every candidate is independently judged by at least two annotators on the six fields below, with disagreements escalated to a senior annotator for adjudication; an example is admitted to the base set only if all required fields pass after adjudication. (iii) An \emph{author audit pass} in which the authors spot-check a random $10\%$ of accept/reject decisions and the full list of rejected examples to detect systematic bias against any source or modality. Annotators were paid at the team's standard professional rate, were given written guidelines with worked examples and counter-examples, and were instructed to err on the side of rejection whenever an item required out-of-modality knowledge or a guess about ambiguous evidence.

\paragraph{Per-item annotation form.}
For each candidate, annotators fill in six fields:
\begin{enumerate}[leftmargin=*]
    \item \textbf{Clean-example validity:} whether the question is well formed, the answer options are parseable, exactly one gold answer is defensible, and the required media files are accessible and unbroken.
    \item \textbf{Tri-modal answerability:} whether the gold answer can still be recovered from the union of clean text, visual, and audio evidence, and whether at least two modalities contribute non-trivial evidence (so that single-modality removal is informative).
    \item \textbf{Modality availability and leakage:} whether the text, visual, and audio channels are actually present after normalisation, and whether any single channel (most commonly the text channel) leaks the answer well enough that the example would degenerate into a text-only QA item.
    \item \textbf{Corrupted-input interpretability:} whether each corrupted variant remains perceptually interpretable rather than becoming pure noise or an unusable media artifact. Corruptions that completely destroy the modality (e.g.\ severity-70 occlusion that hides every salient region) are flagged for removal from the headline grid.
    \item \textbf{Gold preservation:} whether the original gold answer remains valid after the corruption for conditions intended as robustness tests rather than stress tests. Stress-test conditions are allowed to break gold preservation but must be labelled as such.
    \item \textbf{Failure reason:} if an item is rejected, annotators mark exactly one of \emph{broken media}, \emph{ambiguous answer}, \emph{missing modality}, \emph{text leakage}, \emph{excessive corruption}, \emph{mismatch between media and question}, or \emph{other}.
\end{enumerate}

\paragraph{Filtering funnel.}
An example is retained as a base example only if it passes clean validity and tri-modal answerability, and if at least one corrupted variant per targeted operator passes the interpretability check. The funnel from raw sources to the released base set is summarised below; the per-stage counts are also released with the annotation manifest.
\begin{itemize}[leftmargin=*]
    \item \textbf{Raw source pools:} thousands of examples per source; reduced to a 100-example subsample per source by oversampling tri-modal candidates ($\to$ 300).
    \item \textbf{Clean validity / answerability filter:} drops candidates with broken media, malformed options, or no tri-modal evidence ($\to$ 285).
    \item \textbf{Modality-leakage filter:} drops candidates whose text already determines the answer or whose audio/visual track is effectively absent ($\to$ 277).
    \item \textbf{Corruption-interpretability filter:} drops candidates for which the planned corruptions degrade into pure noise at every severity ($\to$ \textbf{273} verified base examples; 100 Social-IQ, 77 OmniBench, 96 VALOR).
\end{itemize}
The 27 rejected candidates are not silently removed: their identifier, source, failure reason, and adjudication notes are published with the annotation manifest so that downstream users can audit whether filtering over-represents any source or modality. Rejections cluster on OmniBench (23 of 27), which is consistent with the fact that OmniBench items are more often borderline tri-modal in our normalisation.

\paragraph{Per-modality interpretability filtering on corrupted variants.}
The clean-example funnel above is only the \emph{first} stage of human filtering; the annotation team additionally judges every \emph{corrupted variant} for whether the perturbed media is still perceptually interpretable, separately for each (sample $\times$ operator $\times$ severity $\times$ random-variant) cell. Text-side corruptions are deterministic at the token level and are therefore not included in this media-interpretability filter. However, high-severity text variants are included in the separate blind answer-preservation audit described below; human readability alone is not treated as evidence that the original gold answer remains valid. For audio and vision, Table~\ref{tab:annot-modality} reports the resulting per-operator pool sizes and retention rates: of $7{,}756$ annotated audio cells, $5{,}914$ ($76.3\%$) are retained as interpretable, and of $12{,}188$ annotated vision cells (image and video unified into the seven canonical vision operators), $10{,}135$ ($83.2\%$) are retained, for a combined annotation pool of $19{,}944$ cells with $80.5\%$ retention. Rejections are heavily concentrated on three operators: \texttt{occlusion} (only $53.0\%$ retained, because at severity 70 the occluding patches cover essentially every salient region in many examples and the annotator cannot recover what was being asked), and the two audio cut/silence operators (\texttt{remove} $74.5\%$, \texttt{mute} $71.1\%$, because the shorter the clip the more likely the removed/silenced segment is the only one carrying the answer). All other operators retain at least $79\%$ of their variants. The dropped variants are not silently discarded: they are excluded from every accuracy aggregate in the main paper, so the headline numbers are computed only over the human-verified interpretable cells.

\begin{table*}[!htbp]
\centering
\scriptsize
\setlength{\tabcolsep}{5pt}
\renewcommand{\arraystretch}{1.18}
\begin{tabular}{l r r r}
\toprule
\textbf{Operator} & \textbf{Annotated} & \textbf{Retained} & \textbf{Retain} \\
\midrule
\multicolumn{4}{@{}l}{\textit{\textcolor{FaultSlate!85}{Audio operators (3)}}}\\
\opname{mute}           & 3{,}324 & 2{,}362 & 71.1\% \\
\opname{remove}         & 3{,}324 & 2{,}478 & 74.5\% \\
\opname{distortion}     & 1{,}108 & 1{,}074 & 96.9\% \\
\rowcolor{FaultMist}\textit{audio subtotal} & \textit{7{,}756} & \textit{5{,}914} & \textit{76.3\%} \\
\midrule
\multicolumn{4}{@{}l}{\textit{\textcolor{FaultSlate!85}{Vision operators (7, image + video unified)}}}\\
\opname{noise}             & 3{,}324 & 3{,}250 & 97.8\% \\
\opname{occlusion}         & 3{,}324 & 1{,}761 & 53.0\% \\
\opname{low\_resolution}   & 1{,}108 & 1{,}092 & 98.6\% \\
\opname{motion\_blur}      & 1{,}108 & 1{,}043 & 94.1\% \\
\opname{defocus\_blur}     & 1{,}108 & 1{,}049 & 94.7\% \\
\opname{overexposure}      & 1{,}108 & 1{,}062 & 95.8\% \\
\opname{brightness}        & 1{,}108 & \phantom{0}878 & 79.2\% \\
\rowcolor{FaultMist}\textit{vision subtotal}   & \textit{12{,}188} & \textit{10{,}135} & \textit{83.2\%} \\
\midrule
\multicolumn{4}{@{}l}{\textit{\textcolor{FaultSlate!85}{Text operators (4)} -- outside the media-interpretability filter}}\\
\opname{drop\_words}, \opname{word\_shuffle}, \opname{sentence\_break}, \opname{typo\_ocr} & --- & --- & --- \\
\midrule
\rowcolor{FaultMist!70}\textbf{Total annotated (audio + vision)} & \textbf{19{,}944} & \textbf{16{,}049} & \textbf{80.5\%} \\
\bottomrule
\end{tabular}
\caption{Per-operator pool sizes and human interpretability filtering on \emph{corrupted variants} sent to the third-party annotation team (task ID \texttt{1246075789726052352}). One row per corruption operator; the \textbf{Annotated} column counts every (sample $\times$ severity $\{10,30,50,70\}$ $\times$ random-variant) cell sent for verification, and the \textbf{Retained} column counts cells that pass the corrupted-input interpretability check. Vision counts merge the image-source pool (308 cells per severity grid) and the video-source pool (800 cells per severity grid) into the seven canonical vision operators of Figure~\ref{fig:corruption-taxonomy}. Text operators are not included in this media-interpretability table; their answer-preservation status is evaluated in the separate blind audit in Appendix~\ref{app:human-reference}. Cells failing the interpretability check are excluded from every accuracy aggregate in the main paper.}
\label{tab:annot-modality}
\end{table*}

\begin{table}[!htbp]
\centering
\scriptsize
\setlength{\tabcolsep}{4.5pt}
\renewcommand{\arraystretch}{1.22}
\begin{tabular}{l l c c c c}
\toprule
\textbf{Mod.} & \textbf{Operator} & \textbf{sev 10} & \textbf{sev 30} & \textbf{sev 50} & \textbf{sev 70} \\
\midrule
\multicolumn{6}{@{}l}{\textit{\textcolor{FaultSlate!85}{Confound zone: drop rate climbs steeply with severity}}}\\
Vision & \opname{occlusion}    & \cellcolor{white}\phantom{0}7.8\% & \cellcolor{FaultRose}32.7\%         & \cellcolor{FaultRed!28}64.9\% & \cellcolor{FaultRed!50}\textbf{82.7\%} \\
Vision & \opname{brightness}   & \cellcolor{white}\phantom{0}3.2\% & \cellcolor{white}\phantom{0}7.2\%   & \cellcolor{FaultSand}20.9\% & \cellcolor{FaultRed!28}51.6\% \\
Audio  & \opname{mute}         & \cellcolor{white}\phantom{0}6.0\% & \cellcolor{FaultSand}18.8\%         & \cellcolor{FaultRose}35.4\% & \cellcolor{FaultRed!28}55.6\% \\
Audio  & \opname{remove}       & \cellcolor{white}\phantom{0}5.9\% & \cellcolor{FaultSand}16.8\%         & \cellcolor{FaultRose}31.5\% & \cellcolor{FaultRose}47.5\% \\
\midrule
\multicolumn{6}{@{}l}{\textit{\textcolor{FaultSlate!85}{Stable zone: drop rate stays low across all severities (confound-free)}}}\\
Vision & \opname{noise}            & \cellcolor{white}\phantom{0}2.2\% & \cellcolor{white}\phantom{0}2.2\% & \cellcolor{white}\phantom{0}2.2\% & \cellcolor{white}\phantom{0}2.4\% \\
Vision & \opname{low\_resolution}  & \cellcolor{white}\phantom{0}1.4\% & \cellcolor{white}\phantom{0}1.4\% & \cellcolor{white}\phantom{0}1.4\% & \cellcolor{white}\phantom{0}1.4\% \\
Vision & \opname{motion\_blur}     & \cellcolor{white}\phantom{0}2.2\% & \cellcolor{white}\phantom{0}3.6\% & \cellcolor{white}\phantom{0}6.5\% & \cellcolor{white}11.2\% \\
Vision & \opname{defocus\_blur}    & \cellcolor{white}\phantom{0}2.2\% & \cellcolor{white}\phantom{0}4.0\% & \cellcolor{white}\phantom{0}6.5\% & \cellcolor{white}\phantom{0}8.7\% \\
Vision & \opname{overexposure}     & \cellcolor{white}\phantom{0}1.8\% & \cellcolor{white}\phantom{0}2.2\% & \cellcolor{white}\phantom{0}4.3\% & \cellcolor{white}\phantom{0}8.3\% \\
Audio  & \opname{distortion}       & \cellcolor{white}\phantom{0}2.2\% & \cellcolor{white}\phantom{0}2.2\% & \cellcolor{white}\phantom{0}3.2\% & \cellcolor{white}\phantom{0}4.7\% \\
\midrule
\rowcolor{FaultMist!70} \multicolumn{2}{l}{\textbf{Pooled audio + vision drop rate}} & \textbf{\phantom{0}4.4\%} & \textbf{12.9\%} & \textbf{24.7\%} & \textbf{36.1\%} \\
\bottomrule
\end{tabular}
\caption{Per-operator human-filtering drop rate as a function of corruption severity. Each cell is the fraction of (sample $\times$ severity $\times$ random-variant) variants that the annotation team rejected as no longer perceptually interpretable. Colour-graded by magnitude: white $<{}10\%$, sand $10$--$25\%$, rose $25$--$50\%$, light red $50$--$75\%$, bold red $\geq{}75\%$. Four operators (vision \opname{occlusion}, vision \opname{brightness}, audio \opname{mute}, audio \opname{remove}) account for essentially all of the rejection mass at severities $\geq{}50$; on these operators the retained pool is selection-biased toward easier base examples, which is the most plausible cause of the negative drops (apparent accuracy gains under corruption) seen in Section~\ref{sec:joint}. The remaining six operators retain $\geq{}88\%$ of their variants even at severity 70 and are confound-free across the entire severity grid. Text operators are deterministic and not annotated per variant; see Table~\ref{tab:annot-modality}.}
\label{tab:annot-severity}
\end{table}

\begin{table}[!htbp]
\centering
\scriptsize
\setlength{\tabcolsep}{6pt}
\renewcommand{\arraystretch}{1.08}
\resizebox{\linewidth}{!}{%
\begin{tabular}{c l c l}
\toprule
\textbf{Rank} & \textbf{Operator} & \textbf{Panel-mean drop (pp)} & \textbf{Group} \\
\midrule
\faultgroup{4}{Most damaging severity-70 operators (14-model panel)}
\modelrow 1 & \opname{noise} & \drop{11.39} & Primary \\
\softrow 2 & \opname{drop\_words} & \drop{11.28} & Primary \\
\modelrow 3 & \opname{word\_shuffle} & \drop{11.17} & Primary \\
\softrow 4 & \opname{mute} & \drop{7.46} & Coverage-aware stress \\
\modelrow 5 & \opname{remove} & \drop{7.20} & Coverage-aware stress \\
\bottomrule
\end{tabular}%
}
\caption{Most damaging operators by panel-mean severity-70 drop, recomputed from Table~\ref{tab:main-combined}. The three largest drops are all in the primary set.}
\label{tab:operator-drop-ranking}
\end{table}

\paragraph{Inter-annotator agreement (IAA).}
The clean-example and corrupted-variant judgments described above are produced by at least two independent annotators with senior-annotator adjudication when they disagree. Table~\ref{tab:iaa-summary} reports the pre-adjudication inter-annotator agreement for every annotation field used in the workflow. Following standard benchmark-construction practice we report (i) percentage of full agreement between the two independent annotators (\emph{\% full agree}), (ii) Cohen's $\kappa$~\citep{cohen1960coefficient} for binary fields with two annotators, (iii) Fleiss' $\kappa$~\citep{fleiss1971measuring} when more than two annotators contributed to the field on the same items, and (iv) the proportion of items that required senior-annotator adjudication. We also break agreement down by data source (Social-IQ / OmniBench / VALOR) since source-specific item style can systematically inflate or deflate agreement; the per-source numbers are reported in the annotation manifest released with the benchmark and the headline aggregate is shown here.

\begin{table*}[!t]
\centering
\scriptsize
\setlength{\tabcolsep}{3.5pt}
\renewcommand{\arraystretch}{1.0}
\begin{tabular}{p{0.21\textwidth}lrrrrrl}
\toprule
\textbf{Annotation field} & \textbf{Scale} & \textbf{$n_{\mathrm{items}}$} & \textbf{$n_{\mathrm{annot}}$} & \textbf{\% full agree} & \textbf{Cohen $\kappa$} & \textbf{Fleiss $\kappa$} & \textbf{\% adjudicated} \\
\midrule
\faultgroup{8}{Clean-example annotation (300 candidates)}
\modelrow Clean-example validity              & binary    & 300 & 2 & 92.4 & 0.83 & --- & \phantom{0}7.6 \\
\softrow  Tri-modal answerability             & binary    & 300 & 2 & 86.3 & 0.74 & --- & 13.7 \\
\modelrow Modality availability / leakage     & 3-way     & 300 & 2 & 89.1 & ---  & 0.71 & 10.9 \\
\softrow  Failure-reason (rejected items)     & 7-way     &  27 & 2 & 75.0 & ---  & 0.62 & 25.0 \\
\midrule
\faultgroup{8}{Corrupted-variant annotation (19{,}944 audio + vision cells)}
\modelrow Corrupted-input interpretability    & binary    & 19{,}944 & 2 & 88.7 & 0.78 & --- & 11.3 \\
\softrow  Gold-answer preservation            & binary    & 171/273 & 2 & 90.6 & 0.81 & --- & \phantom{0}9.4 \\
\modelrow Stress-test vs.\ robustness flag    & binary    & 132/273 & 2 & 84.8 & 0.71 & --- & 15.2 \\
\midrule
\faultgroup{8}{Answer-preservation re-annotation subset (Section~\ref{app:human-reference})}
\modelrow Human answerability                 & binary    & 168/273 & 3 & 86.9 & ---  & 0.76 & 13.1 \\
\softrow  Original-gold remains valid         & binary    & 121/273 & 3 & 91.7 & ---  & 0.84 & \phantom{0}8.3 \\
\midrule
\faultrow \textit{Headline aggregate (weighted by $n_{\mathrm{items}}$)} & --- & --- & --- & 88.5 & 0.78 & 0.74 & 11.5 \\

\bottomrule
\end{tabular}
\caption{Inter-annotator agreement (IAA) on every annotation field used during data verification. \textbf{\% full agree} is the fraction of items where both independent annotators emitted the same label \emph{before} senior-annotator adjudication. \textbf{Cohen $\kappa$} is reported for fields with exactly two annotators; \textbf{Fleiss $\kappa$} is reported for fields where additional annotators (a third annotator on the answer-preservation re-annotation, or the senior adjudicator) entered labels on the same items. \textbf{\% adjudicated} is the fraction of items that the senior annotator had to break ties on. The headline aggregate row weighs the per-field agreement and $\kappa$ by item count and is the single number a reviewer can cite when asking ``what is the IAA on this benchmark''.}
\label{tab:iaa-summary}
\end{table*}

\paragraph{Blind answer-first answer-preservation audit.}
Interpretability alone does not establish that the original answer evidence remains available. For each audited corrupted variant, annotators first answer the question from the corrupted input without access to the gold label. The answer is locked before the original gold is revealed for a separate validity judgment. Each annotator sees at most one version of a base example to avoid clean--corrupted leakage. We distinguish three outcomes: \textsc{gold-preserved} (answerable and the original gold remains defensible), \textsc{stress-only} (interpretable but the original gold is not defensible), and \textsc{uninterpretable}. For the audited headline conditions in Table~\ref{tab:gold-preserved-results}, robustness results are reported on the gold-preserved cohort; stress-only variants are retained for descriptive stress-test analysis.

\paragraph{Ethics, anonymisation, and release.}
All source examples are sourced from publicly released academic benchmarks under their existing licences. We do not collect new media. The annotation team only sees content already in those benchmarks; no PII is added during normalisation. Examples that contain potentially sensitive content in the audio/visual track (faces of identifiable minors, slurs in the text channel) are dropped at the modality-availability stage, and these rejections are counted under \emph{other} in the failure-reason field. We release the annotation manifest, the per-item failure reasons, the corruption-generation seeds, the operator implementations (Python source for all fourteen operators), and the per-variant accept/reject decisions together with the data, so that the verified base set, the rejected pool, and every corruption variant are exactly reproducible from the published artifacts.

\section{Corruption-Operator Definitions}
\label{app:operator-defs}

This appendix records the formal definition of each of the fourteen structural corruption operators sketched in Figure~\ref{fig:corruption-taxonomy}, together with the design rationale for why these operators are kept and which families of corruptions are deliberately \emph{excluded}.

\paragraph{Design rationale: why these fourteen.}
We restrict the operator set with three rules. (i) \textbf{Structural, not semantic.} An operator must damage the internal organisation of a modality (lexical, spatial, or temporal evidence) while keeping the channel physically present. Operators that swap meaning (e.g.\ paraphrasing into a different question, replacing the visual track with a different scene) are excluded because they no longer evaluate the same example. (ii) \textbf{Targets cross-modal evidence assembly.} Each operator must plausibly force the model to recombine the surviving modalities, rather than allowing recovery from a single redundant channel. Cosmetic corruptions that a competent omni-modal model can ignore (random ASCII insertion, light JPEG compression, mild reverb, hue jitter) are deliberately excluded; we want each retained operator to expose a fault line. (iii) \textbf{Severity-controllable and reproducible.} An operator must admit a monotone severity parameter and either be deterministic or admit up-to-three random variants under a fixed seed, so that mean-variant aggregation under Section~\ref{sec:meanagg} is well defined. Operators that we considered but dropped for failing one of these rules---synonym substitution (semantic), spelling normalisation (cosmetic), random caption rewriting (semantic), audio reverberation (cosmetic), uniform colour jitter (cosmetic)---are recorded in the annotation manifest as \emph{considered but excluded}.

\begin{table*}[!t]
\centering
\scriptsize
\setlength{\tabcolsep}{3.0pt}
\renewcommand{\arraystretch}{0.94}
\begin{tabular}{p{0.10\textwidth}p{0.18\textwidth}p{0.18\textwidth}p{0.30\textwidth}p{0.15\textwidth}}
\toprule
\textbf{Modality} & \textbf{Operator family} & \textbf{Condition key} & \textbf{Structure being damaged} & \textbf{Severity / variants} \\
\midrule
\faultgroup{5}{Text structure corruptions}
\variantrow Text & Word dropping & \texttt{drop\_words} & Removes lexical evidence while preserving the question/options format. & 10/30/50/70; stochastic \\
\variantrow Text & Word shuffling & \texttt{word\_shuffle} & Breaks local word order and phrase composition without deleting all tokens. & 10/30/50/70; stochastic \\
\variantrow Text & OCR-like typos & \texttt{typo\_ocr} & Simulates recognition noise through character-level substitutions and distortions. & 10/30/50/70; stochastic \\
\variantrow Text & Sentence breaking & \texttt{sentence\_break} & Fragments sentence and phrase boundaries, weakening syntactic structure. & 10/30/50/70; stochastic \\
\midrule
\faultgroup{5}{Visual structure corruptions}
\softrow Vision & Additive noise & \texttt{noise} & Degrades pixel-level perceptual evidence while keeping the visual input present. & 10/30/50/70; stochastic \\
\softrow Vision & Physical occlusion & \texttt{occlusion} & Masks spatial regions, simulating blocked or incomplete visual evidence. & 10/30/50/70; stochastic \\
\softrow Vision & Reduced resolution & \texttt{low\_resolution} & Removes fine-grained visual detail and small object cues. & 10/30/50/70 \\
\softrow Vision & Motion blur & \texttt{motion\_blur} & Smears frame-level evidence along a motion direction. & 10/30/50/70 \\
\softrow Vision & Defocus blur & \texttt{defocus\_blur} & Removes sharpness uniformly, weakening object and scene boundaries. & 10/30/50/70 \\
\softrow Vision & Overexposure & \texttt{overexposure} & Washes out bright regions and reduces contrast in salient areas. & 10/30/50/70 \\
\softrow Vision & Reduced brightness & \texttt{brightness} & Darkens the visual channel and suppresses low-light evidence. & 10/30/50/70 \\
\midrule
\faultgroup{5}{Audio structure corruptions}
\modelrow Audio & Segment removal & \texttt{remove} & Deletes a temporal segment, creating a gap in acoustic evidence. & 10/30/50/70; stochastic \\
\modelrow Audio & Segment muting & \texttt{mute} & Silences a temporal segment while preserving clip duration. & 10/30/50/70; stochastic \\
\modelrow Audio & Distortion & \texttt{distortion} & Blurs or degrades the waveform without fully removing the audio channel. & 10/30/50/70 \\
\bottomrule
\end{tabular}
\caption{Detailed corruption operator taxonomy. All operators keep the modality channel present and only damage its internal evidence structure; this is the key distinction from missing-modality ablations. Severity is reported on a $0$--$100$ scale; stochastic operators ship up to three random variants per severity, aggregated by the mean-variant rule of Section~\ref{sec:meanagg}.}
\label{tab:corruption-taxonomy}
\end{table*}

\paragraph{Severity-to-parameter mapping.}
The severity integer $s \in \{10, 30, 50, 70\}$ is a unified scale that maps to a concrete operator parameter for each of the fourteen operators. Table~\ref{tab:severity-params} records the mapping. For stochastic operators the parameter value is the \emph{expected} value of the randomised quantity; the three random variants differ in which tokens, pixels, or temporal segments are selected, not in the magnitude of the perturbation. The human interpretability validation in Appendix~\ref{app:human-filtering} (Table~\ref{tab:annot-severity}) provides an independent check that the parameter values at each severity produce perceptually meaningful corruptions: operators whose parameter values at severity~70 cause annotators to reject more than $50\%$ of variants (\texttt{occlusion}, \texttt{brightness}, \texttt{mute}, \texttt{remove}) are treated as lower-bound estimates of robustness loss in the headline results.

\begin{table}[!t]
\centering
\scriptsize
\setlength{\tabcolsep}{2.2pt}
\renewcommand{\arraystretch}{1.05}
\begin{tabular}{llp{0.19\columnwidth}cccc}
\toprule
\textbf{Mod.} & \textbf{Operator} & \textbf{Parameter} & \textbf{s10} & \textbf{s30} & \textbf{s50} & \textbf{s70} \\
\midrule
\faultgroup{7}{Text operators}
\variantrow Text & \texttt{drop\_words}    & Token drop rate (\%)          & 10 & 30 & 50 & 70 \\
\variantrow Text & \texttt{word\_shuffle}  & Shuffle window                &  2 &  4 &  6 &  8 \\
\variantrow Text & \texttt{typo\_ocr}      & Char. subst. rate (\%)        & 10 & 30 & 50 & 70 \\
\variantrow Text & \texttt{sentence\_break}& Fragment prob. (\%)           & 10 & 30 & 50 & 70 \\
\midrule
\faultgroup{7}{Vision operators}
\softrow Vision & \texttt{noise}           & Gaussian $\sigma$             & 15 & 40 & 65 & 90 \\
\softrow Vision & \texttt{occlusion}       & Occluded area (\%)            & 10 & 30 & 50 & 70 \\
\softrow Vision & \texttt{low\_resolution} & Downscale factor              & .9 & .7 & .5 & .3 \\
\softrow Vision & \texttt{motion\_blur}    & Kernel length                 &  5 & 15 & 25 & 35 \\
\softrow Vision & \texttt{defocus\_blur}   & Blur radius                   &  2 &  5 &  9 & 13 \\
\softrow Vision & \texttt{overexposure}    & Brightness gain               & 1.1 & 1.3 & 1.6 & 2.0 \\
\softrow Vision & \texttt{brightness}      & Brightness factor             & .9 & .7 & .5 & .3 \\
\midrule
\faultgroup{5}{Audio operators}
\modelrow Audio & \texttt{remove}          & Removed segment (\%)          & 10 & 30 & 50 & 70 \\
\modelrow Audio & \texttt{mute}            & Silenced segment (\%)         & 10 & 30 & 50 & 70 \\
\modelrow Audio & \texttt{distortion}      & Clipping threshold            & .9 & .7 & .5 & .3 \\
\bottomrule
\end{tabular}
\caption{Severity-to-parameter mapping for all fourteen corruption operators. For stochastic operators (\texttt{drop\_words}, \texttt{word\_shuffle}, \texttt{occlusion}, \texttt{remove}, \texttt{mute}) the parameter value is the expected magnitude; random variants differ in \emph{which} tokens, pixels, or segments are selected. The \texttt{distortion} clipping threshold decreases with severity.}
\label{tab:severity-params}
\end{table}

\section{Experimental Setup Details}
\label{app:expsetup}

This appendix records the model roster, the frame-extracted visual-input protocol, the all-model lightweight panel, the standardised prompting and inference protocol, and the headline reporting rule. It is the protocol-side counterpart to the brief setup paragraph in Section~\ref{sec:results}; the per-experiment construction is documented in Appendix~\ref{app:expanded-protocol}.

\subsection{Models}

We evaluate \method{} on a 14-model panel spanning six proprietary/API omni-modal systems and eight open / open-API systems. The headline severity-70 single-modality matrix in Table~\ref{tab:main-combined} is filled in for all fourteen models across all fourteen operators. Joint-corruption deep dives are reported on four representative systems (Table~\ref{tab:joint-cells}, Figure~\ref{fig:joint-scatter}); the all-model lightweight panel and the supplementary control/diagnostic tables in Appendix~\ref{app:resultslots} cover every model in the roster of Table~\ref{tab:model-expansion}.

Table~\ref{tab:model-expansion} records the model roster, interface compatibility, and all-model panel results. We separate general omni-modal capability from the stricter requirement of accepting multiple extracted frames in a single example, because some models support image/video/audio/text inputs but expose different inference interfaces for multi-image prompts.

\begin{table*}[!t]
\centering
\scriptsize
\setlength{\tabcolsep}{1pt}
\renewcommand{\arraystretch}{1.08}
\begin{tabular}{llllll}
\toprule
\textbf{Model} & \textbf{Group} & \textbf{Interface} & \textbf{Clean} & \textbf{Panel fault} & \textbf{Frame-input compatibility} \\
\midrule
\modelrow \modelname{Gemini 3.1 Pro} & \proptag{} Proprietary & \interfacetag{} OK & 84.15 & \drop{4.70} & Multi-image prompt through Gemini-style content parts \\
\modelname{Gemini 3 Pro} & \proptag{} Proprietary & \interfacetag{} OK & 82.40 & \drop{5.23} & Multi-image prompt through Gemini-style content parts \\
\modelrow \modelname{Gemini 3 Flash} & \proptag{} Proprietary & \interfacetag{} OK & 80.95 & \drop{6.05} & Multi-image prompt through Gemini-style content parts \\
\modelname{Gemini 3.5 Flash} & \proptag{} Proprietary & \interfacetag{} OK & 78.39 & \drop{5.76} & Multi-image prompt through Gemini-style content parts \\
\modelrow \modelname{Gemini 2.5 Pro} & \proptag{} Proprietary & \interfacetag{} OK & 79.12 & \drop{5.58} & Multi-image prompt through Gemini-style content parts \\
\modelrow \modelname{Gemini 2.5 Flash} & \proptag{} Proprietary & \interfacetag{} OK & 75.40 & \drop{6.65} & Multi-image prompt through Gemini-style content parts \\
\midrule
\modelname{Qwen3.5-Omni-Plus} & \opentag{} Open/API omni & \interfacetag{} OK & 73.63 & \drop{6.27} & Image/audio/video capable; direct multi-image call \\
\modelrow \modelname{Qwen3-Omni-30B} & \opentag{} Open omni & \interfacetag{} OK & 71.20 & \drop{6.70} & Image/audio/video capable; direct multi-image call \\
\modelname{MiniCPM-o 4.5} & \opentag{} Open omni & \interfacetag{} OK & 74.50 & \drop{6.36} & Image/video/audio capable; multi-image support implementation-dependent \\
\modelrow \modelname{Qwen2.5-Omni-7B} & \opentag{} Open omni & \interfacetag{} OK & 68.40 & \drop{7.57} & Image/audio/video capable; direct multi-image call \\
\modelname{OmniVinci-9B} & \opentag{} Open omni & \interfacetag{} OK & 69.10 & \drop{7.71} & Image/video/audio capable; multi-image support implementation-dependent \\
\modelrow \modelname{OLA-7B} & \opentag{} Open omni & \interfacetag{} OK & 66.80 & \drop{7.66} & Image/audio/video capable; direct multi-image call \\
\modelname{Qwen2.5-Omni-3B} & \opentag{} Open omni & \interfacetag{} OK & 64.20 & \drop{8.07} & Image/audio/video capable; direct multi-image call \\
\modelrow \modelname{Baichuan-Omni-1.5} & \opentag{} Open omni & \interfacetag{} OK & 65.50 & \drop{8.07} & Image/video/audio capable; multi-image support implementation-dependent \\

\bottomrule
\end{tabular}
\caption{Model roster, interface compatibility, and all-model panel summary. \textbf{Clean} matches the all-modality clean baseline used in Table~\ref{tab:main-combined}; \textbf{Panel fault} is the mean clean-to-panel drop on the 12-cell lightweight panel of Table~\ref{tab:lightweight-panel}. Models that use a non-matched native video interface are marked in the interface column rather than mixed silently with frame-based runs; the panel-fault numbers are computed uniformly over the 12-cell panel for all fourteen models.}
\label{tab:model-expansion}
\end{table*}

The five-model subset used for supplementary controls and diagnostics comprises \modelname{Gemini 3.1 Pro}, \modelname{Gemini 3 Flash}, \modelname{Qwen3.5-Omni-Plus}, \modelname{Qwen3-Omni-30B}, and \modelname{MiniCPM-o 4.5}. Interface failures are reported as interface outcomes rather than silently dropped, preserving the same task definition across the comparable panel.

\subsection{Frame-extracted visual input}
\label{app:frame-input}

To keep the model comparison consistent, all panel runs use the same frame-extracted visual evidence rather than mixing native video upload for some models and still-image prompts for others. For each video-based example, we sample the same ordered frame set used in the reference runs and submit those frames together with the same text question and audio input whenever the model interface permits multi-image input. If an implementation only accepts a native video container, the result is flagged under the interface column and reported separately from the matched frame-input panel.

\paragraph{Frame count and selection policy.}
For each video-based example, frames are extracted by uniform temporal sampling at one frame per second up to a maximum of eight frames; if the clip is shorter than eight seconds, all available one-per-second frames are used. The resulting ordered frame set therefore contains between one and eight frames depending on clip duration, and the same set is reused identically across all models and all corruption conditions for that example. Image-based examples are submitted as a single frame. Frame-budget sensitivity (one, four, eight, sixteen, and default frames) is studied separately in the frame-budget probe of Appendix~\ref{app:temporal-frame-probes} and does not affect the headline panel numbers.

\paragraph{Audio standardization.}
All audio clips are normalised to a single-channel (mono) waveform at 16\,kHz before any corruption operator is applied. Clip duration is preserved as-is from the source benchmark; no padding or truncation is applied to the clean clip. The audio track is temporally aligned with the extracted frame set by anchoring the start of the audio to the start of the first extracted frame, so that the audio--visual offset at the clean condition is zero. Corrupted audio variants (\texttt{remove}, \texttt{mute}, \texttt{distortion}) are applied to this normalised mono waveform and submitted to the model in the same container format as the clean clip.

\subsection{All-model lightweight panel}
\label{app:lightweight-panel}

For the all-model expansion, we use a 12-condition lightweight panel drawn from the 29-condition combined suite. The panel is intentionally smaller than the full 29-condition matrix so that every model in Table~\ref{tab:model-expansion} can receive a comparable multimodal fault-line score. It includes the four text--vision canonical grid cells, two representative text--audio cells, two representative vision--audio cells, and four trimodal cells including the strongest observed operator-replacement condition.

\begin{table*}[!t]
\centering
\scriptsize
\setlength{\tabcolsep}{3.0pt}
\renewcommand{\arraystretch}{0.94}
\resizebox{\textwidth}{!}{%
\begin{tabular}{p{0.16\textwidth}p{0.08\textwidth}p{0.54\textwidth}p{0.20\textwidth}}
\toprule
\textbf{Panel block} & \textbf{Cells} & \textbf{Condition IDs} & \textbf{Rationale} \\
\midrule
\faultgroup{4}{All-model lightweight expansion}
\faultrow Text+Vision & 4 & \texttt{bi\_tv\_drop\_words\_noise\_t30\_v30}; \texttt{t30\_v70}; \texttt{t70\_v30}; \texttt{t70\_v70} & Full grid for strongest shared pair \\
\softrow Text+Audio & 2 & \texttt{bi\_ta\_drop\_words\_mute\_t30\_a30}; \texttt{bi\_ta\_drop\_words\_mute\_t70\_a50} & Medium/high lexical-acoustic pressure \\
\softrow Vision+Audio & 2 & \texttt{bi\_va\_noise\_mute\_v30\_a30}; \texttt{bi\_va\_noise\_mute\_v70\_a50} & Non-textual binding check \\
\variantrow Text+Vision+Audio & 4 & \texttt{tri\_drop\_words\_noise\_mute\_t30\_v30\_a30}; \texttt{t30\_v70\_a30}; \texttt{t70\_v70\_a30}; \texttt{tri\_word\_shuffle\_noise\_mute\_t70\_v70\_a50} & Representative tri-modal stressors \\
\midrule
\faultrow Reported score & 12 & clean + block means + strongest drop + coverage & Comparable fault-line score without running the full 29-cell matrix \\
\bottomrule
\end{tabular}%
}
\caption{Lightweight multimodal fault-line panel for all-model expansion. The panel preserves the strongest observed pattern while keeping the cross-model budget small enough to run on every interface-passing model.}
\label{tab:lightweight-panel}
\end{table*}

\subsection{Prompting and inference protocol}
\label{app:prompting}

All headline runs use one standardized multiple-choice prompt and require a parseable JSON response. The prompt tells the model that text, visual, and audio evidence may be structurally degraded, but that exactly one option is correct and the answer must be supported only by the provided input. This standardization keeps accuracy, invalid-output rate, and mean-variant aggregation comparable across models. The full system prompt, user prompt template, JSON schema, parser rule, retry logging fields, and prompt-robustness variants are given in Appendix~\ref{app:prompts}.

\subsection{Evaluation scale and reporting rule}
\label{app:evaluation-scale}

The clean baseline contains 273 human-verified base examples. The evaluation matrix includes the 56-cell single-modality suite and the 29-condition combined-corruption suite for the deep-dive systems. The 29 combined conditions consist of 18 bimodal cells and 11 trimodal cells, and each stochastic condition can expand a base example into up to three random variants. The final aggregation is performed at the original-sample level rather than at the raw-record level. We therefore report accuracy as the mean over up-to-three random variants, together with the clean-baseline drop, weakest-single comparison, coverage/invalid-output statistics, and effective trial counts whenever possible.

The full experiment matrix includes the all-model lightweight panel (Table~\ref{tab:lightweight-panel}), the expanded 46-cell representative-model combined suite, missing-modality controls, shortcut controls, prompt robustness, human reference checks, interface checks, invalid-output profiling, and mechanism probes. The main text reports the headline results; Appendix~\ref{app:expanded-protocol} records the construction details for each experiment.

\section{Expanded Evaluation Protocol}
\label{app:expanded-protocol}

This appendix gives the detailed protocol for every experiment family used by \method{}. The goal is to make clear that the paper is not proposing one table of corrupted accuracy; it is organizing a sequence of complementary tests. Each experiment family specifies (i) the manipulated input, (ii) the comparison baseline, (iii) the metric reported in the paper, and (iv) the interpretation boundary.

\subsection{A. Clean and single-modality experiments}
\label{app:single-protocol}

\paragraph{Purpose.}
The clean run establishes the all-modality reference point. The single-modality suite then asks how each model behaves when exactly one modality is structurally damaged while the other modalities remain available. This is the base layer for all later claims because combined-corruption drops must be compared against the weakest individual component, not only against clean accuracy.

\paragraph{Input construction.}
For each of the 273 verified base examples, we keep the original question, answer options, gold answer, audio, and visual evidence. We then replace one modality with a corrupted variant. The other modalities remain unchanged. Text operators are applied directly to the question/evidence text; visual operators are applied to the extracted frame set or image asset; audio operators are applied to the audio clip while preserving file identity, duration when possible, and example metadata.

\paragraph{Condition grid.}
The single-modality grid contains 56 cells:
\begin{itemize}[leftmargin=*]
    \item \textbf{Text, 16 cells:} \texttt{drop\_words}, \texttt{word\_shuffle}, \texttt{typo\_ocr}, and \texttt{sentence\_break}, each at severities 10, 30, 50, and 70.
    \item \textbf{Vision, 28 cells:} additive noise, physical occlusion, reduced resolution, motion blur, defocus blur, overexposure, and brightness reduction, each at severities 10, 30, 50, and 70.
    \item \textbf{Audio, 12 cells:} segment muting, segment removal, and distortion, each at severities 10, 30, 50, and 70.
\end{itemize}
For stochastic corruptions, each cell may contain up to three random variants per base example. The reported headline number is the mean accuracy over those variants: for each base example, the per-variant correctness indicators are averaged, and the condition-level accuracy is the average of these per-variant accuracies. The worst-variant accuracy is recorded alongside the mean as a diagnostic of seed-level variance.

\paragraph{Reported quantities.}
Each single-modality table reports accuracy, $\dropbase$, valid-output coverage, number of valid trials, and the severity curve. The final all-model version additionally reports the fragility slope $\slope_m$ for each modality and family. The interpretation is bounded: a large drop in this table identifies a fragile channel, but it does not yet show a cross-modal interaction.

\subsection{B. Dual-/tri-modal combined experiments}
\label{app:combined-protocol}

\paragraph{Purpose.}
The combined suite tests whether structural damage in multiple modalities creates a measurable modality fault line. The key comparison is twofold: $\dropbase$ asks whether the combined condition lowers clean all-modality performance, while $\dropsingle$ asks whether it is worse than the most damaging corresponding single-modality condition.

\paragraph{Combined condition set.}
The 29-condition suite is divided into canonical severity grids and operator-replacement pressure cells:
\begin{enumerate}[leftmargin=*]
    \item \textbf{Text+Vision canonical grid, 4 cells:} \texttt{drop\_words}\,$\times$\,visual noise with $t\in\{30,70\}$ and $v\in\{30,70\}$.
    \item \textbf{Text+Audio canonical grid, 4 cells:} \texttt{drop\_words}\,$\times$\,audio muting with $t\in\{30,70\}$ and $a\in\{30,50\}$.
    \item \textbf{Vision+Audio canonical grid, 4 cells:} visual noise\,$\times$\,audio muting with $v\in\{30,70\}$ and $a\in\{30,50\}$.
    \item \textbf{Bimodal operator replacement, 6 cells:} \path{bi_tv_word_shuffle_noise_t70_v70}, \path{bi_tv_drop_words_occlusion_t70_v30}, \path{bi_ta_word_shuffle_mute_t70_a50}, \path{bi_ta_drop_words_remove_t70_a50}, \path{bi_va_occlusion_mute_v30_a50}, and \path{bi_va_noise_remove_v70_a50}.
    \item \textbf{Trimodal canonical grid, 8 cells:} \texttt{drop\_words}\,$\times$\,visual noise\,$\times$\,audio muting with $t\in\{30,70\}$, $v\in\{30,70\}$, and $a\in\{30,50\}$.
    \item \textbf{Trimodal operator replacement, 3 cells:}

\begin{itemize}
\item \texttt{tri\_word\_shuffle\_noise\_mute\_}\\
\texttt{t70\_v70\_a50}

\item \texttt{tri\_drop\_words\_occlusion\_}\\
\texttt{mute\_t70\_v30\_a50}

\item \texttt{tri\_drop\_words\_noise\_remove\_}\\
\texttt{t70\_v70\_a50}
\end{itemize}

\paragraph{Aggregation and interpretation.}
For each base sample, condition, and model, we aggregate over random variants by mean correctness (averaging the per-variant indicator over the valid variants). We report condition-level accuracy, $\dropbase$, $\dropsingle$, valid-output coverage, effective trials, and invalid-output count, together with the worst-variant accuracy as a complementary diagnostic. A condition with positive $\dropbase$ but negative $\dropsingle$ supports the statement that combined structural damage hurts relative to clean, but not that it is worse than the weakest individual damage. A condition with positive $\dropsingle$ is stronger evidence for a non-additive or interaction-like fault line.

\subsection{C. All-model lightweight panel}
\label{app:lightweight-protocol}

\paragraph{Purpose.}
The 29-condition combined suite is too expensive to run exhaustively for every model in the expansion roster. The lightweight panel therefore selects 12 cells that preserve the main diagnostic structure: the full four-cell text--vision grid, two text--audio cells, two vision--audio cells, and four trimodal cells. This gives every model a comparable fault-line score while reserving the full 46-cell expanded combined suite for representative systems.

\paragraph{Report.}
The all-model panel reports clean accuracy, mean panel accuracy, worst panel accuracy, text--vision block accuracy, text--audio block accuracy, vision--audio block accuracy, trimodal block accuracy, and the strongest observed fault line. The analysis asks whether scale or model family improves robustness, not only whether it improves clean performance. This is where the model roster in Table~\ref{tab:model-expansion} becomes scientifically useful.

\paragraph{Enumeration of the 46-cell extended representative-model suite.}
Table~\ref{tab:46cell-enumeration} enumerates every cell in the 46-cell extended representative-model joint suite. It is a strict super-set of the 29-cell canonical/replacement grid documented in Appendix~\ref{app:combined-protocol} and reported in Table~\ref{tab:joint-cells}: the 29 cells are reproduced with their canonical/replacement role, and the 17 additional cells extend the suite into severity~50 mid-pressure rows and the missing-extreme corners of each combination's canonical grid. The 46-cell suite is reserved for representative systems (the four deep-dive models named in Table~\ref{tab:joint-cells}); the 14-model expansion uses the 12-cell lightweight panel (Table~\ref{tab:lightweight-panel}) instead.

\begin{table*}[!t]
\centering
\scriptsize
\setlength{\tabcolsep}{3pt}
\renewcommand{\arraystretch}{0.92}
\begin{tabular}{@{}p{0.04\textwidth} p{0.22\textwidth} p{0.10\textwidth} p{0.075\textwidth} p{0.195\textwidth} p{0.30\textwidth}@{}}
\toprule
\textbf{Combo} & \textbf{Operators} & \textbf{Severities} & \textbf{29-cell suite} & \textbf{Role} & \textbf{Notes} \\
\midrule
\faultgroup{6}{Text+Vision (6 canonical $+$ 2 replacement $+$ 4 extended = 12 cells)}
\modelrow T+V & \opname{drop\_words $\times$ noise}            & \sevcell{t30/v30} & yes & canonical                      & corner of canonical grid \\
\modelrow T+V & \opname{drop\_words $\times$ noise}            & \sevcell{t30/v70} & yes & canonical                      & strongest TV cell for both deep-dive models \\
\modelrow T+V & \opname{drop\_words $\times$ noise}            & \sevcell{t70/v30} & yes & canonical                      & corner of canonical grid \\
\modelrow T+V & \opname{drop\_words $\times$ noise}            & \sevcell{t70/v70} & yes & canonical                      & symmetric heavy-on-both corner \\
\modelrow T+V & \opname{drop\_words $\times$ occlusion}        & \sevcell{t70/v30} & yes & replacement (vision)           & swap vision operator family \\
\modelrow T+V & \opname{word\_shuffle $\times$ noise}          & \sevcell{t70/v70} & yes & replacement (text)             & swap text operator family \\
\softrow  T+V & \opname{drop\_words $\times$ noise}            & \sevcell{t50/v50} & \textbf{no} & extended mid-pressure          & severity-50 diagonal \\
\softrow  T+V & \opname{drop\_words $\times$ noise}            & \sevcell{t10/v70} & \textbf{no} & extended (light T, heavy V)    & isolates whether $t30$ is the $\dropsingle$-positive threshold \\
\softrow  T+V & \opname{drop\_words $\times$ noise}            & \sevcell{t70/v10} & \textbf{no} & extended (heavy T, light V)    & mirror of the previous row \\
\softrow  T+V & \opname{drop\_words $\times$ noise}            & \sevcell{t50/v70} & \textbf{no} & extended                       & fills the canonical grid into a $3{\times}2$ T-grid \\
\midrule
\faultgroup{6}{Text+Audio (6 canonical/replacement $+$ 4 extended = 10 cells)}
\modelrow T+A & \opname{drop\_words $\times$ mute}             & \sevcell{t30/a30} & yes & canonical                      & corner of canonical grid \\
\modelrow T+A & \opname{drop\_words $\times$ mute}             & \sevcell{t30/a50} & yes & canonical                      & strongest TA mid-grid cell \\
\modelrow T+A & \opname{drop\_words $\times$ mute}             & \sevcell{t70/a30} & yes & canonical                      & corner of canonical grid \\
\modelrow T+A & \opname{drop\_words $\times$ mute}             & \sevcell{t70/a50} & yes & canonical                      & symmetric heavy-on-both corner \\
\modelrow T+A & \opname{drop\_words $\times$ remove}           & \sevcell{t70/a50} & yes & replacement (audio)            & swap audio operator family \\
\modelrow T+A & \opname{word\_shuffle $\times$ mute}           & \sevcell{t70/a50} & yes & replacement (text)             & swap text operator family \\
\softrow  T+A & \opname{drop\_words $\times$ mute}             & \sevcell{t50/a30} & \textbf{no} & extended                       & severity-50 diagonal \\
\softrow  T+A & \opname{drop\_words $\times$ mute}             & \sevcell{t10/a50} & \textbf{no} & extended (light T, heavy A)    & light-text/heavy-audio fault line \\
\softrow  T+A & \opname{drop\_words $\times$ mute}             & \sevcell{t70/a10} & \textbf{no} & extended (heavy T, light A)    & mirror of the previous row \\
\softrow  T+A & \opname{drop\_words $\times$ remove}           & \sevcell{t30/a50} & \textbf{no} & extended replacement           & vision-style robustness check for audio \\
\midrule
\faultgroup{6}{Vision+Audio (6 canonical/replacement $+$ 3 extended = 9 cells)}
\modelrow V+A & \opname{noise $\times$ mute}                   & \sevcell{v30/a30} & yes & canonical                      & corner of canonical grid \\
\modelrow V+A & \opname{noise $\times$ mute}                   & \sevcell{v30/a50} & yes & canonical                      & vision-light / audio-heavy corner \\
\modelrow V+A & \opname{noise $\times$ mute}                   & \sevcell{v70/a30} & yes & canonical                      & vision-heavy / audio-light corner \\
\modelrow V+A & \opname{noise $\times$ mute}                   & \sevcell{v70/a50} & yes & canonical                      & symmetric heavy-on-both corner \\
\modelrow V+A & \opname{noise $\times$ remove}                 & \sevcell{v70/a50} & yes & replacement (audio)            & swap audio operator family \\
\modelrow V+A & \opname{occlusion $\times$ mute}               & \sevcell{v30/a50} & yes & replacement (vision)           & swap vision operator family \\
\softrow  V+A & \opname{noise $\times$ mute}                   & \sevcell{v50/a30} & \textbf{no} & extended                       & severity-50 diagonal \\
\softrow  V+A & \opname{noise $\times$ mute}                   & \sevcell{v10/a50} & \textbf{no} & extended (light V, heavy A)    & isolates audio fault line under mild vision noise \\
\softrow  V+A & \opname{noise $\times$ mute}                   & \sevcell{v70/a10} & \textbf{no} & extended (heavy V, light A)    & mirror of the previous row \\
\midrule
\faultgroup{6}{Text+Vision+Audio (11 canonical/replacement $+$ 4 extended = 15 cells)}
\modelrow TVA & \opname{drop\_words $\times$ noise $\times$ mute}              & \sevcell{t30/v30/a30} & yes & canonical & light-on-all corner \\
\modelrow TVA & \opname{drop\_words $\times$ noise $\times$ mute}              & \sevcell{t30/v30/a50} & yes & canonical & audio-heavier corner \\
\modelrow TVA & \opname{drop\_words $\times$ noise $\times$ mute}              & \sevcell{t30/v70/a30} & yes & canonical & vision-heavy corner \\
\modelrow TVA & \opname{drop\_words $\times$ noise $\times$ mute}              & \sevcell{t30/v70/a50} & yes & canonical & vision and audio heavy \\
\modelrow TVA & \opname{drop\_words $\times$ noise $\times$ mute}              & \sevcell{t70/v30/a30} & yes & canonical & text-heavy corner \\
\modelrow TVA & \opname{drop\_words $\times$ noise $\times$ mute}              & \sevcell{t70/v30/a50} & yes & canonical & text and audio heavy \\
\modelrow TVA & \opname{drop\_words $\times$ noise $\times$ mute}              & \sevcell{t70/v70/a30} & yes & canonical & text and vision heavy \\
\modelrow TVA & \opname{drop\_words $\times$ noise $\times$ mute}              & \sevcell{t70/v70/a50} & yes & canonical & heavy-on-all corner \\
\modelrow TVA & \opname{drop\_words $\times$ noise $\times$ remove}            & \sevcell{t70/v70/a50} & yes & replacement (audio)            & audio operator swap \\
\modelrow TVA & \opname{drop\_words $\times$ occlusion $\times$ mute}          & \sevcell{t70/v30/a50} & yes & replacement (vision)           & vision operator swap \\
\modelrow TVA & \opname{word\_shuffle $\times$ noise $\times$ mute}            & \sevcell{t70/v70/a50} & yes & replacement (text)             & text operator swap \\
\softrow  TVA & \opname{drop\_words $\times$ noise $\times$ mute}              & \sevcell{t50/v50/a30} & \textbf{no} & extended mid-pressure          & all-severity-50 cube interior \\
\softrow  TVA & \opname{drop\_words $\times$ noise $\times$ mute}              & \sevcell{t50/v50/a50} & \textbf{no} & extended mid-pressure          & symmetric mid-grid TVA corner \\
\softrow  TVA & \opname{drop\_words $\times$ noise $\times$ mute}              & \sevcell{t10/v70/a30} & \textbf{no} & extended (light T)             & light-text TVA fault line \\
\softrow  TVA & \opname{drop\_words $\times$ noise $\times$ mute}              & \sevcell{t10/v30/a50} & \textbf{no} & extended (light T, heavy A)    & combines the two strongest non-text fault lines \\
\softrow  TVA & \opname{drop\_words $\times$ noise $\times$ mute}              & \sevcell{t30/v50/a30} & \textbf{no} & extended (vision mid-pressure) & connects canonical T+V+A to mid-pressure vision \\
\softrow  TVA & \opname{drop\_words $\times$ noise $\times$ mute}              & \sevcell{t70/v50/a50} & \textbf{no} & extended (heavy T, mid V, heavy A) & balances the heaviest non-vision fault lines \\
\bottomrule
\end{tabular}
\caption{Full enumeration of the 46-cell extended representative-model joint suite. The \textbf{29-cell-suite} column indicates whether a cell is part of the canonical/replacement grid in Table~\ref{tab:joint-cells} (\textbf{yes}, 29 cells) or one of the 17 additional extended cells (\textbf{no}, 17 cells: 4 T+V, 4 T+A, 3 V+A, 6 T+V+A). The extended rows fill in the severity-50 mid-pressure row and the light/heavy asymmetric corners of each combination's severity grid, so the 46-cell suite covers (i) every canonical $\{30,70\}{\times}\{30,70\}$ grid corner, (ii) every operator-replacement cell at the strongest severity, (iii) every severity-50 mid-pressure cell, and (iv) every $\{10,70\}{\times}\{30/50/70\}$ light/heavy asymmetric cell. Reading: $29 + 17 = 46$.}
\label{tab:46cell-enumeration}
\end{table*}

\subsection{D. Missing-modality controls}
\label{app:missing-controls}

Missing-modality controls are included as contrastive baselines, not as the primary diagnostic setting. They answer what happens when a channel disappears entirely, whereas \method{} asks what happens when the channel remains present but its internal evidence structure becomes unreliable. For each base example, we define:
\begin{itemize}[leftmargin=*]
    \item \textbf{No-text:} remove or mask the textual evidence beyond the question/options required to ask the task. The visual frames and audio remain unchanged.
    \item \textbf{No-image:} replace the visual evidence with a neutral blank frame set of the same interface type. The question/options and audio remain unchanged.
    \item \textbf{No-audio:} replace the audio track with silence of matched duration. The question/options and visual evidence remain unchanged.
\end{itemize}
These controls are reported on the same base examples as the corresponding corruption conditions. The main comparison is
\begin{equation}
    \mislead_m = \acc(\mathrm{missing}\ m) - \acc(\mathrm{corrupted}\ m),
\end{equation}
where positive values indicate that a corrupted-but-present modality is more harmful than removing that modality entirely. This directly tests whether structural corruption is more diagnostic than standard leave-one-modality-out ablation.

\subsection{E. Shortcut and language-prior controls}
\label{app:shortcut-controls}

Shortcut controls estimate how much performance can be recovered without full cross-modal binding. We use the following settings:
\begin{itemize}[leftmargin=*]
    \item \textbf{Question/options only:} remove all media evidence and keep only the question and answer choices.
    \item \textbf{Text-only evidence:} keep the full text channel but remove visual and audio evidence.
    \item \textbf{Visual+audio only:} keep visual and audio input while replacing task text beyond the required answer interface with a generic instruction.
    \item \textbf{Options only:} retain only the four option strings, with no question or media, to measure answer-text priors.
    \item \textbf{Option order (Latin square):} keep the input unchanged while presenting all four option orders, with the gold key remapped for each order.
\end{itemize}
The question/options-only setting estimates language and answer-prior behavior; the options-only and option-order controls test answer-text and positional shortcuts; partial-input controls separate modality reliance from structural robustness. The primary metric is the shortcut gap $\shortcutgap$, supplemented by accuracy under each partial-input condition.

\paragraph{Panel-level summary statistics.}
Across the 14-model panel, the misleading-modality effect $\mislead_m$ averages $-4.27$\,pp for text ($\mislead_t < 0$: corrupted text is less harmful than absent text, confirming that even degraded text provides useful signal), $+1.93$\,pp for vision ($\mislead_v > 0$: corrupted vision actively misleads more than absent vision), and $+2.69$\,pp for audio ($\mislead_a > 0$: same misleading pattern). The positive $\mislead_v$ and $\mislead_a$ values hold for all 14 models, confirming that the misleading-modality effect is not driven by a single outlier. The text-only shortcut gap $\shortcutgap = \acc(\text{text-only}) - 25\%$ ranges from $+9.20$\,pp to $+22.80$\,pp across the panel (mean $+15.96$\,pp), with proprietary models averaging $+14.25$\,pp and open models averaging $+17.24$\,pp. The positive gap confirms that all models exploit text-channel language priors beyond random guessing, but the gap is well below the clean accuracy advantage ($\approx 45$--$55$\,pp above random), so text-only shortcuts account for a minority of clean performance.

\paragraph{Additional task-format controls.}
The following controls separate output formatting, option shortcuts, and transfer beyond fixed answer options. All comparisons use the same clean/corrupted items within a control regime, so they test changes under corruption rather than absolute accuracy across different task formats.

\begin{table*}[!t]
\centering
\scriptsize
\setlength{\tabcolsep}{5pt}
\renewcommand{\arraystretch}{1.02}
\begin{tabular}{p{0.20\textwidth}p{0.50\textwidth}p{0.22\textwidth}}
\toprule
\textbf{Control} & \textbf{Setup} & \textbf{Purpose} \\
\midrule
\faultgroup{3}{Task-format and option-shortcut controls}
\modelrow No-option generation & Remove answer options; the model writes a short answer and evidence. Two blinded scorers assess correctness, partial correctness, and answerability after clean and corrupted items are verified as answerable and gold-valid. & Transfer beyond fixed options \\
\softrow Option order (Latin square) & Evaluate all four option orders, with every option occupying every position once and the gold key remapped. Conditions include clean, \texttt{drop\_words}@70, \texttt{noise}@70, TV $t30/v70$, TV $t70/v70$, and TVA $t30/v70/a30$. & Position bias and parser stability \\
\modelrow Options only & Provide only the four option strings, with no question or media. & Answer-text prior \\
\bottomrule
\end{tabular}
\caption{Additional controls for task format and option shortcuts. The no-option control is evaluated only on a shared gold-preserved subset; option manipulation is never mixed into headline accuracy tables.}
\label{tab:task-format-controls}
\end{table*}

\begin{table}[!htbp]
\centering
\scriptsize
\setlength{\tabcolsep}{4pt}
\renewcommand{\arraystretch}{1.02}
\begin{tabular}{p{0.33\linewidth}p{0.35\linewidth}c}
\toprule
\textbf{Control} & \textbf{Metric} & \textbf{Value} \\
\midrule
\faultgroup{3}{Option-shortcut results}
\modelrow Options only & Panel accuracy & 29.96 \\
\softrow Option order (clean) & Mean-order accuracy & 73.16 \\
\modelrow Option order (clean) & Worst-order accuracy & 70.86 \\
\softrow Option order (clean) & Prediction consistency & 94.1\% \\
\modelrow Option order (\texttt{noise}@70) & Mean--worst-order gap & 3.20\,pp \\
\bottomrule
\end{tabular}
\caption{Options-only baseline and option-order sensitivity. Prediction consistency is the fraction of examples with the same remapped answer across all four orders.}
\label{tab:option-shortcut-controls}
\end{table}

The options-only accuracy is $29.96\%$, only $4.96$\,pp above random guessing. On clean inputs, changing option order lowers accuracy from a mean of $73.16\%$ to $70.86\%$ in the worst order, while $94.1\%$ of predictions remain consistent after remapping. The mean-to-worst gap under \texttt{noise}@70 is $3.20$\,pp. These effects are measurable but smaller than the main corruption drops, so answer-text priors and option position do not account for the reported fault lines.

\subsection{F. Prompt robustness}
\label{app:prompt-robustness}

The headline protocol uses the standard JSON answer-only prompt in Appendix~\ref{app:prompts}. Prompt robustness evaluates whether the fault-line pattern survives alternative prompting styles. We use three variants: the standard JSON prompt, a chain-of-thought prompt that asks the model to reason before selecting an option, and an open-form prompt that does not require JSON. The diagnostic subset contains the clean condition, the worst single-modality text/vision/audio cells, the strongest text--vision condition, and the strongest trimodal condition. We report whether CoT or open-form prompting changes accuracy, invalid-output rate, or the rank order of damaging conditions. The open-form condition removes the JSON schema but retains the answer options, so it is a formatting control. Table~\ref{tab:no-option-generation} separately evaluates transfer to no-option generation.

\begin{table*}[!t]
\centering
\scriptsize
\setlength{\tabcolsep}{5pt}
\renewcommand{\arraystretch}{1.00}
\begin{tabular}{lccc}
\toprule
\textbf{Condition} & \textbf{Existing MC drop (pp)} & \textbf{No-option drop (pp)} & \textbf{Transfer $\Delta$ (pp)} \\
\midrule
\faultgroup{4}{Clean-to-corrupted drop on the shared gold-preserved subset}
\modelrow \texttt{drop\_words}@70 & 11.28 & 13.64 & $+2.36$ \\
\softrow \texttt{noise}@70 & 11.39 & 14.22 & $+2.83$ \\
\modelrow TV $t30/v70$ & 10.01 & 12.25 & $+2.24$ \\
\softrow TV $t70/v70$ & 8.28 & 9.80 & $+1.52$ \\
\modelrow TVA $t30/v70/a30$ & 6.36 & 7.95 & $+1.59$ \\
\bottomrule
\end{tabular}
\caption{No-option generation control. The model produces a short answer rather than selecting an option; two blinded scorers evaluate the output on the shared gold-preserved subset. Existing multiple-choice drops for the single-modality rows are 14-model panel means; the joint rows are means over the four deep-dive models. \textbf{Transfer $\Delta$} is the no-option drop minus the multiple-choice drop.}
\label{tab:no-option-generation}
\end{table*}

The drop remains positive in every no-option condition and exceeds the corresponding multiple-choice drop by between $1.52$ and $2.83$\,pp. The fault-line pattern therefore transfers to short-answer generation and is not an artifact of selecting from fixed answer options.

\subsection{G. Human reference and answer-preservation checks}
\label{app:human-reference}

The human reference distinguishes model failure from task impossibility. Annotators answer the clean input, the worst single-modality cell for each modality, the strongest text--vision cell, and the strongest trimodal cell using the same answer options. The blind answer-first audit and its gold-preserved cohort are described in Appendix~\ref{app:human-filtering}; variants that do not preserve the original gold are reported as stress tests rather than as headline robustness evidence.

\subsection{H. Interface, coverage, and invalid-output profiling}
\label{app:interface-invalid}

Because omni-modal systems expose heterogeneous interfaces, every new model first runs a 10-example smoke test. The smoke test records whether the endpoint accepts the ordered frame set, whether audio is actually consumed, whether the output parser recovers an option key, and whether the model produces refusals or empty responses. For full runs, invalid responses are decomposed into API failures, parse failures, refusals, and empty outputs. Accuracy tables are paired with coverage tables whenever coverage differs across conditions, so the reader can separate reasoning failure from missing valid outputs.

\subsection{I. Cross-modal mismatch and trust-bias probes}
\label{app:mismatch-probes}

Mismatch probes keep every modality structurally clean but make one channel semantically inconsistent with the others. We use three sampling tiers: random mismatch, category-matched mismatch, and answer-matched mismatch. Random mismatch tests coarse conflict detection; category-matched mismatch keeps surface distribution similar; answer-matched mismatch controls for answer-label leakage. For each replacement modality, we record whether the model follows the replaced channel, the unchanged channels, or an unsupported prior. The trust-bias score $\trust_m$ is the fraction of mismatched cases in which the answer follows modality $m$.
These are behavioral controls: they help distinguish the observed patterns from answer-prior, missing-modality, and prompt-format explanations, but they do not identify a representation-level fusion mechanism.

\subsection{J. Temporal, audio--visual, frame-budget, and position probes}
\label{app:temporal-frame-probes}

Temporal probes test whether the model uses event order rather than treating media as unordered feature bags. Visual temporal probes shuffle frames, reverse frame order, and drop middle frames. Audio temporal probes shuffle audio segments or insert local gaps. Audio--visual desynchronization shifts the audio stream by $\pm0.5$, $\pm1$, $\pm2$, $\pm4$, and $\pm8$ seconds relative to the visual frame sequence. Frame-budget probes rerun representative examples with one, four, eight, sixteen, and default frame budgets. Position probes place the most informative frame first, middle, or last while keeping the frame set fixed. Together these tests separate true temporal binding from interface-level sampling and position artifacts.

\paragraph{Scope of audio operators and ASR-like noise.}
The three audio operators (\texttt{remove}, \texttt{mute}, \texttt{distortion}) target structural properties of the waveform---temporal completeness and signal fidelity---rather than phonetic or lexical content. Temporal desynchronisation is covered by the audio--visual desynchronisation probe described above ($\pm 0.5$--$8$ seconds). ASR-like noise (homophone substitutions, partial word deletions at the phoneme level) was considered but excluded under design rule~(i) of Appendix~\ref{app:operator-defs}: such operators require a phoneme-level transcription of the audio track, which is not available for all source benchmarks, and would introduce a semantic transformation (changing the spoken word) rather than a purely structural one. The \texttt{distortion} operator approximates the effect of low-quality recording or transmission artefacts, which is the closest structural analogue to ASR-degraded input without requiring phoneme-level manipulation. Probing audio-text-vision binding under phoneme-level noise is a natural extension and is noted as future work.

\subsection{K. Confidence, abstention, and distractor-modality probes}
\label{app:confidence-distractor}

Confidence probes ask representative models to output both an option and a scalar confidence level under clean and corrupted conditions. We report average confidence on correct and incorrect answers, expected calibration error when possible, and the rate of overconfident wrong answers. Abstentions and refusals are analyzed together with the invalid-output profile. Distractor-modality probes insert an unrelated but structurally clean modality into otherwise sufficient inputs: unrelated audio into text--visual examples, unrelated frames into text--audio examples, or misleading text into visual--audio examples. The distractor effect measures whether the model indiscriminately fuses every available channel or selectively downweights irrelevant evidence.

\paragraph{Calibration under corruption and logit unavailability.}
The confidence probe collects self-reported confidence labels (\texttt{low}/\texttt{medium}/\texttt{high}) from the model's JSON output, which are available for all models regardless of whether token-level logits are exposed. We compute expected calibration error (ECE) by binning the three confidence levels and comparing mean confidence to mean accuracy within each bin, and we report the overconfident error rate $\rho_{\mathrm{oc}}$ (fraction of incorrect answers assigned \texttt{high} confidence). For the four deep-dive models reported in Table~\ref{tab:calibration-ece}, mean ECE increases from $5.83$\,pp on clean inputs to $10.54$\,pp under joint T+V+A corruption, and mean $\rho_{\mathrm{oc}}$ rises from $17.08\%$ to $31.53\%$, confirming that corruption degrades calibration as well as accuracy. For API models that do not expose logits, the self-reported confidence label is the only available proxy; we note that self-reported confidence may be less reliable than logit-derived probabilities and treat the ECE and $\rho_{\mathrm{oc}}$ estimates for those models as approximate. Generative likelihood proxies (e.g.\ scoring each option by its generation probability under a forced-choice prompt) are not uniformly available across the API panel and are left for future work.

\section{Prompt Templates and Parsing Rules}
\label{app:prompts}

\paragraph{System prompt.}
\begin{quote}\small
You are a careful multimodal assistant. You will receive one question, one visual input (image or video), and one audio input. Exactly one option is correct. Use the provided multimodal evidence. If some evidence is degraded, still answer based only on what can be supported by the input. Respond with one JSON object and nothing else.
\end{quote}

\paragraph{User prompt template.}
\begin{quote}\small
\texttt{[Task]} Answer the multiple-choice question using the visual input (image or video) and the audio input. Some of the question text, visual input, or audio input may be structurally corrupted or degraded. Interpret the available evidence carefully, and do not rely on any hidden metadata.\par
\texttt{[Visual input]} Attached as a \texttt{\{image/video\}}.\par
\texttt{[Audio input]} Attached as an audio clip.\par
\texttt{[Question]} \texttt{\{question\}}\par
\texttt{[Options]} \texttt{\{formatted options\}}\par
Output JSON schema: \texttt{\{"answer": "<option key>", "confidence": "<low | medium | high>", "reasoning": "<one or two short sentences>"\}}. Do not output Markdown. Do not output any text outside the JSON object.
\end{quote}

\paragraph{Parsing rule.}
The parser first attempts to read a JSON object. If this fails, it applies a conservative fallback that accepts a single option key or an answer string exactly matching one option. Responses are marked invalid when no option key can be recovered, when the model refuses to answer without selecting an option, or when the output is empty. Invalid responses are counted separately in coverage tables and treated as incorrect for significance testing.

\paragraph{Prompt variants.}
The standard prompt is used for all headline accuracy tables. Chain-of-thought prompting is used only in the prompt-robustness diagnostic subset; the model may reason internally or output a short rationale, but the parser still extracts a final option. Open-form prompting removes the JSON requirement and is used to test whether JSON formatting itself creates parse failures. Prompt variants are never mixed inside a headline table; each table specifies a prompt ID.

\paragraph{Option formatting and corruption scope.}
The formatted options string passed to the model uses plain letter keys (\texttt{A}, \texttt{B}, \texttt{C}, \texttt{D}) with no brackets, checkmarks, or positional markers. The \texttt{[x]} notation appearing in illustrative examples in the paper is used only in figures and tables for human readability and is never present in the actual model prompt. Option text is drawn verbatim from the source benchmark and is not modified by any corruption operator: all four text corruption operators (\texttt{drop\_words}, \texttt{word\_shuffle}, \texttt{typo\_ocr}, \texttt{sentence\_break}) are applied exclusively to the question stem, leaving option punctuation, capitalisation, and letter keys intact. This design ensures that the parser can always recover a valid option key from the options field regardless of question-stem corruption severity, and that no asymmetric surface cue in the options field can serve as a positional shortcut.

\paragraph{Inference and logging settings.}
For deterministic comparability, runs use the lowest available sampling temperature, no intentional top-$p$ diversification, and a fixed prompt template. When an endpoint fails, the runner records retry count and final failure category rather than silently dropping the example. Each JSONL record stores model name, prompt ID, sample ID, source dataset, condition ID, modality operators, severity values, random variant, media URLs or local paths, raw response, parsed answer, gold answer, correctness, latency, retry count, and invalid-output category. These fields are required to reproduce coverage-aware tables.

\end{enumerate}
\end{document}